\documentclass[11pt]{article}
\usepackage[utf8]{inputenc}
\usepackage[margin=1.0in]{geometry}  

\usepackage{cite}

\usepackage{graphics} 
\usepackage{epstopdf}

\usepackage{amsmath}
\usepackage{amssymb}

\newtheorem{remark}{Remark}

\usepackage{algorithmic}
\usepackage{algorithm}

\usepackage{array}

\usepackage[caption=false,font=footnotesize]{subfig}
\usepackage[export]{adjustbox}
\usepackage{multicol} 

\usepackage{url}

\usepackage[shortlabels]{enumitem}

\usepackage{xcolor}

\usepackage[title]{appendix}

\usepackage{authblk}

\usepackage{xspace}
\newcommand{\namingConvention}[1]{#1\xspace}

\newcommand{\CFTOCshort}{\namingConvention{CFTOC}}
\newcommand{\CFTOClong}{\namingConvention{Constrained Finite Time Optimal Control Problem}}

\newcommand{\SQPlong}{\namingConvention{Sequential Quadratic Programming}}
\newcommand{\SQPshort}{\namingConvention{SQP}}
\newcommand{\QPlong}{\namingConvention{Quadratic Program}}
\newcommand{\QPshort}{\namingConvention{QP}}
\newcommand{\RTIlong}{\namingConvention{Real Time Iteration}}
\newcommand{\RTIshort}{\namingConvention{RTI}}
\newcommand{\SAARTIlong}{\namingConvention{Sampling Augmented Adaptive RTI}}
\newcommand{\SAARTIshort}{\namingConvention{SAA-RTI}}

\begin{document}
\title{\Huge{Traction Adaptive Motion Planning and \\ Control at The Limits of Handling}}
\author[$^{1}$]{Lars~Svensson}
\author[$^{2}$]{Monimoy~Bujarbaruah}
\author[$^{3}$]{Arpit~Karsolia}
\author[$^{4}$]{Christian~Berger}
\author[$^{1}$]{Martin~T\"orngren}

\affil[$^{1}$]{Department of Machine Design, KTH Royal Institute of Technology}
\affil[$^{2}$]{Model Predictive Control Laboratory, University of California Berkeley}
\affil[$^{3}$]{Department of Electrical Engineering, Chalmers University of Technology}
\affil[$^{4}$]{Department of Computer Science and Engineering, University of Gothenburg}


\maketitle

\begin{abstract}
\noindent In this paper, we address the problem of motion planning and control at the limits of handling, under locally varying traction conditions. We propose a novel solution method where traction variations over the prediction horizon are represented by time-varying tire force constraints, derived from a predictive friction estimate. A constrained finite time optimal control problem is solved in a receding horizon fashion, imposing these time-varying constraints. Furthermore, our method features an integrated sampling augmentation procedure that addresses the problems of infeasibility and sensitivity to local minima that arise at abrupt constraint alterations, e.g., due to sudden friction changes.

We validate the proposed algorithm on a Volvo FH16 heavy-duty vehicle, in a range of critical scenarios. Experimental results indicate that traction adaptive motion planning and control improves the vehicle's capacity to avoid accidents, both when adapting to low local traction, by ensuring dynamic feasibility of the planned motion, and when adapting to high local traction, by realizing high traction utilization. 
\end{abstract}

\section{Introduction}
\label{sec:intro}
\begin{figure}[t]
\captionsetup[subfigure]{}
\centering
    \subfloat[Dry road surface]{%
        \includegraphics[width=0.6\columnwidth]{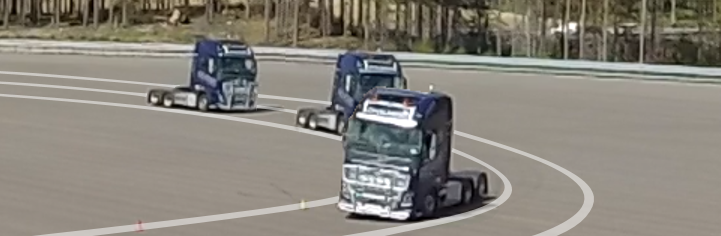}
        \label{fig:motionfig:dry}
    }\\
    \subfloat[Wet road surface]{%
        \includegraphics[width=0.6\columnwidth]{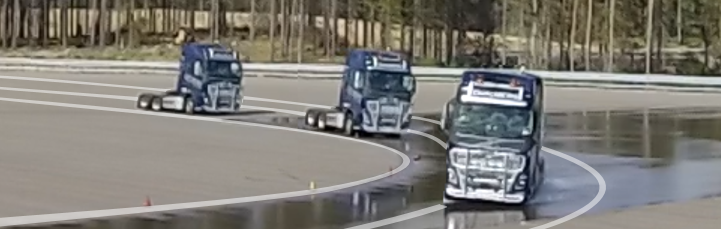}
        \label{fig:motionfig:wet}
    }
    \caption{Example of how operational conditions dictate the dynamic capabilities of a road vehicle. Entering the curve at 15m/s, the vehicle comfortably negotiates the corner when the road is dry, but slides out into the opposing lane when the road is wet, causing a hazardous situation.}
    \label{fig:motionfig:full}
    \vspace{-0.3cm}
\end{figure}

Automated driving and advanced driver assistance technology show increasing potential to improve safety and mobility of transportation systems in the future. A major challenge in driving automation is handling of critical situations, i.e., situations that appear suddenly, in which an accident is imminent. Critical situations may originate from multiple sources, e.g., unpredictable behavior of other traffic agents or rapid changes in the operational conditions. 
The steady progress in sensors, perception and prediction algorithms indicate that a subset of critical situations can be anticipated and avoided ahead of time. However, considering the inherent unpredictability of humans in the traffic environment and the wide range of operational conditions, it is unlikely that all critical situations can be anticipated and avoided ahead of time, without sacrificing efficiency and availability by enforcing overly cautious behavior. 
When a critical situation does occur, passenger comfort is no longer a priority, and if necessary, we wish to utilize the full physical capability of the vehicle to avoid the imminent accident. This motivates research in the field of motion planning and control at the limits of handling. Previous research efforts present methods for which performance approaches the physical limits \cite{liniger2015optimization, subosits2019qcqp}. 
However, the motion capability of a road vehicle is greatly affected by locally varying traction conditions of the road. This phenomenon, that can be intuitively understood by the simple experiment depicted in Fig.~\ref{fig:motionfig:full}, has been well studied and is included in standard vehicle dynamics literature \cite{fambro1997nchrp,rajamani2011vehicle}, but has only recently been considered in the context of motion planning at the limits of handling \cite{alsterda2019contingency}.

The work presented in this paper represents a further step towards traction adaptive motion planning and control at the limits of handling. We propose a novel optimization based framework, in which the locally varying traction limit is represented as time-varying constraints on tire forces. The constraints are updated online as a function of the estimated friction coefficient and dynamic tire loads. Furthermore, we employ sampling augmentation to address the non-convexity and infeasibility issues of the arising optimization problem. We evaluate the proposed method in comparison with an equivalent non-adaptive scheme, in a sequence of experiments on various road surfaces. 

The paper presents an extended and refined version of the method that was originally introduced in \cite{svensson2019adaptive}. The extension includes refinement of the mathematical framework, a more accurate model of the vehicle and its traction limitations, a novel, GPU accelerated method to generate feasible sample trajectories, and an extensive experimental evaluation. 

We summarize the contributions of the paper as follows: 

\begin{enumerate}
    \item A real-time capable algorithm for traction adaptive motion planning and control that produces optimal solutions with respect to time-varying tire force constraints.
    \item An experimental evaluation of traction adaptive motion planning and control, showing that the concept improves capacity to avoid accidents in a range of critical scenarios, when adapting to low as well as high local traction. 
\end{enumerate}


\section{Motivation}
\label{sec:prel}
Critical scenarios at the limits of handling put the following high level requirements on the local motion planning functionality: 
\begin{enumerate}
    \item A relatively complex vehicle model is required to accurately describe the vehicle in aggressive motions.
    \item A sufficiently long planning horizon is required to the handle the inertia and limited actuation capacity of the vehicle.
    \item A short planning time is required in order to react promptly to a dynamically changing traffic scene
    \item The planner has to be able to make discrete decisions, e.g., to go left or right of an obstacle
    \item Local traction variations should be considered.
\end{enumerate}
Requirements 1) - 4) have been been identified in previous academic works with state of the art approaches presented in Section \ref{sec:relwork}. Next, we motivate why 5) should also be considered in this context. 

From basic mechanics we have that for a single tire, the combined longitudinal, $F_x$, and lateral, $F_y$, force that can be exerted between tire and road is upper bounded by the friction coefficient $\mu$ times the normal force acting on the tire, $F_z$, i.e.,
\begin{equation}
    \sqrt{F_x^2 + F_y^2} \leq \mu F_z.
    \label{eq:forcelimit}
\end{equation}
This limit forms a circle in the lateral-longitudinal plane and is therefore commonly referred to as the friction circle\footnote{The friction circle is an approximation of the true force boundary of a tire. In reality, different tire-surface combinations will have differently shaped force boundaries. Therefore, the framework presented in Section \ref{sec:method} is designed to allow any convex shape to represent the force limit. In this study however, we use a circular representation for simplicity.}.

\begin{figure}[t]
    \centering
    \includegraphics[width=0.6\columnwidth]{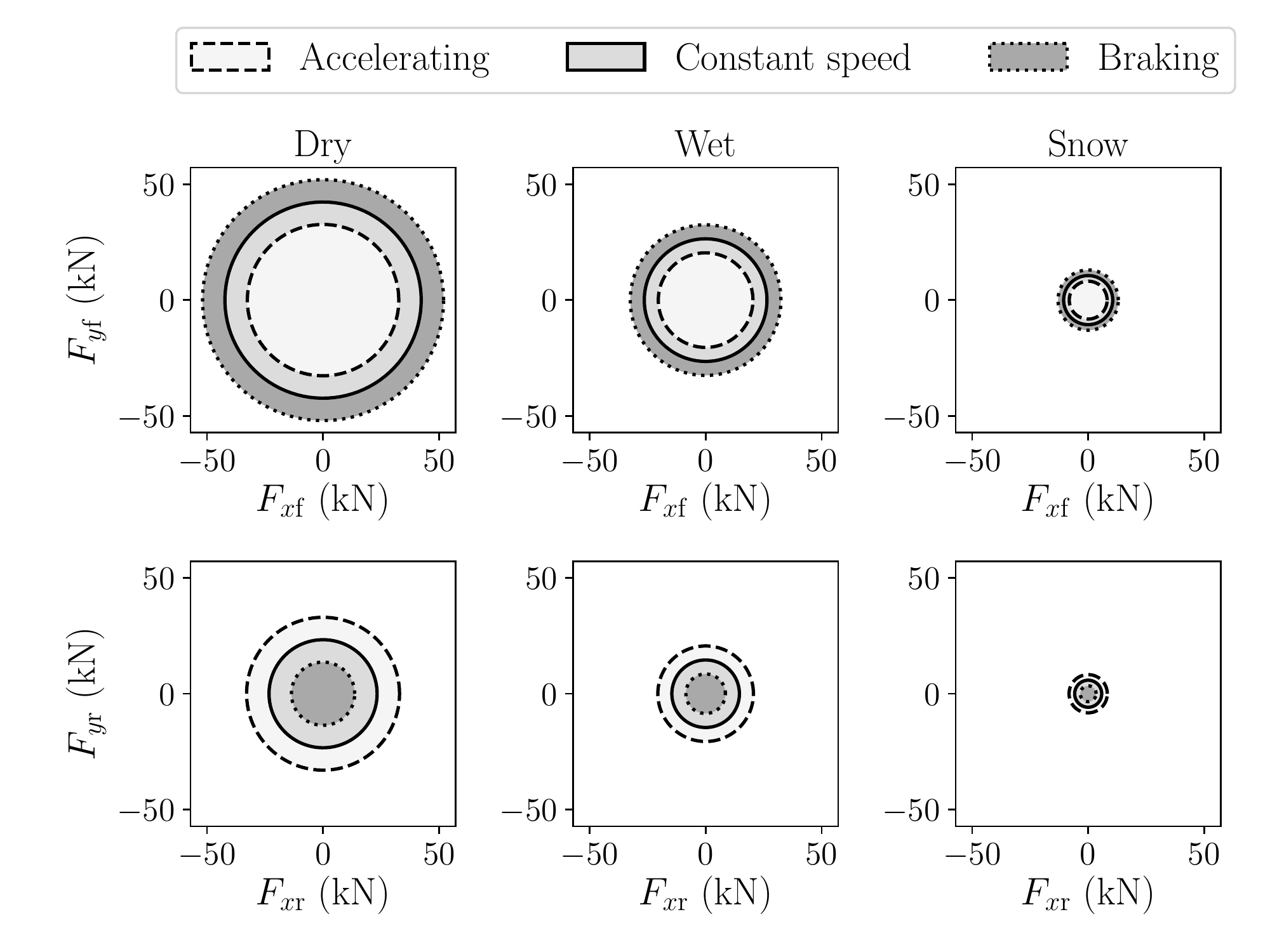}\\
    \caption{Numerical example of how tire force limits of our test vehicle vary at different operational conditions and motion cases, here represented by the following parameters: Dry: $\mu = 0.8$, Wet: $\mu = 0.5$, Snow: $\mu = 0.3$, Accelerating: $a_x = 0.5\mathrm{m/s^2}$, Constant speed: $a_x = 0\mathrm{m/s^2}$, Braking: $a_x = -0.5\mathrm{m/s^2}$}
    \label{fig:traction_limits}
    \vspace{-0.3cm}
\end{figure}

Both $\mu$ and $F_z$ typically vary substantially with time. The friction coefficient varies with, e.g., road surface, tire temperature and wear, and the normal force varies as a result of road inclination and dynamic forces caused by vehicle motion. Fig.~\ref{fig:traction_limits} highlights the impact of this phenomenon in terms of motion planning near the physical limits. The figure visualizes to what extent the force limits of the front and rear tires vary under different operational conditions and motion cases. 

The phenomenon presents an additional challenge to motion planning in critical situations. For example, not considering it and assuming static bounds on tire forces will have one of two consequences when planning aggressive evasive maneuvers
\begin{enumerate}
    \item \textbf{Planning of infeasible motions}. The real vehicle will be unable to track the plan. 
    \item \textbf{Reduced capability to avoid accident}. The planner will not be able to fully utilize the available traction to avoid an accident.
\end{enumerate}

Both 1) and 2) reduce the vehicle's capacity to avoid accidents in critical situations. This conclusion motivates us to pursue the development of algorithms for motion planning and control at the limits of handling, with a time-varying traction boundary. We refer to this concept as \textit{traction adaptive} motion planning and control. Next, we give a brief summary of relevant academic works in related fields. 

\section{Related Work}
\label{sec:relwork}

Motion planning algorithms for automated vehicles can be roughly classified into random sampling based, graph search based, trajectory roll-out and optimization based \cite{paden2016survey,gonzalez2016review}. Considering the special requirements in critical scenarios stated in Section \ref{sec:prel}, there is an evident computational trade-off between properties 1), 2) and 3), and therefore, efficient solving of the motion planning problem is a key challenge. Previous studies, e.g., \cite{liniger2015optimization,subosits2019qcqp} show that optimization based methods are well suited to handle such a computational trade-off, but have less favorable properties for requirement 4), compared to the other three classes. 

Random sampling based and graph search based algorithms are primarily designed to produce complex maneuver sequences \cite{paden2016survey}, which is typically not required for short-term evasive maneuvers in critical situations. Also, such algorithms are inherently sequential, restricting efficient implementation through parallel computing. Therefore, we delimit our related work section to optimization based and trajectory rollout algorithms, as well as a few relevant concepts from the fields of tire-road friction estimation and adaptive control.

\subsection{Optimization Based Methods}
\label{sec:relwork:optbased}
Optimization based methods dominate the field of motion planning at the limits of handling. Compared to other methods, they are not restricted to a discretized search space, scale comparatively well with model complexity \cite{ziegler2014trajectory}, and allow explicit representation of constraints on states and inputs.

The key concept of optimization based motion planning is to solve a \CFTOClong (\CFTOCshort) at each planning iteration, where the solution is the control and state sequence that obeys the constraints and minimizes a cost function over a fixed time horizon. In general, the \CFTOCshort representing a collision avoidance motion planning problem is nonconvex, due to the nonlinear vehicle dynamics and nonconvex state constraints representing obstacles \cite{boyd2004convex}. The nonconvexity of the problem renders the problem nontrivial to solve under the time requirements associated with critical evasive maneuvers. However, there exist several methods such as quasi-Newton algorithms with line search or \SQPlong (\SQPshort) that are capable of solving the problem given a near-optimal initial guess of the optimal trajectory \cite{betts1998survey,hargraves1987direct,von1992direct}. \SQPshort is an iterative approach in which a quadratic, and hence convex, approximation of the problem is formed and solved at each iteration. The approximation is obtained by performing a second order approximation of the cost function and a linearization of dynamics and constraints at the solution from the previous iteration. Thus, a \QPlong (\QPshort) is solved at each iteration. 
Diehl et al.~proposed selecting the solution trajectory from the previous planning iteration as  the initial guess, and perform a single \SQPshort iteration per planning iteration \cite{diehl2005real}. This approach, called \RTIlong (\RTIshort), is exceptionally computationally efficient and has been proven successful in numerous applications, including motion planning for automated vehicles \cite{werling2012opt,ziegler2014trajectory,liniger2015optimization}. 

Ziegler et al.~highlighted a drawback of the \RTIshort approach in \cite{ziegler2014trajectory}, noticing a sensitivity to local minima in situations where the motion planning problem contains discrete decision making. The intuition behind the phenomenon is that at each iteration the optimizer searches for the best solution locally around the initial guess, and therefore it is unlikely to find a solution in a different part of the state space, that may contain the globally optimal solution. Liniger et al. \cite{liniger2015optimization} identify the same issue and mitigate it by means of a high level path planner based on dynamic programming, a solution that works well in the particular application, but not for the general case.

\subsection{Trajectory Roll-Out Methods}
\label{sec:relwork:rollout}
The trajectory roll-out concept represents a radically different approach to solving the motion planning problem. A large set of candidate trajectories are computed followed by a feasibility check and cost function evaluation that determines the optimal feasible candidate within the set \cite{howard2008state, Ferguson2011bossmp, werling2012_IJRRl}. 
Since the candidate trajectories are distributed across the drivable area, the method is less sensitive to local minima and thus performs well in situations where the motion planning problem contains discrete decisions. 

Computing the individual trajectories connecting the initial state to a sampled goal state means solving a nonlinear optimal control problem. This is computationally intractable for large numbers of samples. However, for particular classes of trajectory candidates, e.g., quintic and quartic polynomials, these problems have closed form solutions \cite{werling2012_IJRRl}. While such a solution makes trajectory roll-out computationally tractable, the resulting reference trajectories are not suitable for maneuvers at the limits of handling, due to the inaccurate representation of vehicle dynamics and limitations.

\subsection{Adaptive Control}
To account for local variations in physical capabilities of the vehicle, we draw from developments in the field of adaptive control. Predictive control under model uncertainty is a well-studied topic \cite{tanaskovic2014adaptive,cautiousGP_TCST_Hewing,kohler2019linear}. Such frameworks allow the system to dynamically re-plan safer and more cost efficient trajectories with time, as model parameters get updated.

These and related concepts are increasingly being adopted to account for traction variations in motion planning and control of automated vehicles. For example, Alsterda et al. \cite{alsterda2019contingency} present a novel approach in which a contingency plan, based on an alternative model parameter set, is computed in parallel to the nominal plan. As such, the method maintains a dynamically feasible contingency plan at all times, even in the event of, e.g., a rapid friction reduction. In this work, we instead leverage the notion of adaptive control by updating model parameters based on a \textit{predictive friction estimate}.



\subsection{Predictive Tire-Road Friction Estimation}
\label{sec:relwork:friction_est}
There exists a large body of research on the topic of tire road friction estimation from on-board vehicle measurements \cite{gustafsson1997slip,rajamani2012algorithms}. 
Such estimates have previously been shown to improve performance in several vehicle control applications, e.g., automatic emergency braking \cite{yi2002adaptiveAEB,alvarez2002adaptive} and path tracking control at deteriorated traction \cite{falcone2007slippery,gao2010predictive}. 
With the recent developments in computer vision and machine learning, methods for predictive friction estimation using camera images have emerged \cite{sohini2018predfrictionest,nolte2018predfrictionest,holzmann2006predfrictionest,omer2010predfrictionest,qian2016predfrictionest}. 
A \textit{predictive} friction estimate enables motion planning according to the traction conditions \textit{ahead} of the vehicle. 
This could also be obtained using a classical method such as, \cite{gustafsson1997slip,rajamani2012algorithms}, under the assumption that the traction conditions are unchanged for a short distance ahead of the vehicle.

To perform motion planning and control with respect to such predictive friction estimates, the planner must reliably handle situations where the friction estimate varies over the prediction horizon, or changes between subsequent planning iterations.


\subsection{Summary}
To summarize, optimization based methods (\RTIshort in particular) provide good performance in handling the trade-off between requirements 1), 2) and 3) in Section \ref{sec:prel}, but struggle with 4). Trajectory roll-out methods on the other hand, inherently handle 4), but have much lower performance in the trade-off between 1), 2) and 3). The method proposed in this work combines \RTIshort, trajectory roll-out and adaptive control to collectively address requirements 1) through 5). 
We leverage the notion of adaptive control by utilizing updated model information derived from a predictive friction estimate\footnote{We designed the method to receive a predictive friction estimate from a state of the art method, but the estimation itself is not included in the scope of this paper.}, to account for time-varying traction limitations in local motion planning, including discrete decisions.

\section{Problem Formulation}
\label{sec:pf}
Our goal in algorithm design is to develop a method that efficiently computes dynamically feasible reference trajectories at a high degree of traction utilization on varying road surfaces, based on a predictive friction estimate. Furthermore, the algorithm must handle cases where the predictive friction estimate changes over the course of the planning horizon and/or changes abruptly between planning iterations.
In this section we introduce the mathematical details of the considered motion planning problem and state the challenges it presents in further detail.

\subsection{Planning Model}
We derive a continuous time vehicle model with time varying parameters, shown in Appendix \ref{app:vehiclemodel}, and denote it $f_c(\cdot, \cdot)$. We let $t_c$ denote continuous time. For controller design, we obtain a discrete time version of this model as: 
\begin{align}\label{f_c_mod}
f(x_t, u_t) & = x_t + T_s f_c(x(tT_s), u(tT_s)), 
\end{align}
with time index $t$, time step $T_s$ and the state and input vectors 
\begin{align*}
    & x = [s, d, \Delta \psi, \dot{\psi}, v_{x}, v_{y}]^\top \in \mathbb{R}^n, \\
    & u = [F_{{y\mathrm{f}}}, F_{{x\mathrm{f}}}, F_{{x\mathrm{r}}}]^\top \in \mathbb{R}^m.
\end{align*}
\begin{remark}
We avoid separate notations for ``nominal" states from \eqref{f_c_mod} and the true states from the vehicle for the sake of notational simplicity in the remainder of the paper. This can be done as we do not opt for a robust control design approach, such as \cite{chisci2001systems, langson2004robust, bujarACC}.
\end{remark}

State variables $s$, $d$ and $\Delta \psi$ represent the progress along, the lateral deviation from and relative orientation to the lane center. Variables $\dot{\psi}$, $v_{x}$ and $v_{y}$ denote yaw angular velocity, longitudinal and lateral velocity of the vehicle. 
We employ the single track modelling technique \cite{rajamani2011vehicle}, in which the combined influence of two wheels on an axle is represented by a single tire in the middle of the axle, and select as control inputs: the lateral force on the front tire, $F_{{y\mathrm{f}}}$, the longitudinal force on the front tire, $F_{{x\mathrm{f}}}$, and the longitudinal force on the rear tire $F_{{x\mathrm{r}}}$. 
\begin{remark}
The primary reason for selecting tire force control inputs is that it allows us to explicitly model the locally varying traction limit as a time-varying constraint on the inputs. Furthermore, it contributes to the generality of the solution, since it can be instantiated on a wide range of vehicle types by setting a small number of easily measured parameters. A drawback compared to selecting steering angle as the control input for the lateral dimension is that it is non-trivial to represent actuator limitations such as maximum steering angle. However, for the speed range associated with the type of maneuvers we are targeting ($v_x > 5m/s$) this problem does not occur. In order to translate the resulting tire force commands into the vehicle's actual control inputs, a vehicle specific control interface is required. The control interface developed for our test vehicle is described in Appendix \ref{app:ctrl_interface}.
\end{remark}

In addition to the dynamics in the xy-plane, the model includes relations to determine the front and rear normal forces $F_{z\mathrm{f}}$ and $F_{z\mathrm{r}}$ from the state $x_t$. This, together with a predictive friction estimate, allows us to obtain a time-varying traction force boundary through~\eqref{eq:forcelimit}. Further details on the model are provided in Appendix \ref{app:vehiclemodel}.

\subsection{Optimal Control Problem}
In optimization based motion planning, a \CFTOClong (\CFTOCshort) is solved at regular time intervals, producing a new motion plan at each planning iteration as its solution. For the motion planning problem with traction limit variation we consider the \CFTOCshort

\begin{equation}
\begin{array}{ll}
\underset{u_{0|t}, \dots, u_{N-1|t}}{\mbox{min}} & J(\mathcal{T}_t)   \\\
~~~~~~\mbox{s.t.,} 	& x_{k+1|t} = f \left(x_{k|t},u_{k|t} \right), \\
			    & u_{k|t} \in \mathcal{U}_{k|t}, \\
			    & x_{k|t} \in \mathcal{X}_{k|t}, \\ 
			    & \forall k \in \{0,1,\dots,(N-1)\},\\
			    & x_{0|t} = x_t,~x_{N|t} \in \mathcal{X}_{k|t},  
\end{array}
\label{eq:cftoc}
\end{equation}
where the variables $\{u_{0|t},u_{1|t},\dots,u_{N-1|t}\}$ denote a predicted input sequence at time step $t$, which, when applied through the vehicle model $f(\cdot,\cdot)$, yields the corresponding predicted state sequence $\{x_{0|t},x_{1|t}, \dots, x_{N|t}\}$. 
The closed-loop state is reset again when $x_{t+1}$ is measured. We denote a predicted trajectory consisting of the associated predicted state and control sequences as 
\begin{align*}
\mathcal{T}_t = \{\{x_{k|t}\}_{k=0}^{N}, \{ u_{k|t}\}_{k=0}^{N-1}\}, 
\end{align*}
which is used in the selected quadratic cost function 
\begin{equation}
J(\mathcal{T}_t) = x_{N|t}^\top Q_N x_{N|t}  + \displaystyle\sum_{k=0}^{N-1} (x_{k|t}^\top Q x_{k|t} + u_{k|t}^\top R u_{k|t}),
\label{eq:cost}
\end{equation}
where $Q_N \succ 0$, $Q\succ 0$ and $R\succ 0$ are weight matrices that can be tuned by the designer. The choice here of a quadratic cost in \eqref{eq:cost} is done without loss of generality. Standard methods exists to make local quadratic approximations of non-quadratic cost functions, see Section \ref{sec:relwork:optbased}. 

In \eqref{eq:cftoc}, the state trajectory constraints $\mathcal{X}_{k|t}$, for all $k \in \{0,1,\dots,N\}$ represent collision free configurations of the vehicle with respect to the road geometry, static obstacles as well as positions and predicted positions of dynamic obstacles.

Notice that the input constraints $\mathcal{U}_{k|t}$ are \emph{time-varying} along the prediction horizon for $k \in \{0,1,\dots,(N-1)\}$. This is the mechanism we use to adapt the planned motion to local traction limitations.
The set $\mathcal{U}_{k|t}$ is computed at run-time as a function of the predicted friction estimate $\mu_{k|t}$ and normal forces $F_{z\mathrm{f}}$ and $F_{z\mathrm{r}}$. 
A further description on how the tire force limits are obtained is described in Appendix \ref{app:vehiclemodel}. Finally, we let $\mathcal{T}^\star_t = \{\{x^\star_{k|t}\}_{k=0}^{N}, \{ u^\star_{k|t}\}_{k=0}^{N-1}\}$ denote the optimal predicted trajectory. 

\subsection{Applying a State of the Art Approach: RTI}
\label{sec:RTI}

The Real Time Iteration approach \cite{diehl2005real,gros2016linear}, \RTIshort for short, is a state of the art approach for solving CFTOCs, that enables extraordinary computational efficiency \cite{Houska2011micro}.
In the context of motion planning in critical situations, computational efficiency is essential to achieving an acceptable trade-off between properties 1) 2) and 3) of Section \ref{sec:prel}. Hence, we select \RTIshort as a starting point for algorithm development. 

The central idea of the algorithm is to use the solution from the previous planning iteration to formulate a \QPshort approximation of the \CFTOCshort, which can be solved efficiently. The quadratic formulation is obtained by linearizing the dynamics and expressing the constraints as linear inequalities \cite{boyd2004convex}. 
At a single planning iteration starting at time $t$, a standard \RTIshort algorithm goes through the following steps:

First, the solution from the previous iteration $\mathcal{T}^\star_{t-1}$ is shifted one step forward in time, to compensate for the time passing between planning iterations. Thus, at time step $t$, an initial guess is obtained as,
\begin{align*}
\hat{\mathcal{T}}_t = \{\{\hat{x}_{k|t}\}_{k=0}^{N}, \{ \hat{u}_{k|t}\}_{k=0}^{N-1}\}, 
\end{align*}
where $\hat{x}_{k|t} = x^\star_{k+1|t-1}$ and $\hat{u}_{k|t} = u^\star_{k+1|t-1}$ for $k \in \{1,2,\dots,(N-1)\}$
\footnote{Forward shifting is not possible for the $N$-th predicted state. So the final control input is duplicated $\hat{u}_{N-1|t} = u^\star_{N-1|t-1}$ and the final state is obtained by integrating the dynamics forward $\hat{x}_{N|t} = f (x^\star_{N-1|t-1},u^\star_{N-1|t-1})$.}.
Next, the dynamics are linearized about $\hat{\mathcal{T}}_t$. The system model matrices are obtained as 
\begin{align*}
& A_{k|t} = \frac{\partial f}{\partial x}  \bigg|_{(\hat{x}_{k|t}, \hat{u}_{k|t})},~B_{k|t} = \frac{\partial f}{\partial u}  \bigg|_{(\hat{x}_{k|t}, \hat{u}_{k|t})},
\end{align*}
for all $k \in \{0,1,\dots,(N-1)\}$. State and input constraints are expressed as sets of linear inequalities:
\begin{align*}
& \mathcal{P}^\mathcal{X}_{k|t}= \{x: H^\mathcal{X}_{k|t} x \leq h^\mathcal{X}_{k|t}\} \subseteq \mathcal{X}_{k|t},\\  
& \mathcal{P}^\mathcal{U}_{k|t}= \{u: H^\mathcal{U}_{k|t} u \leq h^\mathcal{U}_{k|t}\} \subseteq \mathcal{U}_{k|t}, \end{align*}
with $H^\mathcal{X}_{k|t} \in \mathbb{R}^{s_x \times n}, h^\mathcal{X}_{k|t} \in \mathbb{R}^{s_x}$, for all $k \in \{0,1,\dots, N\}$, and $H^\mathcal{U}_{k|t} \in \mathbb{R}^{s_u \times m}$, $h^\mathcal{U}_{k|t} \in \mathbb{R}^{s_u}$ for all $k \in \{0,1,\dots, (N-1)\}$. 
The procedure we use to linearize the input constraints of our vehicle model is described in Appendix \ref{app:vehiclemodel}. 

The final \RTIshort style reformulation of \eqref{eq:cftoc} around an initial guess $\hat{\mathcal{T}}_t$ is shown in \eqref{eq:mpc_problem_rti}.
\begin{equation}\label{eq:mpc_problem_rti}
\begin{array}{ll}
\underset{\Delta u_{0|t},\dots, \Delta u_{N-1|t}}{\mbox{min}} &~~ \! \! \! \! J(\mathcal{T}_t),~\textnormal{with }\mathcal{T}_t = \{\{x_{k|t}\}_{k=0}^{N}, \{ u_{k|t}\}_{k=0}^{N-1}\},  \\\
~~~~~\mbox{s.t.,} 	& x_{k+1|t} \! = \! A_{k|t}(\Delta x_{k|t}) \! + \! B_{k|t} (\Delta u_{k|t}) \! + \! \hat{x}^\star_{k+1|t}, \\
& \textrm{(linearized about $\hat{\mathcal{T}}_{t}$)}\vspace{1mm}\\ 
& u_{k|t} \in \mathcal{P}_{k|t}^\mathcal{U},\\
& \forall k \in \{0,1,\dots,(N - 1)\},~\textrm{with }\\
& [\Delta x_{k|t}, \Delta u_{k|t}] = [x_{k|t} - \hat{x}'_{k|t}, u_{k|t}-\hat{u}'_{k|t}],\vspace{1mm} \\ 
& x_{k|t} \in \mathcal{P}_{k|t}^\mathcal{X},\\
& \forall k \in \{0,1,\dots,N\},\\
& x_{0|t} = x_t, 
\end{array}
\end{equation}
When a solution of \eqref{eq:mpc_problem_rti} for time step $t$ is obtained, the control 
\begin{align*}
    u_t = u^\star_{0|t},
\end{align*}
is applied to the vehicle. At the next time step, $(t+1)$, problem \eqref{eq:mpc_problem_rti} is solved again. 

\subsection{Challenges in Solving the RTI Problem \eqref{eq:mpc_problem_rti}}\label{sec:rti_lim}
The \RTIshort algorithm builds on the assumption that $\mathcal{T}^\star_{t-1}$ is a feasible and near-optimal initial guess for time step $t$. If this is not the case, performance of the algorithm deteriorates, leading to three essential drawbacks with respect to the problem formulation of this paper.  

\begin{enumerate}[(i)]
\item \textit{Infeasibility w.r.t. $\mathcal{P}_{k|t}^\mathcal{U}$:} In the presence of time-varying input constraints, there is no guarantee that at time step $t$ the input sequence from, $\mathcal{T}^\star_{t-1}$ is feasible with respect to $\mathcal{P}_{k|t}^\mathcal{U}$, i.e., the condition  $u^\star_{k|t-1} \in \mathcal{P}_{k|t}^\mathcal{U}$ may or may not hold for all $k \in \{1,2,\dots,N\}$. \label{probitem:infeasU}
\item \textit{Infeasibility w.r.t. $\mathcal{P}_{k|t}^\mathcal{X}$:} Irrespective of \ref{probitem:infeasU}, in a dynamically changing environment with sensor limitations and unpredictably moving human actors, there is no guarantee that at time step $t$, the state sequence of $\mathcal{T}^\star_{t-1}$ will be collision free, i.e., the condition $x^\star_{k|t-1} \in \mathcal{P}_{k|t}^\mathcal{X}$ may or may not hold for all $k \in \{1,2,\dots,N\}$. \label{probitem:infeasX}
\item \textit{Sensitivity to Local Minima:} Due to the fact that the search for $\mathcal{T}^\star_{t}$ is only performed locally around $\mathcal{T}^\star_{t-1}$, \RTIshort is prone to suboptimal decision making where the planning problem includes discrete decision making. In realistic traffic, new discrete decisions may appear dynamically \cite{ziegler2014trajectory}. \label{probitem:localminima}
\end{enumerate}
Problem \ref{probitem:infeasU} is unique to our problem formulation, and stems from the addition of traction adaptation through time-varying input constraints $\mathcal{P}_{k|t}^\mathcal{U}$ introduced in \eqref{eq:cftoc}. A naive solution is to project $u^\star_{k|t-1}$ onto $\mathcal{P}_{k|t}^\mathcal{U}$ for all $k \in \{1,2,\dots,(N-1)\}$ and recompute $\hat{\mathcal{T}}_t$ from the dynamics model (this is the first step of our proposed algorithm, see Section \ref{sec:ens_feas_previous}). However, this is not a complete solution as it, for non-trivial cases, merely transforms problem \ref{probitem:infeasU} into problem \ref{probitem:infeasX}.  
Problem \ref{probitem:infeasX} has been identified in previous works and is unavoidable in a dynamic traffic environment. Computational feasibility can be obtained by introducing slack variables, such that state constraint violations are allowed, but with a very high cost \cite{kerrigan2000soft,liniger2015optimization}. Commonly however, this approach merely transforms Problem \ref{probitem:infeasX} into Problem \ref{probitem:localminima}.
Problem \ref{probitem:localminima} in turn may be handled for specific applications by a higher level decision making layer \cite{liniger2015optimization}. In critical situations however, the optimal decision, for example to swerve left or right of a suddenly appearing obstacle, is highly dependent on, e.g., the geometry of the traffic scene, the vehicle's state, dynamics and dynamic limitations. This information is typically only available at the local motion planning level and therefore, an informed decision can only be made there. 
In such critical decision making, insufficient scene understanding may lead to suboptimal decisions regarding the selected maneuvers, which in turn reduces capacity to avoid accidents. 
Trajectory roll-out methods, in contrast to optimization based methods, inherently handle such decisions by construction at the local planning level, as described in Section \ref{sec:relwork:rollout}.   

In the next section, we present a novel variant of \RTIshort, which we refer to as \SAARTIlong (\SAARTIshort), to jointly address \ref{probitem:infeasU}, \ref{probitem:infeasX} and \ref{probitem:localminima}.

\section{Sampling Augmented Adaptive RTI}
\label{sec:method}

The \SAARTIshort algorithm decomposes the solving of \eqref{eq:cftoc} into four steps using the same planning horizon $N$ and time discretization step $T_s$. Furthermore, steps C and D use the same cost function \eqref{eq:cost}. 
The steps are:
\begin{enumerate}[A.]
    \item Ensuring input feasibility of $\mathcal{T}^\star_{t-1}$, 
    \item Generating additional initial guess candidates through sampling, 
    \item Selection of initial guess trajectory,
    \item Constraint adaptive trajectory optimization. 
\end{enumerate}
Together, steps A-D remedy the limitations \ref{probitem:infeasU}, \ref{probitem:infeasX} and \ref{probitem:localminima} of \RTIshort introduced in Section~\ref{sec:rti_lim}. In the following sub-sections we explain each step in further detail.

\subsection{Ensuring Input Feasibility of $\mathcal{T}^\star_{t-1}$}
\label{sec:ens_feas_previous}

Due to the time-varying input constraints, situations when $u^\star_{k|t-1} \notin \mathcal{P}_{k|t}^\mathcal{U}$ for $k \in \{1,2,\dots,(N-1)\}$ may occur. To resolve this, we replace the shifting procedure in conventional \RTIshort as:
\begin{align}
    \tilde{u}_{k|t} = \mathrm{Proj}_{\mathcal{P}_{k|t}^\mathcal{U}}(u^\star_{k+1|t-1}),~\forall k \in \{0,1,\dots, (N-1) \},
\end{align}
where $\mathrm{Proj}_\Omega(\cdot)$ denotes Euclidean projection operation to the set $\Omega$. 
Thus, we modify individual inputs if needed, such that $\tilde{u}_{k|t} \in \mathcal{P}_{k|t}^\mathcal{U}$ for $k \in \{0,1,\dots,(N-1)\}$. Then, under the new input sequence, we propagate the system forward from $\tilde{x}_{0|t} = x^\star_{1|t-1}$ to determine the state sequence $\tilde{x}_{k+1|t} = f (\tilde{x}_{k|t},\tilde{u}_{k|t})$, $k \in \{1,2, \dots, (N-1)\}$. 
Thereby, we obtain an initial guess 
\begin{equation}
    \tilde{\mathcal{T}}_t = \{\{\tilde{x}_{k|t}\}_{k=0}^{N}, \{ \tilde{u}_{k|t}\}_{k=0}^{N-1}\},
    \label{eq:T_tilde}
\end{equation}
that satisfies dynamics and time-varying input constraints, resolving limitation \ref{probitem:infeasU} of Section \ref{sec:pf}. 

We acknowledge that the procedure may alter the state sequence such that state constraints are violated, i.e., $\tilde{x}_{k|t} \notin \mathcal{P}_{k|t}^\mathcal{X}$, for some $ k \in \{0,1,\dots,N\}$. However, as stated in Problem \ref{probitem:infeasX} in Section \ref{sec:rti_lim}, this problem has to be addressed regardless of this addition, due to the dynamic environment. We address the problem in the next step of the algorithm by employing the trajectory roll-out concept \cite{howard2008state}.

\subsection{Generating Additional Initial Guess Candidates Through Sampling}
\label{sec:feasibleinitguess}

To handle the case when the initial guess \eqref{eq:T_tilde} violates the state constraints, $\tilde{x}_{k|t} \notin \mathcal{P}_{k|t}^\mathcal{X}$ for any $k \in \{1,2,\dots,N\}$, i.e., Problem \ref{probitem:infeasX}, and to avoid highly suboptimal local minima, i.e., Problem \ref{probitem:localminima}, we employ a further developed version of the sampling augmentation concept that was first introduced in  \cite{svensson2019adaptive}.  

The procedure starts by selecting $N_s$ reference states $x^{(i)}_{\textnormal{ref}}$, $i \in \{1,2, \dots, N_s\}$. 
Reference values for $d$ and $v_x$ are uniformly distributed within admissible values, the $s$-reference selected as the current $s$ of the vehicle, and remaining state references are set to zero.
For each reference state, we integrate the dynamics forward, 
\begin{align}
    \hat{x}^{(i)}_{k+1|t} = f (\hat{x}^{(i)}_{k|t},\hat{u}^{(i)}_{k|t}),~\forall k \in \{0,1, \dots, (N-1)\},
\end{align}
from the initial state $\hat{x}^{(i)}_{0|t} = x_t$, under a Linear Quadratic Tracking (LQT) controller, for all $i \in \{1,2,\dots, N_s\}$, that minimizes deviation from the reference state. At each time step $k$, the optimal control input is computed as 
\begin{align}
    \hat{u}^{(i)}_{k|t} = -K_t(\hat{x}_{k|t}^{(i)} - x^{(i)}_\textnormal{ref}),
\end{align}
where $K_t$ is the optimal gain matrix, recomputed at each planning iteration by linearizing the dynamics around the the initial state $x_t$ and solving the Discrete-time Algebraic Riccati Equation. 
The cost function weights are selected such that behavior is primarily dominated by errors in $d$ and $v_x$. Errors for example in $s$, have small influence.

To ensure $\hat{u}^{(i)}_{k|t} \in \mathcal{P}_{k|t}^\mathcal{U}$ for $k \in \{1,2,\dots,N\}$ we apply the procedure described in Section \ref{sec:ens_feas_previous} to project all control inputs onto the feasible set before propagating the state.
The $N_s$ candidate trajectories 
\begin{align*}
\hat{\mathcal{T}}^{(i)}_t = \{\{\hat{x}^{(i)}_{k|t}\}_{k=0}^{N}, \{ \hat{u}^{(i)}_{k|t}\}_{k=0}^{N-1}\}, 
\end{align*}
for all $i \in \{1,2,\dots,N_s\}$ are stored in the trajectory set 
\begin{align}\label{eq:s_hatt}
\hat{\mathcal{S}}_t = \bigcup_{i=1}^{N_s} \hat{\mathcal{T}}^{(i)}_t.
\end{align}
Notice that following the computation of $K_t$, each trajectory is computed independently, enabling efficient GPU acceleration.

\subsection{Selection of Initial Guess Trajectory}
\label{sec:selectinitguess}
To determine the best available initial guess, each candidate trajectory in the set $\hat{\mathcal{S}}_t \cup \tilde{\mathcal{T}}_{t}$ is evaluated based on the shared cost function $J(\cdot)$ in \eqref{eq:cost}. 
The lowest cost candidate, i.e.,
\begin{align}\label{eq:thattilde}
\hat{\mathcal{T}}'_{t} = \arg \min \limits_{{\hat{\mathcal{T}}_t \in \hat{\mathcal{S}}_t \cup \tilde{\mathcal{T}}_t} } (J(\hat{\mathcal{T}}_{t}))
\end{align}
is selected for subsequent optimization. Trajectory candidates that violate state constraints, i.e., $\hat{x}_{k|t} \notin \mathcal{P}_{k|t}^\mathcal{X}$ for any $k \in \{1,2,\dots,N\}$ are disregarded in the selection. 

In case $\tilde{\mathcal{T}}_t$ is colliding or is in a highly suboptimal local minima, an alternative feasible and near-optimal initial guess can be obtained from $\hat{\mathcal{S}}_t$. Thus, we exploit the inherent redundancy in the trajectory roll-out concept \cite{werling2012_IJRRl}, to resolve Problems \ref{probitem:infeasX} and \ref{probitem:localminima} of Section \ref{sec:pf}. This property of the algorithm is experimentally verified in Section \ref{sec:results:discrete_decisions}. 

\begin{remark}
In cases when changes to the time-varying state and input constraints are small between planning iterations, it is highly probable that $\hat{\mathcal{T}}'_{t} = \tilde{\mathcal{T}}_{t}$. In this case, algorithm behavior is identical to that of \RTIshort. This is further elaborated in Section \ref{sec:results:discrete_decisions}. 
\end{remark}

\subsection{Constraint Adaptive Trajectory Optimization}
\label{sec:trajopt}
With the initial guess $\hat{\mathcal{T}}'_t$ obtained from \eqref{eq:thattilde}, the final step of the \SAARTIshort algorithm is to solve the \QPshort 

\begin{equation}\label{eq:mpc_problem}
\begin{array}{ll}
\underset{\Delta u_{0|t},\dots, \Delta u_{N-1|t}}{\mbox{min}} &~~\! \! \! \! J(\mathcal{T}_t),~\textnormal{with } \mathcal{T}_t = \{\{x_{k|t}\}_{k=0}^{N}, \{ u_{k|t}\}_{k=0}^{N-1}\},  \\\
~~~~\mbox{s.t.,} 	& x_{k+1|t} \! = \! A_{k|t}(\Delta x_{k|t}) \! + \! B_{k|t} (\Delta u_{k|t}) \! + \! \hat{x}'_{k+1|t}, \\
& \textrm{(linearized about $\hat{\mathcal{T}}'_{t}$)},\vspace{1mm}\\
& u_{k|t} \in \mathcal{P}_{k|t}^\mathcal{U},\\
& \forall k \in \{0,1,\dots,(N - 1)\},~\textrm{with }\\
& [\Delta x_{k|t}, \Delta u_{k|t}] = [x_{k|t} - \hat{x}'_{k|t}, u_{k|t}-\hat{u}'_{k|t}],\vspace{1mm} \\ 
& x_{k|t} \in \mathcal{P}_{k|t}^\mathcal{X}, \\
& \forall k \in \{0,1,\dots,N\},\\
& x_{0|t} = x_t.
\end{array}
\end{equation}
In contrast to \eqref{eq:mpc_problem_rti}, the \QPshort is initialized at $\hat{\mathcal{T}}'_t$, i.e., with the system model matrices given as
\begin{align*}
A_{k|t} = \frac{\partial f}{\partial x}  \bigg|_{(\hat{x}'_{k|t}, \hat{u}'_{k|t})},~B_{k|t} = \frac{\partial f}{\partial u}  \bigg|_{(\hat{x}'_{k|t}, \hat{u}'_{k|t})},
\end{align*}
for all $k \in \{0,1,\dots,(N-1)\}$. After solving \eqref{eq:mpc_problem} at any time step $t$, we apply the closed-loop command
\begin{align}\label{cl_control}
    u_t = u^\star_{0|t}
\end{align}
to the vehicle system through the low level control interface described in Appendix \ref{app:ctrl_interface}. At time step $(t+1)$, \eqref{eq:mpc_problem} is re-solved.

\subsection{Summary and Implications for Traction Adaptation}
We summarize \SAARTIshort in Algorithm~\ref{alg:saa-rti}. Here, $M_t$ represents local map features, e.g., lane boundaries and static obstacles, and $O_t$ denotes dynamic obstacles at time step $t$. Note that we assume availability of a predictive tire-road friction estimates $\mu_{k|t}$ for $k \in \{0,1,\dots, (N-1)\}$ at any time step $t$.
\begin{algorithm}[h] 
    \caption{The SAA-RTI Algorithm}
    \begin{algorithmic}[1] \label{alg:saa-rti}
    \STATE \textbf{Initialization:} $\mathcal{T}^\star_{t-1} = \emptyset$
    \WHILE {$t \geq 0$}
    \STATE \textbf{Inputs:} $x_t$, $\mathcal{T}^\star_{t-1}$, $M_t$, $O_t$, $\mu_{k|t},\forall k \in \{0,1,\dots,N\}$
    \IF{$\mathcal{T}^\star_{t-1} \neq \emptyset$}
    \STATE  $\tilde{\mathcal{T}}_{t}$ $\leftarrow$ \text{ShiftAndEnsureFeasibility}($\mathcal{T}^\star_{t-1}$, $\mu_{k|t},~\forall k$)
    \ELSE 
    \STATE $\tilde{\mathcal{T}}_{t} = \emptyset$
    \ENDIF
    \STATE $\hat{\mathcal{S}}_t$ $\leftarrow$ \text{FeasibleTrajRollout}($x_t, M_t, \mu_{k|t},~\forall k$) from \eqref{eq:s_hatt}
    \STATE $\hat{\mathcal{T}}'_{t} = \arg \min \limits_{{\hat{\mathcal{T}}_t \in \hat{\mathcal{S}}_t \cup \tilde{\mathcal{T}}_t} } (J(\hat{\mathcal{T}}_{t}))$ from \eqref{eq:thattilde}
    \STATE $A_{k|t}, B_{k|t}$ $\leftarrow$ \text{LinearizeModel}($\hat{\mathcal{T}}'_t,~\forall k$)
    \STATE $\mathcal{P}^\mathcal{U}_{k|t}$ $\leftarrow$ \text{InputConstraints}($\hat{\mathcal{T}}'_t, \mu_{k|t},~\forall k$)
    \STATE $\mathcal{P}^\mathcal{X}_{k|t}$ $\leftarrow$ \text{StateConstraints}($\hat{\mathcal{T}}'_t$, $O_t$, $M_t,~\forall k$) 
    \STATE $\mathcal{T}^\star_{t} =  \{\{x^{\star }_{k|t}\}_{k=0}^{N},\{u^{\star }_{k|t}\}_{k=0}^{N-1}\}$ $\leftarrow$ Solve \eqref{eq:mpc_problem}
    \STATE Apply $u^\star_{t|t}$ to the vehicle
    \ENDWHILE
    \end{algorithmic}
\end{algorithm}

\begin{figure}[t]
    \captionsetup[subfigure]{}
    \centering
    \subfloat[Position and orientation of planned motion]{%
        \includegraphics[width=0.5\columnwidth]{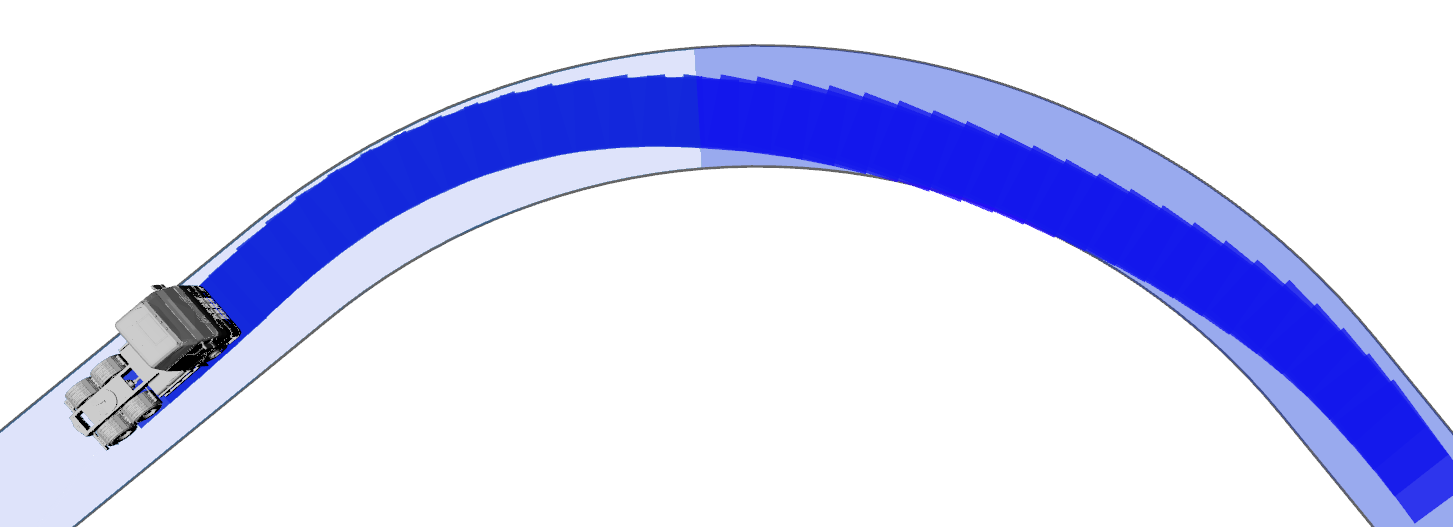}
        \label{fig:mu_drop_oh}
    }\\
    \subfloat[Velocity and front tire force of planned motion]{%
        \includegraphics[width=0.6\columnwidth]{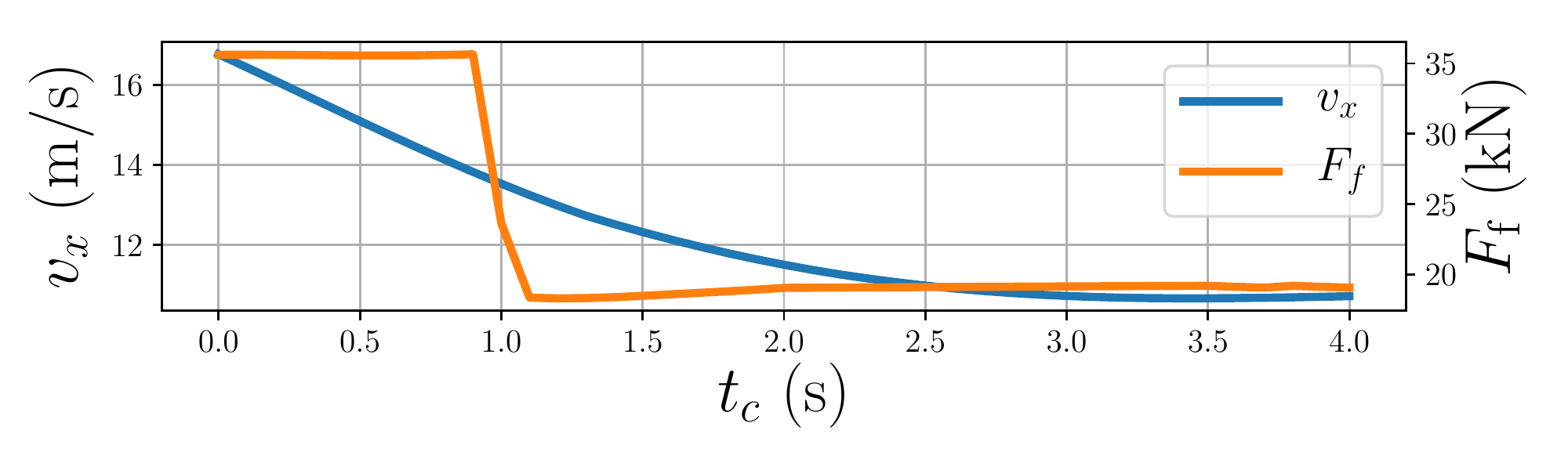}
        \label{fig:mu_drop_v_F}
    }
    \caption{Example of planned behavior when the friction coefficient varies over the prediction horizon. The proposed approach plans dynamically feasible motions at the (varying) limit of tire adhesion.} 
    \label{fig:mu_drop}
\end{figure}

\begin{figure}[t!]
    \captionsetup[subfigure]{}
    \centering
    \subfloat[Position and orientation of initial guess candidates and optimized solution.]{%
        \includegraphics[width=0.6\columnwidth]{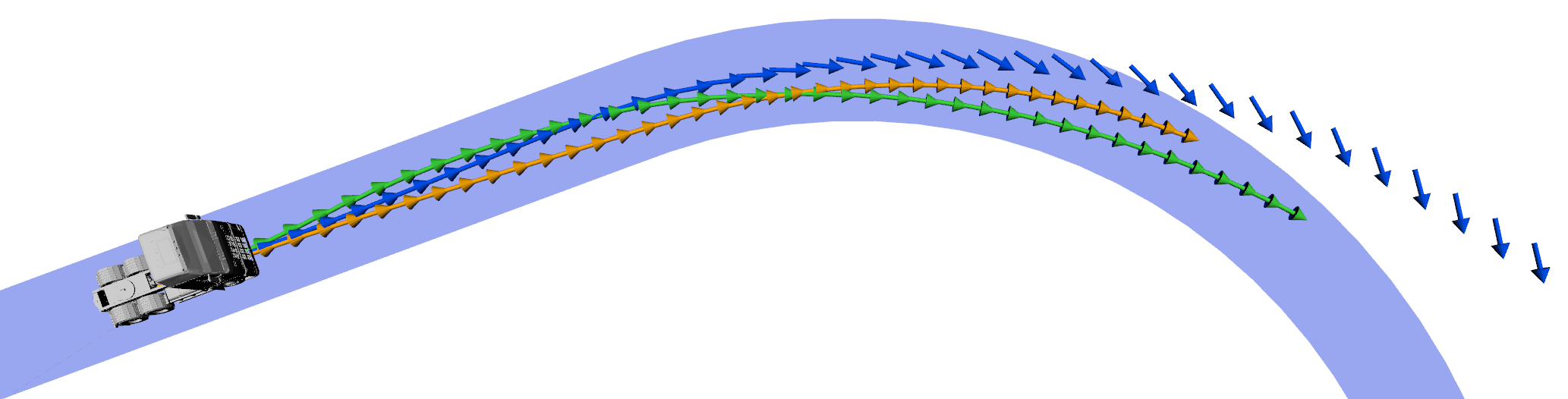}
        \label{fig:guesses_XYpsi}
    }\\
    \subfloat[Velocity state of initial guess candidates and optimized solution.]{%
        \includegraphics[width=0.6\columnwidth]{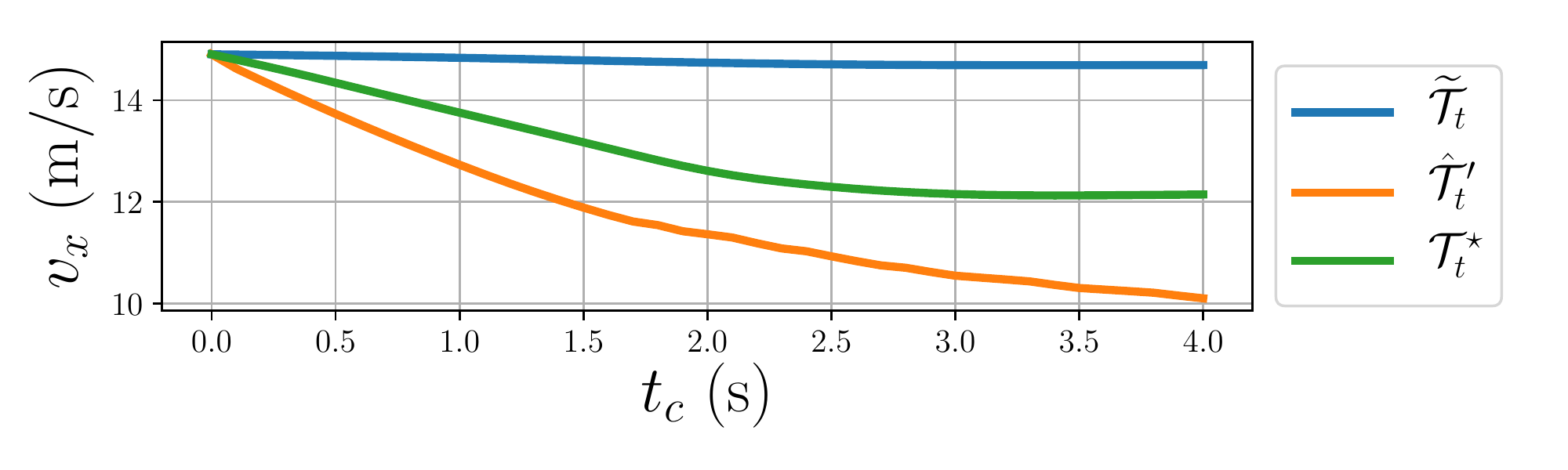}
        \label{fig:guesses_vx}
    }
    \caption{Initial guess candidates when estimated traction changes abruptly between planning iterations. Sampling augmentation mitigates the problems of using the solution from the previous iteration as initial guess.}
    \label{fig:guesses}
\end{figure}

The proposed approach of modelling locally varying tire force limits as time-varying input constraints allows the vehicle to plan motions that optimally utilizes traction throughout the prediction horizon. 
Fig.~\ref{fig:mu_drop}, shows an example where $\mu$ drops from 0.8 to 0.3 in the middle of the horizon. Recall that $t_c$ denotes our continuous time variable. The planned motion exploits the initial high traction to reduce velocity, Fig.~\ref{fig:mu_drop_v_F}, and position the vehicle such that the curvature of the trajectory is reduced over the low traction part of the corner, Fig.~\ref{fig:mu_drop_v_F}. This is done in a coordinated manner such that planned tire forces, shown for the front tire in Fig.~\ref{fig:mu_drop_v_F}, are at the (varying) limit throughout the planned maneuver, i.e., for the front tire, $\sqrt{(F_{x\text{f}})_{k|t}^2 + (F_{y\text{f}})_{k|t}^2} = \mu_{k|t} (F_{z\text{f}})_{k|t}$.

The SAA-RTI algorithm ensures efficient and reliable solving of the traction adaptive motion planning problem. The sampling augmentation procedure ensures that the optimization procedure is always provided a feasible and near-optimal initial guess, even when the predictive friction estimate changes dramatically between planning iterations.
Fig.~\ref{fig:guesses} shows an example where the estimate of $\mu$ has dropped from 0.8 to 0.3 between the previous and the current planning iteration. The figure shows position and orientation, Fig.~\ref{fig:guesses_XYpsi}, and velocity states, Fig.~\ref{fig:guesses_vx} of the forward shifted and input feasibility adjusted solution from the previous iteration $\tilde{\mathcal{T}}_{t}$ (blue), the lowest cost SAA-RTI initial guess candidate $\hat{\mathcal{T}}'_{t}$ (orange) and the final optimized trajectory $\mathcal{T}^\star_t$ (green). It is evident from Fig.~\ref{fig:guesses_XYpsi}, that even when adjusted for input feasibility, Problem (i), $\tilde{\mathcal{T}}_{t}$ is still not feasible with respect to state constraints, Problem (ii). The addition of sampling augmentation reliably alleviates this problem by providing an abundance of additional feasible candidates and selecting the best available option $\hat{\mathcal{T}}'_{t}$. In addition, since the candidates are spread across the drivable area, sampling augmentation greatly reduces sensitivity to local minima in the presence of discrete decisions, Problem (iii), which we will revisit in Section.  \ref{sec:results:discrete_decisions}.

\section{Experimental Evaluation}
\label{sec:exp_setup}

We evaluate the performance of traction adaptive motion planning and control, realized by Algorithm~\ref{alg:saa-rti}, in terms of its capacity to avoid accidents in a set of critical scenarios. We compare the traction adaptive algorithm (abbreviated TA) with a baseline non-adaptive \RTIshort scheme with static tire force constraints (abbreviated NA). 
The TA scheme uses time-varying tire force constraints computed as a function of predicted normal forces and a predictive tire-road friction estimate $\mu_{k|t}=\mu_{\textnormal{est}}(s)$ for all $k \in \{0,1,\dots,(N-1)\}$ as per equations \eqref{eq:pitchdynamics} and \eqref{eq:tireforce_bounds} in Appendix \ref{app:vehiclemodel} with traction utilization factor $\lambda = 0.90$ for all scenarios. The NA scheme on the other hand uses static tire force constraints associated with a static friction estimate $\mu_{k|t} = \mu_{\textnormal{sta}}$ for all $k \in \{0,1,\dots,(N-1)\}$ and $a_x = 0$. 
All evaluated configurations use $N=40$ and $T_s=0.1$s, i.e., a planning horizon length of 4s.

The critical scenarios selected for evaluation are:
\begin{enumerate}[(i)]
    \item \textit{Turn at Low Local Traction} 
    \item \textit{Collision Avoidance at Low Local Traction} 
    \item \textit{Collision Avoidance at High Local Traction} 
    \item \textit{Collision Avoidance with Discrete Decisions}
\end{enumerate}

Scenarios (i), (ii) and (iii) are selected to test the accident avoidance performance impact of traction adaptation, whereas Scenario (iv) highlights the impact of avoiding local minima in the planning problem.

For the evaluation in this paper, we assume that an accurate predictive friction estimate is available, i.e., $\mu_{\textnormal{est}}(s) = \mu_{\textnormal{gt}}(s)$, where $\mu_{\textnormal{gt}}(s)$ denotes the ground truth tire-road friction, which was obtained for the experiments by performing friction measurements ahead of time and storing the values in a map of the track. 
The tests were performed on flat and level road surfaces and therefore, evaluation of the effects of road inclination is not included in this work, i.e., $\theta = \phi = 0$ rad. in  \eqref{eq:dynamics_cont}, Appendix \ref{app:vehiclemodel}. 
In the collision avoidance scenarios, all suddenly appearing obstacles remain at a static position after appearing. 
\begin{remark}
This delimitation is \textit{not} done because of limitations in the evaluated algorithms with respect to dynamic obstacles. For both the TA and NA schemes, state constraints $\mathcal{P}_{k|t}^\mathcal{X}$ are set individually for each $k \in \{0,1,\dots, N\}$. Hence, including avoidance of for example a confidence interval of predicted positions of the obstacle, is trivial from the planning perspective. Rather, the delimitation was made to ensure consistency between runs, and to avoid introducing additional uncertainty from the prediction functionality. Since the traction adaptive and the non-adaptive schemes are equivalent in this regard, we are confident that the conclusions we make for static obstacles will also hold for dynamic obstacles.
\end{remark}

In the following sub-sections, 
\ref{sec:results:turn_with_reduced_mu}, 
\ref{sec:results:coll_avoid_reduced_mu},  \ref{sec:results:coll_avoid_high_mu} and \ref{sec:results:discrete_decisions}, we present and discuss results from each of the scenarios. 
The experimental setup is detailed in Appendix \ref{app:exp_setup}.

\subsection{Scenario (i): Turn at Low Local Traction}
\label{sec:results:turn_with_reduced_mu}
\begin{figure}[!]
\begin{minipage}[c]{0.5\textwidth}
\captionsetup[subfigure]{}
\centering
    \subfloat[NA, $t_c=0$s]{%
        \includegraphics[width=0.45\columnwidth]{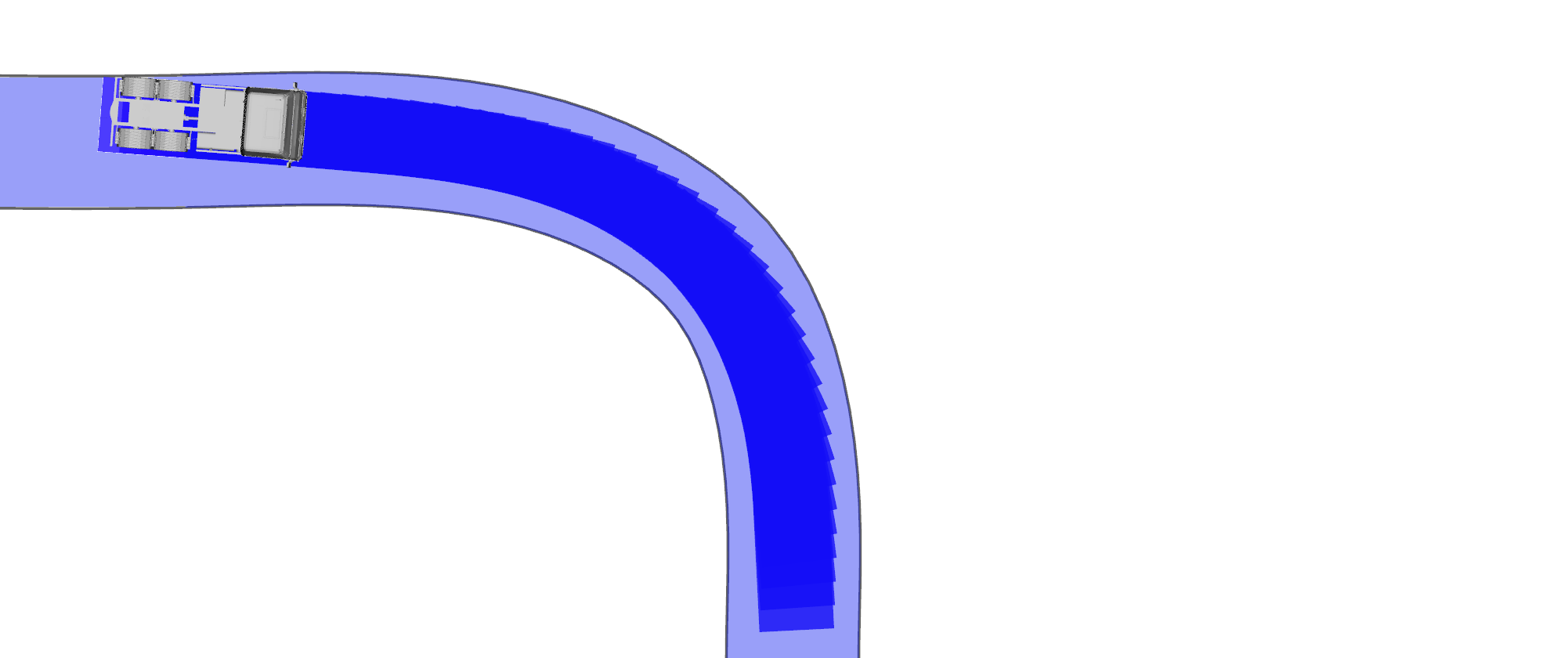}
        \label{fig:reducedmuturn:NA0}
    }
    \subfloat[TA, $t_c=0$s]{%
        \includegraphics[width=0.45\columnwidth]{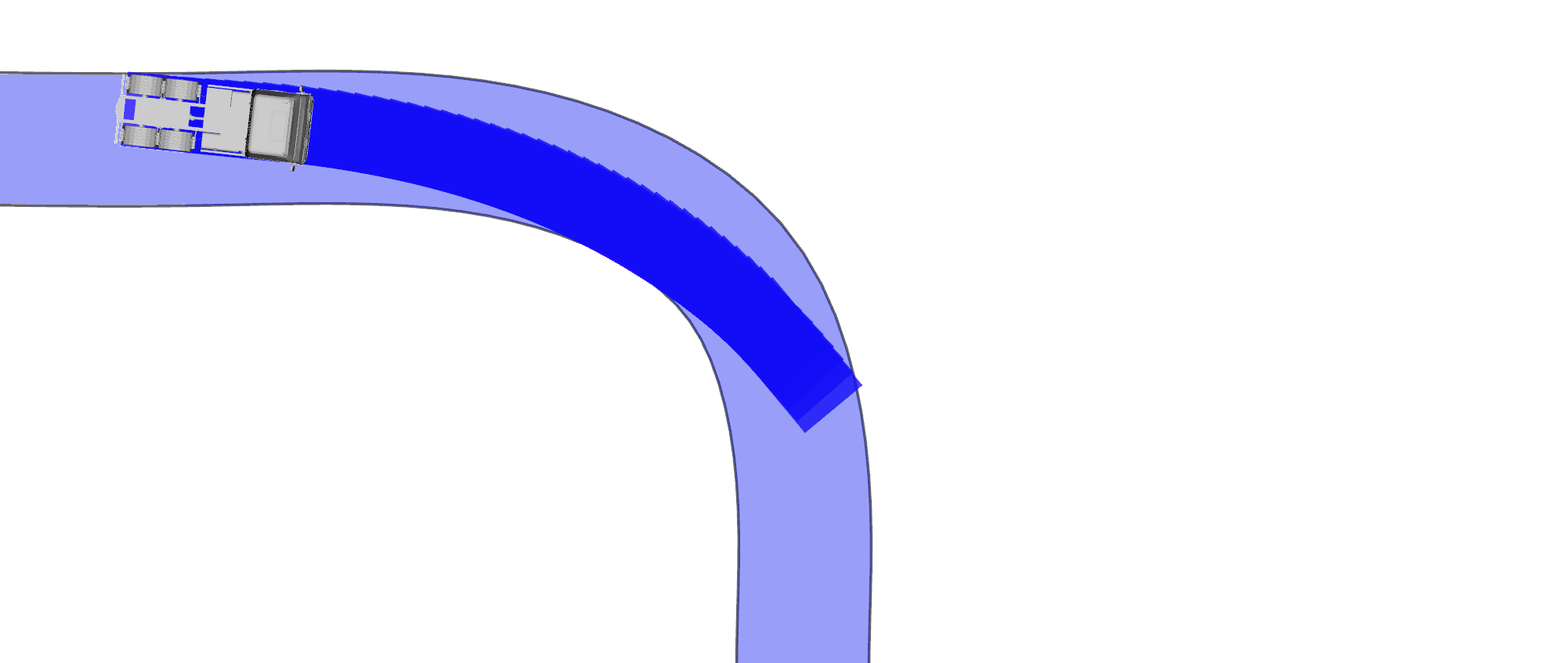}
        \label{fig:reducedmuturn:TA0}
    }\\ 
    \subfloat[NA, $t_c=2.5$s \mbox{(under-steering)}]{%
        \includegraphics[width=0.45\columnwidth]{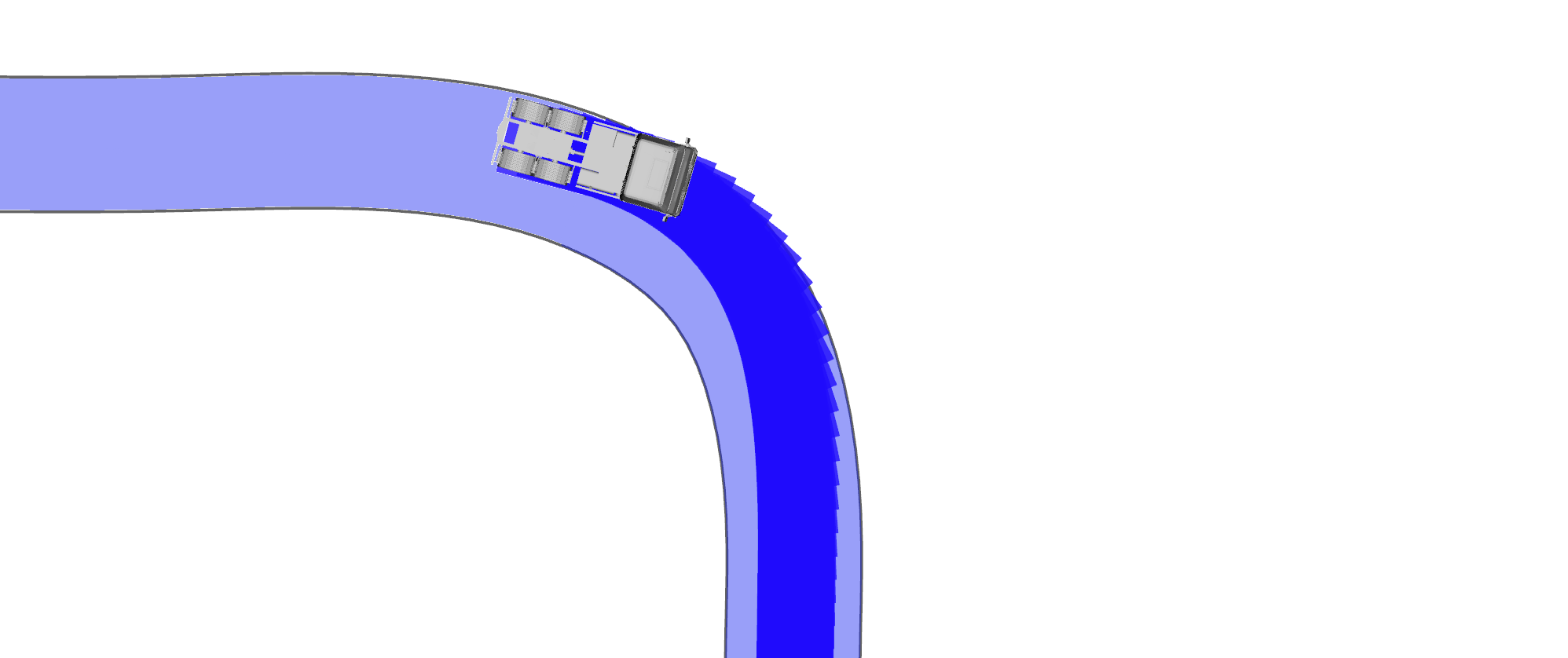}
        \label{fig:reducedmuturn:NA1}
    }
    \subfloat[TA, $t_c=2.5$s]{%
        \includegraphics[width=0.45\columnwidth]{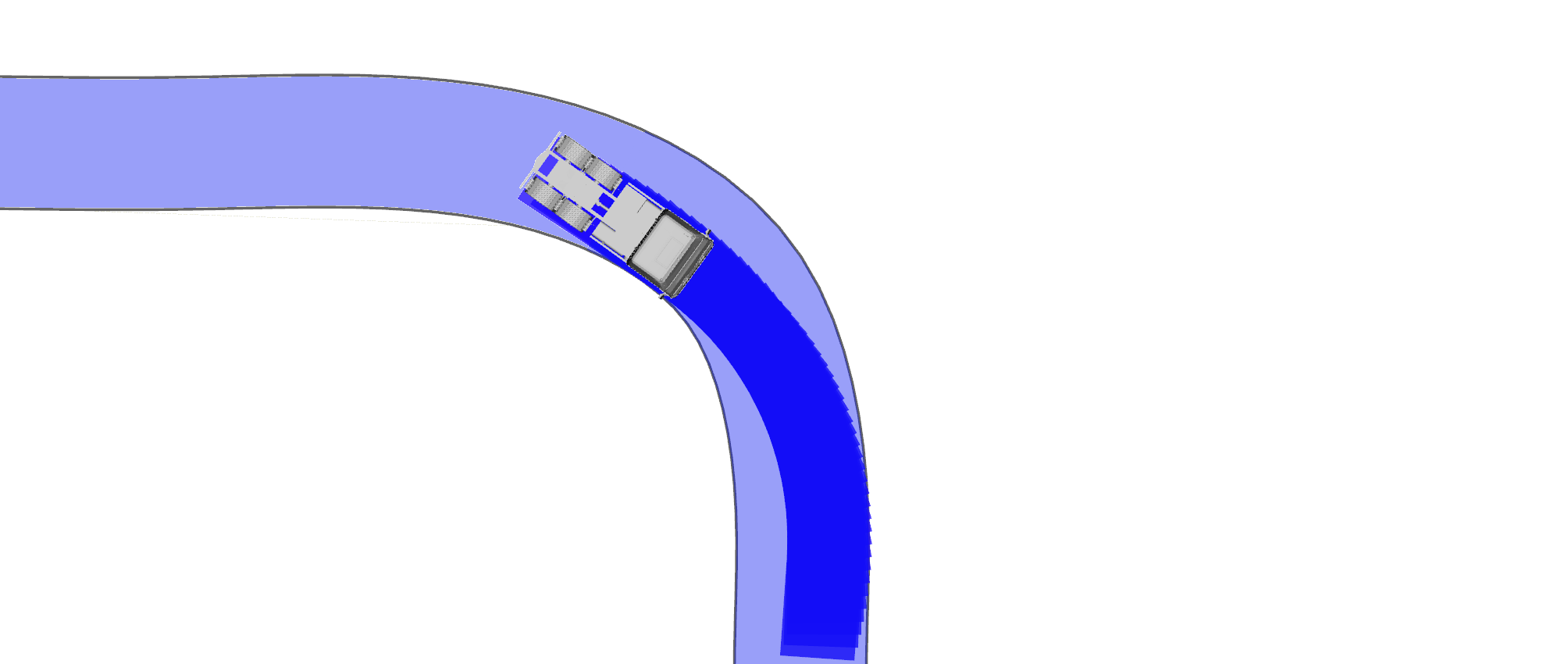}
        \label{fig:reducedmuturn:TA1}
    }\\
    \subfloat[NA, $t_c=3.5$s \mbox{(exiting lane)}]{%
        \includegraphics[width=0.45\columnwidth]{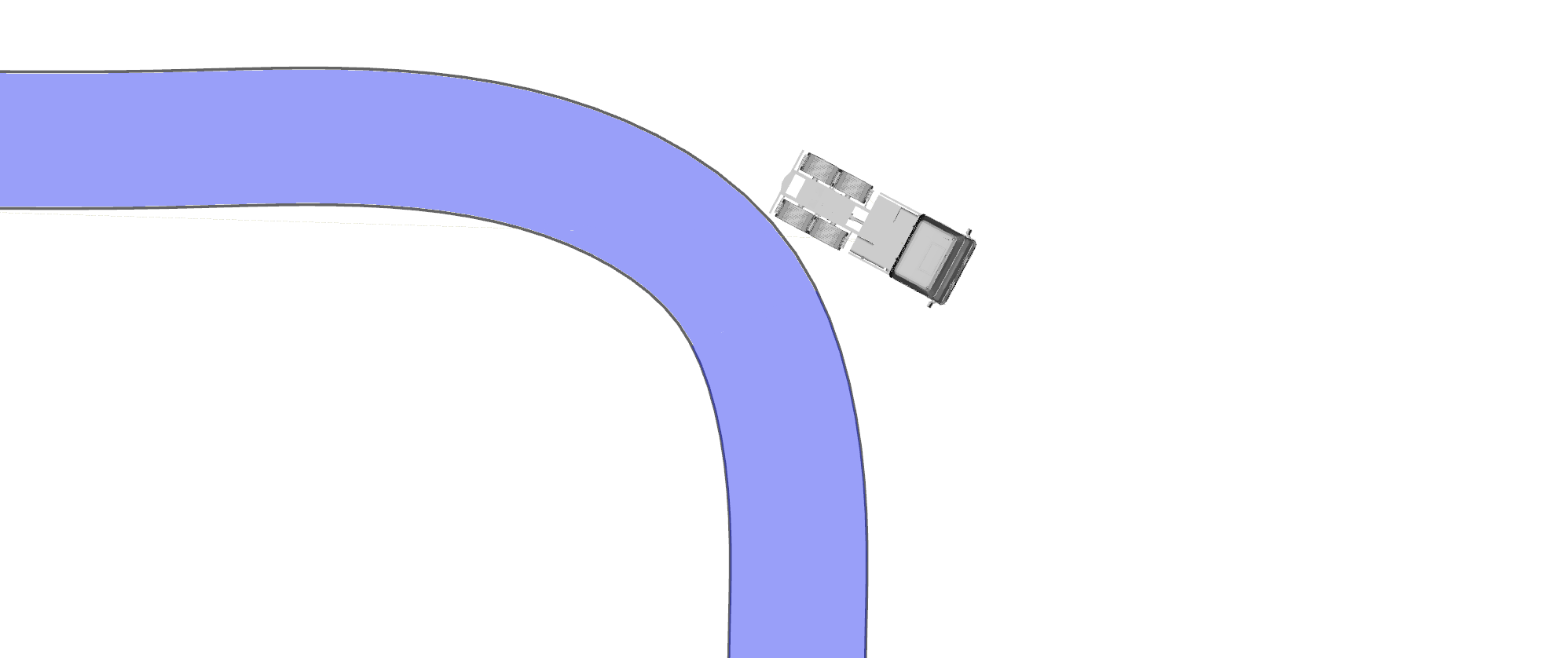}
        \label{fig:reducedmuturn:NA2}
    }
    \subfloat[TA, $t_c=4$s]{%
        \includegraphics[width=0.45\columnwidth]{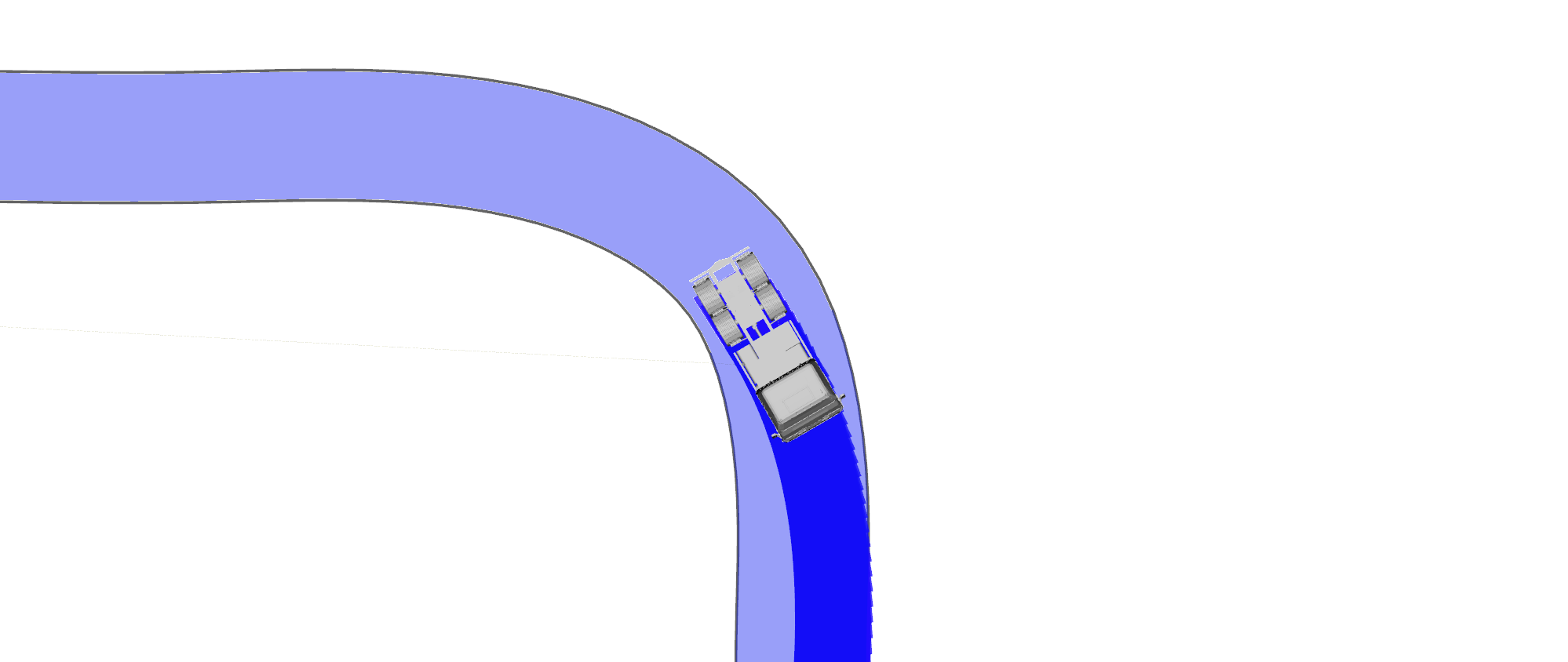}
        \label{fig:reducedmuturn:TA2}
    }\\
    \subfloat[NA, $t_c=7$s \mbox{(accident)}]{%
        \includegraphics[width=0.45\columnwidth]{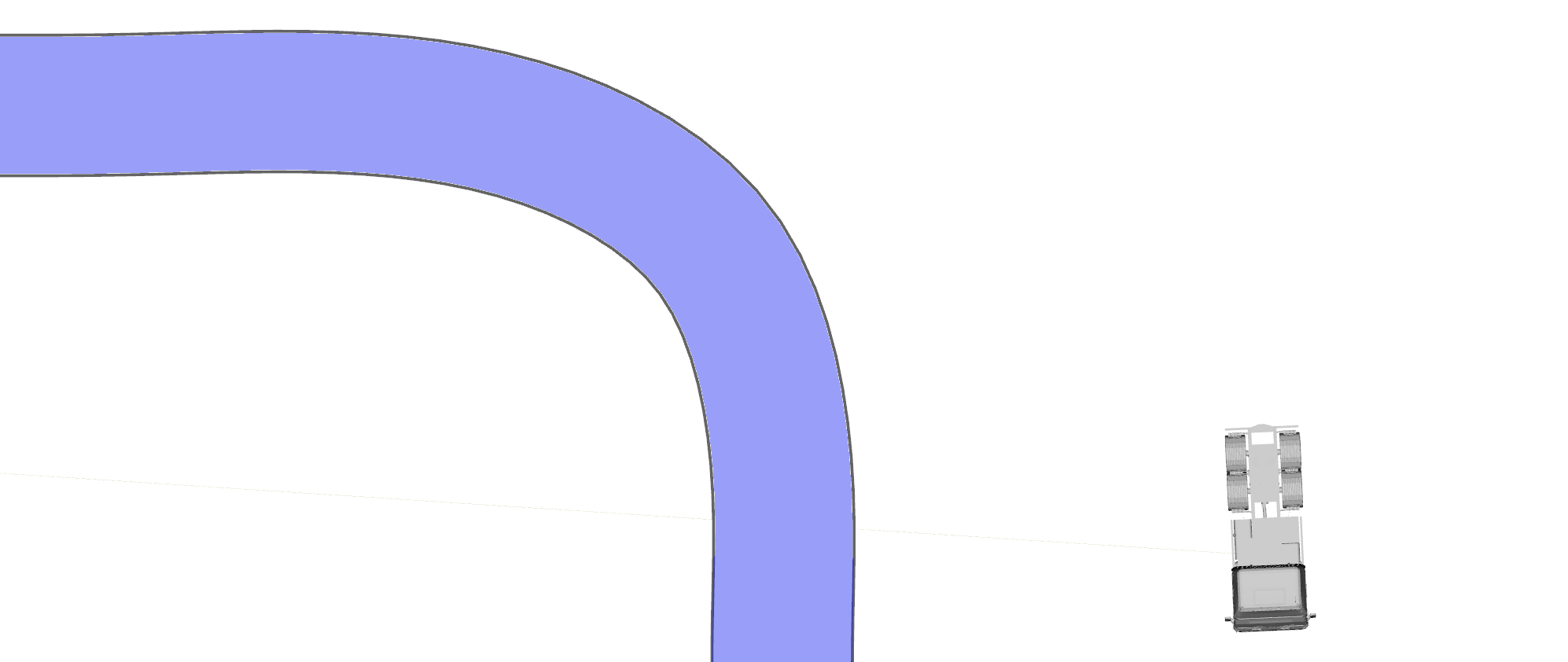}
        \label{fig:reducedmuturn:NA3}
    }
    \subfloat[TA, $t_c=5$s \mbox{(no accident)}]{%
        \includegraphics[width=0.45\columnwidth]{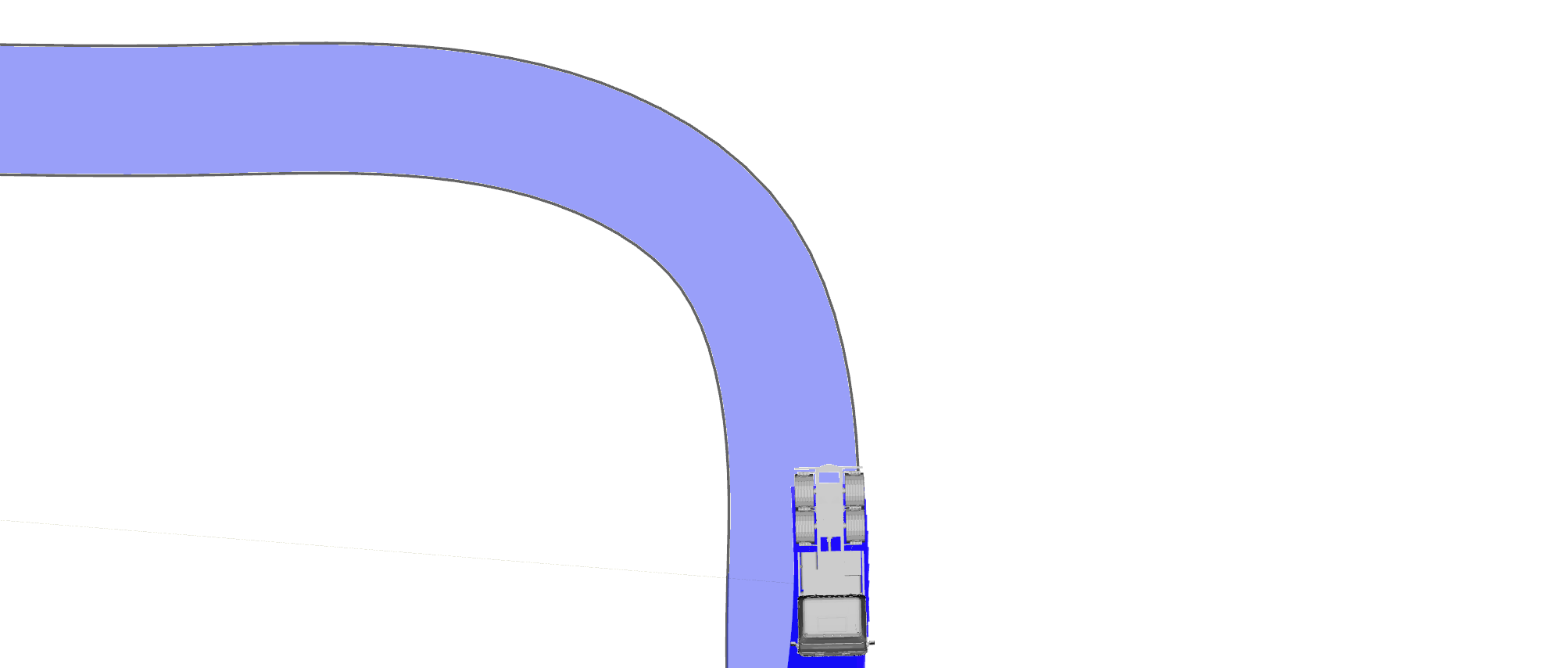}
        \label{fig:reducedmuturn:TA3}
    }\\
    \subfloat[Planned tire forces at $t_c=0$s (corresponding to \ref{fig:reducedmuturn:NA0} and \ref{fig:reducedmuturn:TA0})]{%
        \includegraphics[width=0.95\columnwidth]{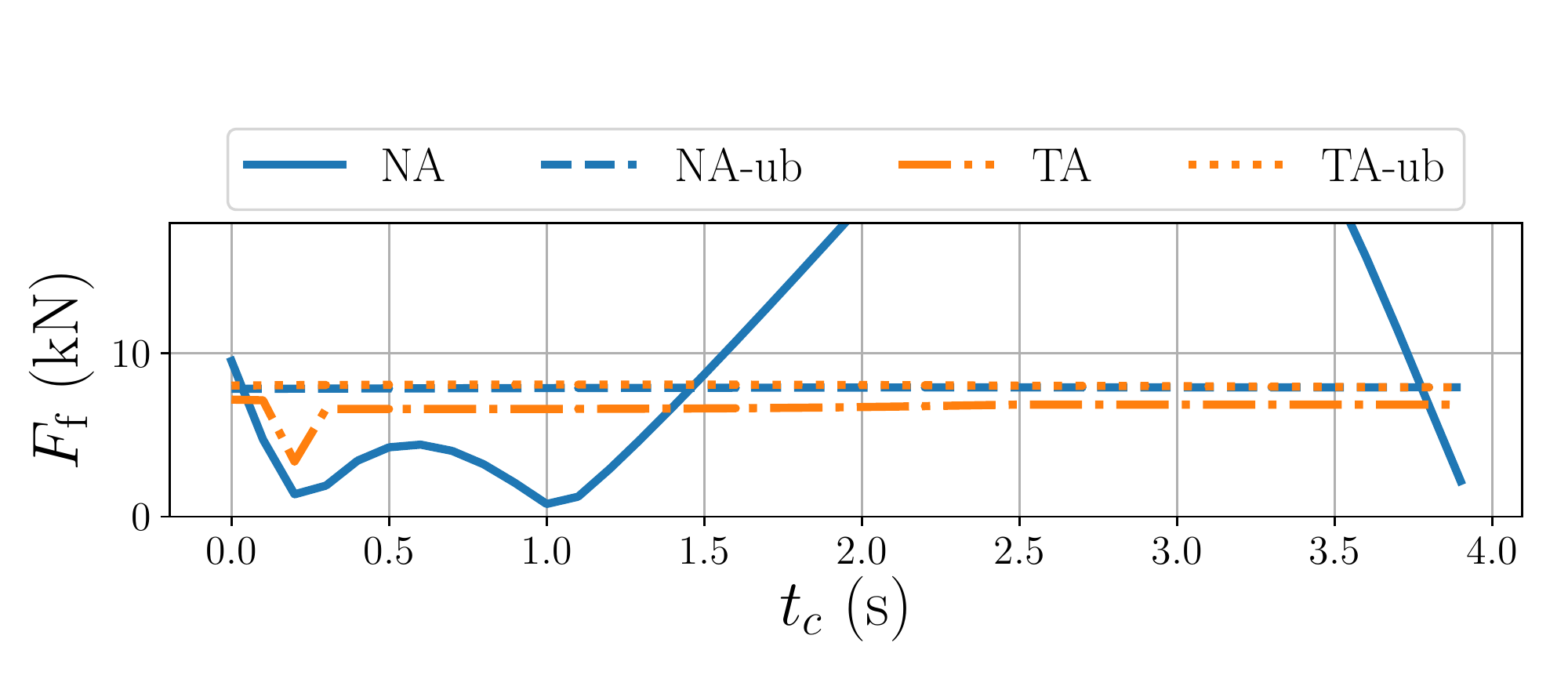}
        \label{fig:reducedmuturn:F}
    }\\
    \subfloat[Position w.r.t lane center]{%
        \includegraphics[width=0.45\columnwidth]{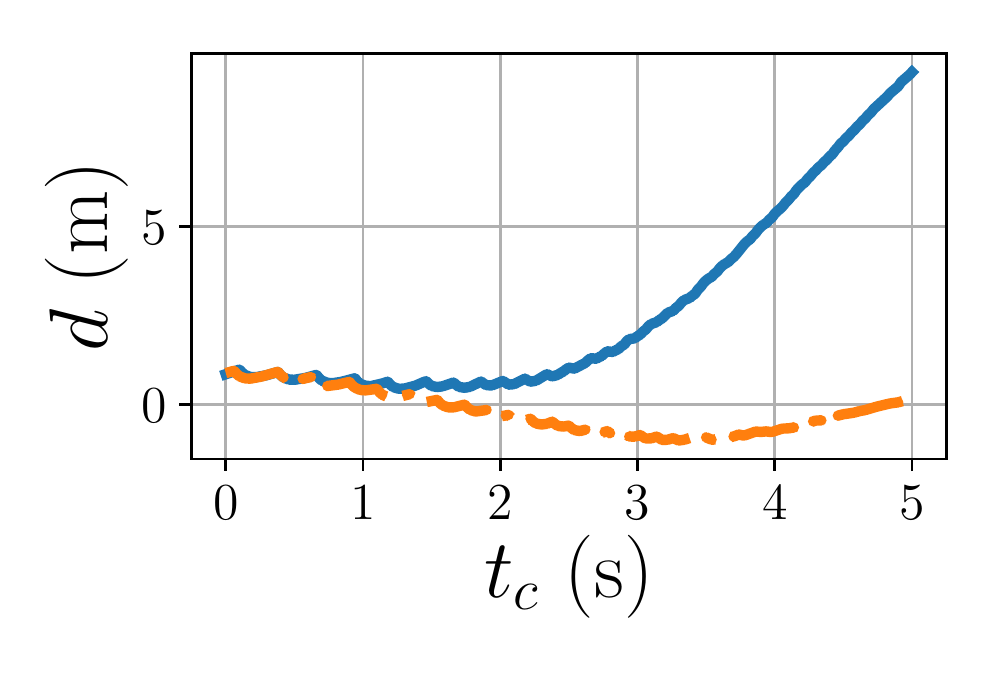}
        \label{fig:reducedmuturn:d}
    }
    \subfloat[Forward velocity]{%
        \includegraphics[width=0.45\columnwidth]{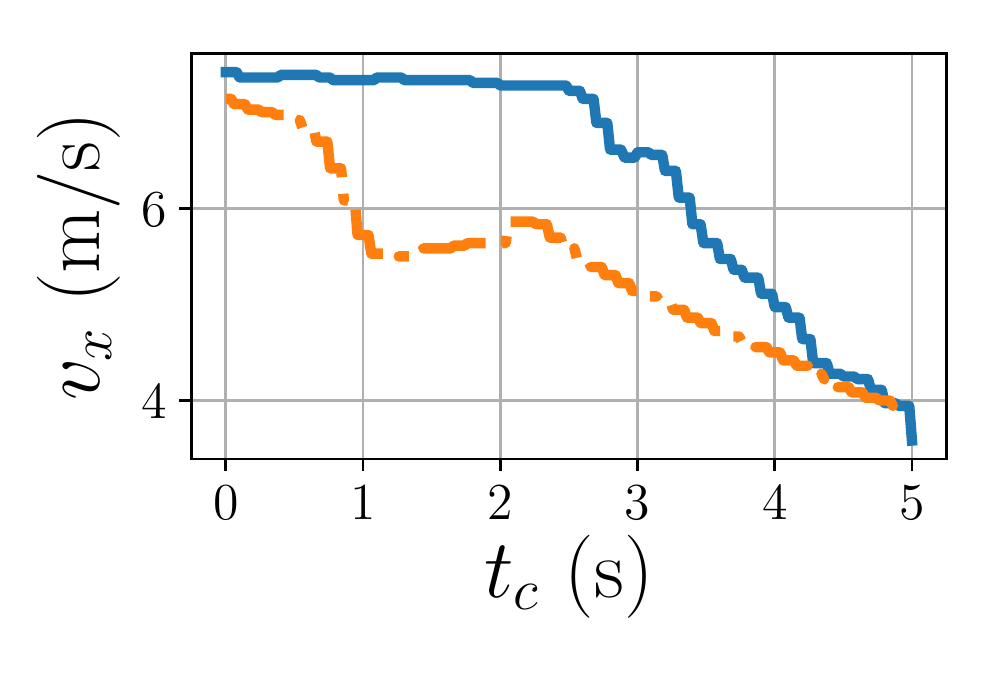}
        \label{fig:reducedmuturn:vx}
    }\\
    \subfloat[Front wheel slip angle]{%
        \includegraphics[width=0.45\columnwidth]{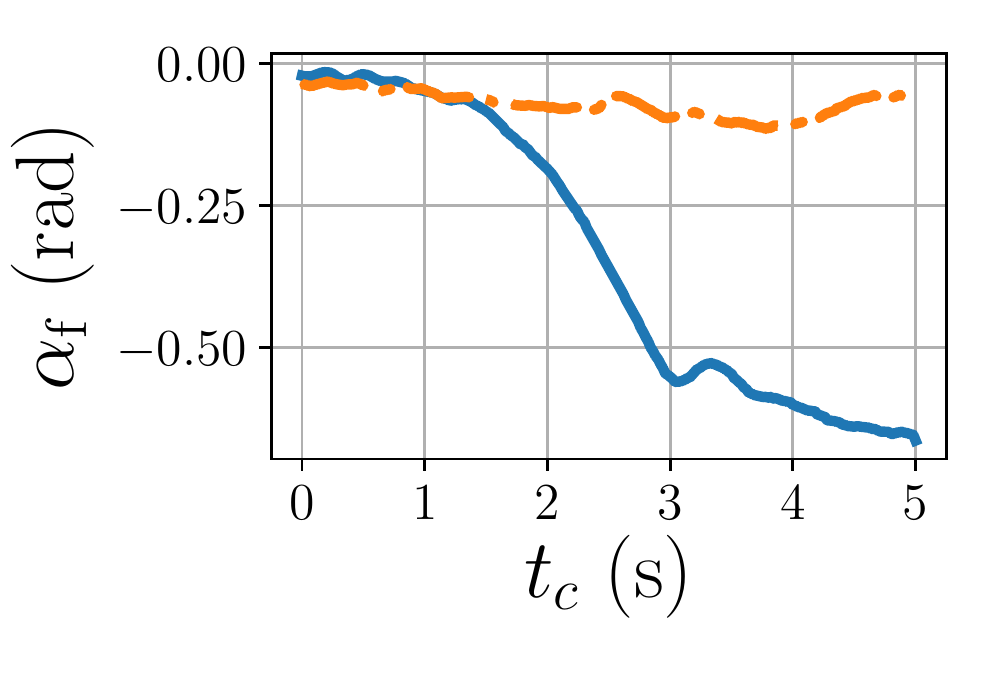}
        \label{fig:reducedmuturn:alphaf}
    }
    \subfloat[Lateral acceleration]{%
        \includegraphics[width=0.45\columnwidth]{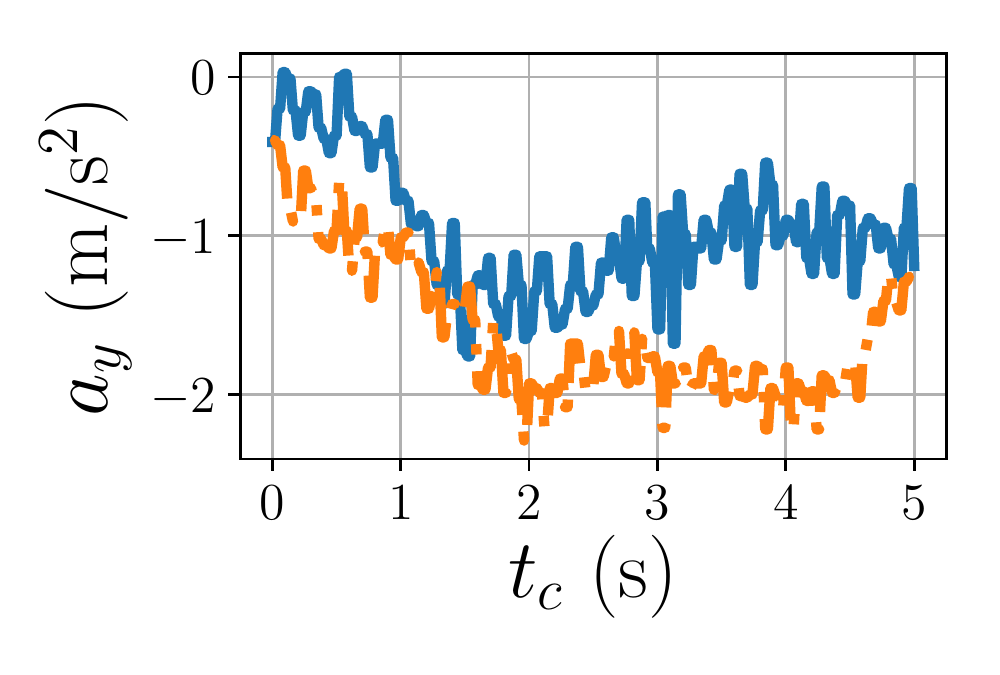}
        \label{fig:reducedmuturn:ay}
    }\\
    \end{minipage}\hfill
    \begin{minipage}[c]{0.45\textwidth}
    \centering
    \caption{Scenario (i), turn at low $\mu$. Subfigures \ref{fig:reducedmuturn:NA0} through \ref{fig:reducedmuturn:TA3} show overhead visualizations of the planned motion. The left and right columns correspond to the non-adaptive (NA) and traction adaptive (TA) schemes respectively. Subfigure \ref{fig:reducedmuturn:F} show planned tire forces and prevailing force limits for the front tires at $t_c=0$s. Subfigures \ref{fig:reducedmuturn:d}, \ref{fig:reducedmuturn:vx}, \ref{fig:reducedmuturn:alphaf} and \ref{fig:reducedmuturn:ay} show comparisons in terms of deviation from the lane center, velocity, front wheel slip angle and lateral acceleration.}
    \label{fig:reducedmuturn:full}
    \end{minipage}
\end{figure}

At the start of the first critical scenario at $t_c=0$s and $s=0$m, the vehicle approaches a turn at 8 m/s. Traction has deteriorated such that $\mu_{\textnormal{gt}}(s) = 0.2$ for $s \geq 0$m instead of the previous $\mu_{\textnormal{gt}}(s) = 0.8$ for $s < 0$m. The objective of the vehicle is to maintain its initial velocity and stay close to the center of the lane. Fig.~\ref{fig:reducedmuturn:full}, shows a driving behavior analysis comparing the TA scheme with time-varying tire force constraints corresponding to $\mu_{\textnormal{est}}(s) = \mu_{\textnormal{gt}}(s)$ and the NA scheme with static tire force constraints corresponding to $\mu_{\textnormal{sta}} = 0.8$. 

\subsubsection*{Not Adapting to Low Local Traction}
We make the following observations of the behavior under the NA scheme. At $t_c=0$s, before reaching the corner, Fig.~\ref{fig:reducedmuturn:NA0}, the planned motion appears to follow the intended objective of tracking the center of the lane and maintaining velocity. Immediately after entering the corner, Fig.~\ref{fig:reducedmuturn:NA1}, the vehicle starts to exhibit under-steering behavior that was not previously anticipated, i.e., the planned motion at $t_c=0$s did not include this motion, and the plan gets gradually shifted towards the outside of the corner. The vehicle continues to under-steer heavily, (see front slip angle in Fig.~\ref{fig:reducedmuturn:alphaf}), such that it eventually enters the opposing lane, Fig.~\ref{fig:reducedmuturn:NA2}, at a velocity of $7$m/s Fig.~\ref{fig:reducedmuturn:vx}. At this point in the experiment, the planner disengages and full braking is applied. The vehicle eventually comes to a stop 15m outside of the lane, Fig.~\ref{fig:reducedmuturn:NA3}.
The planned trajectory at the beginning of the scenario, Fig.~\ref{fig:reducedmuturn:NA0}, differs dramatically from the resulting closed loop trajectory, Figs.~\ref{fig:reducedmuturn:NA1}, \ref{fig:reducedmuturn:NA2}, \ref{fig:reducedmuturn:NA3}.

The reason behind this undesirable behavior is evident from the planned tire forces at $t_c=0$s, plotted in Fig.~\ref{fig:reducedmuturn:F}. The dashed blue and dotted orange lines represent the maximum horizontal forces that can be generated by the front and rear tires respectively, according to \eqref{eq:forcelimit}, given the recorded state trajectory. It is clear that the non-adaptive scheme plans tire forces (solid blue) that violate the physical limitations. Therefore, the vehicle is unable to realize the planned motion and promptly veers out of its lane. 

\subsubsection*{Adapting to Low Local Traction}
The adaptive scheme however, sets the tire force constraints dynamically according to the predictive friction estimate $\mu_\mathrm{est}(s)$. Therefore the planned tire forces (dash-dotted orange in Fig.~\ref{fig:reducedmuturn:F}) do not violate the physical limitations of the tires. Instead, it plans a motion, Fig.~\ref{fig:reducedmuturn:TA0}, that minimizes the deviation from the objective (track the center of the lane and maintain 8 m/s), while satisfying the updated tire force constraints. 
With the TA scheme, the vehicle slows down from $8$m/s to $5.5$m/s before entering the corner, Fig.~\ref{fig:reducedmuturn:vx} and keeps a position on the outside of the turn. About a third of the way through the turn, Fig.~\ref{fig:reducedmuturn:TA1}, the vehicle has cut across the lane and is positioned along the inside of the corner, Fig.~\ref{fig:reducedmuturn:d}. Towards the end of the turn, Fig.~\ref{fig:reducedmuturn:TA3}, it starts cutting back across to the outside of the corner. Finally, the vehicle completes the turn safely, positioned at the outer edge of the lane, Fig.~\ref{fig:reducedmuturn:TA3}, Fig.~\ref{fig:reducedmuturn:d}.
In contrast to the NA scheme, here the planned trajectory at the beginning of the scenario, Fig.~\ref{fig:reducedmuturn:TA0}, corresponds well with the closed loop trajectory of the vehicle, Figs.~\ref{fig:reducedmuturn:TA1}, \ref{fig:reducedmuturn:TA2}, \ref{fig:reducedmuturn:TA3}, 

We break this behavior down into three coordinated characteristics of the maneuver that enable the vehicle to successfully negotiate the turn. 
First, contrary to the NA scheme, the TA scheme generates feasible tire force commands, utilizing approximately 90\% ($\lambda = 0.90$) of the available traction for this surface, see Fig.~\ref{fig:reducedmuturn:F}. This in turn generates slip angles within acceptable limits, Fig~\ref{fig:reducedmuturn:alphaf}, yielding high lateral acceleration, Fig~\ref{fig:reducedmuturn:ay}.  
Second, the maneuver utilizes the full width of the lane to minimize the curvature of the vehicle trace, thus reducing the lateral acceleration required to negotiate the turn. 
Third, the longitudinal velocity is reduced early in the maneuver such that the maximum tire forces are sufficient to negotiate the turn. 

These coordinated adjustments to the motion plan are a direct result of the reduced tire force constraint set $\mathcal{P}_{k|t}^\mathcal{U}$, for all $k \in \{0,1,\dots,(N-1)\}$ in the optimization problem \eqref{eq:mpc_problem}. The planned motion utilizes 90$\%$ of the available local traction, such that control authority is maintained throughout the maneuver. Thus, the planning/control scheme safely mitigates the critical situation.
Next, we evaluate the traction adaptation concept in collision avoidance scenarios.  

\subsection{Scenario (ii): Collision Avoidance at Low Local Traction}
\label{sec:results:coll_avoid_reduced_mu}
\begin{figure}[!]
\captionsetup[subfigure]{}
\begin{minipage}[c]{0.5\textwidth}
\centering
    \subfloat[NA, $t_c=0$s]{%
        \includegraphics[width=0.45\columnwidth]{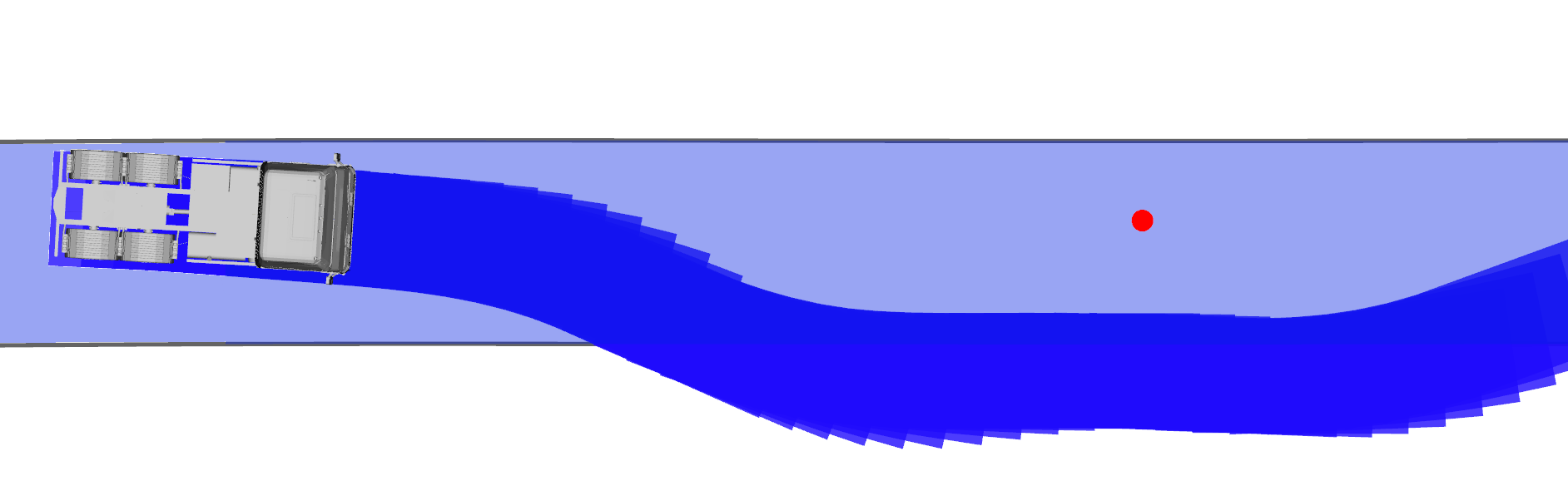}
        \label{fig:reducedmuCA:NA0}
    }
    \subfloat[TA, $t_c=0$s]{%
        \includegraphics[width=0.45\columnwidth]{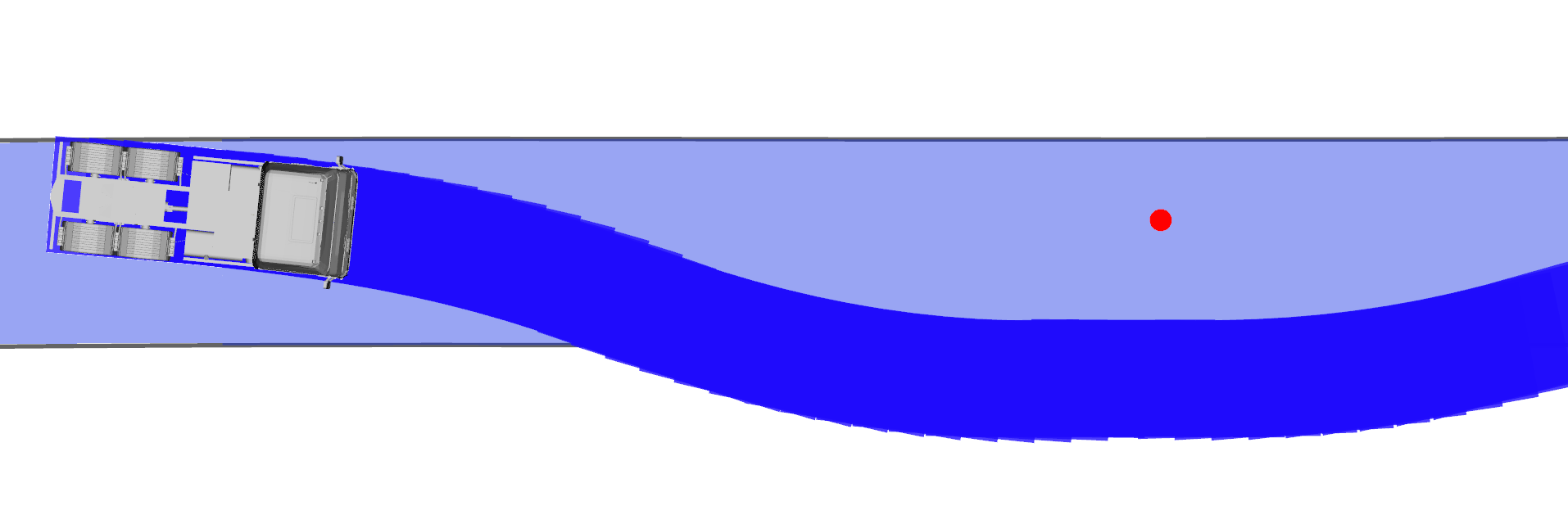}
        \label{fig:reducedmuCA:TA0}
    }\\
    \subfloat[NA, $t_c=2$s \mbox{{(under-steering)}}]{%
        \includegraphics[width=0.45\columnwidth]{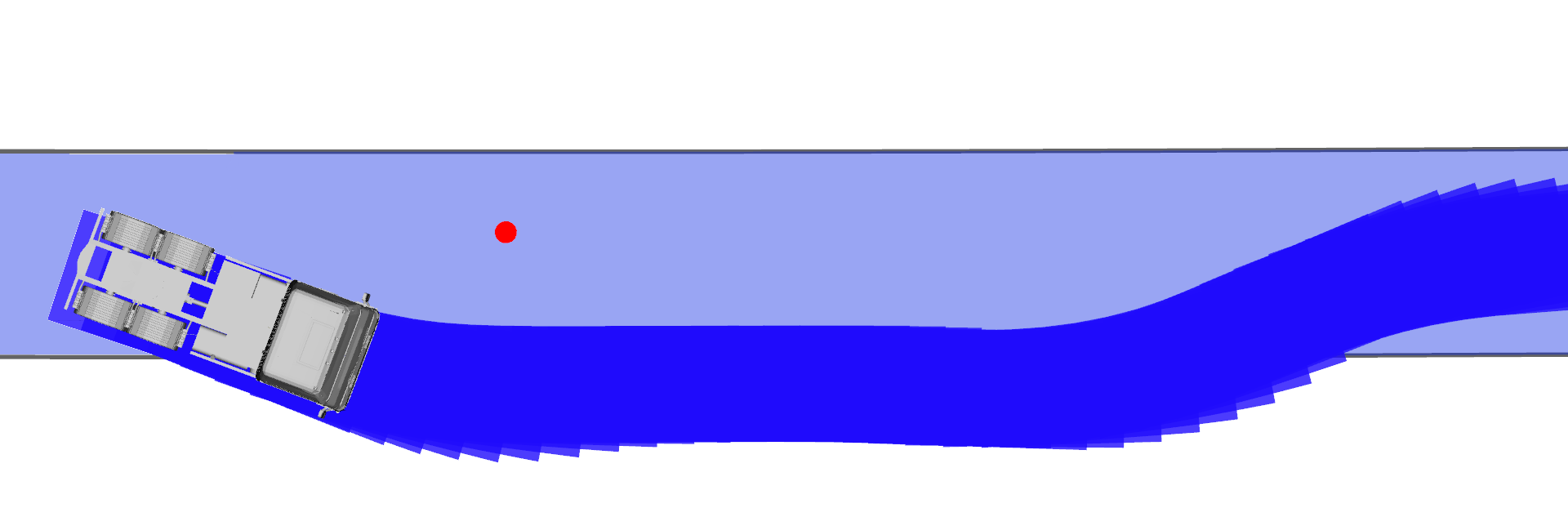}
        \label{fig:reducedmuCA:NA1}
    }
    \subfloat[TA, $t_c=2$s]{%
        \includegraphics[width=0.45\columnwidth]{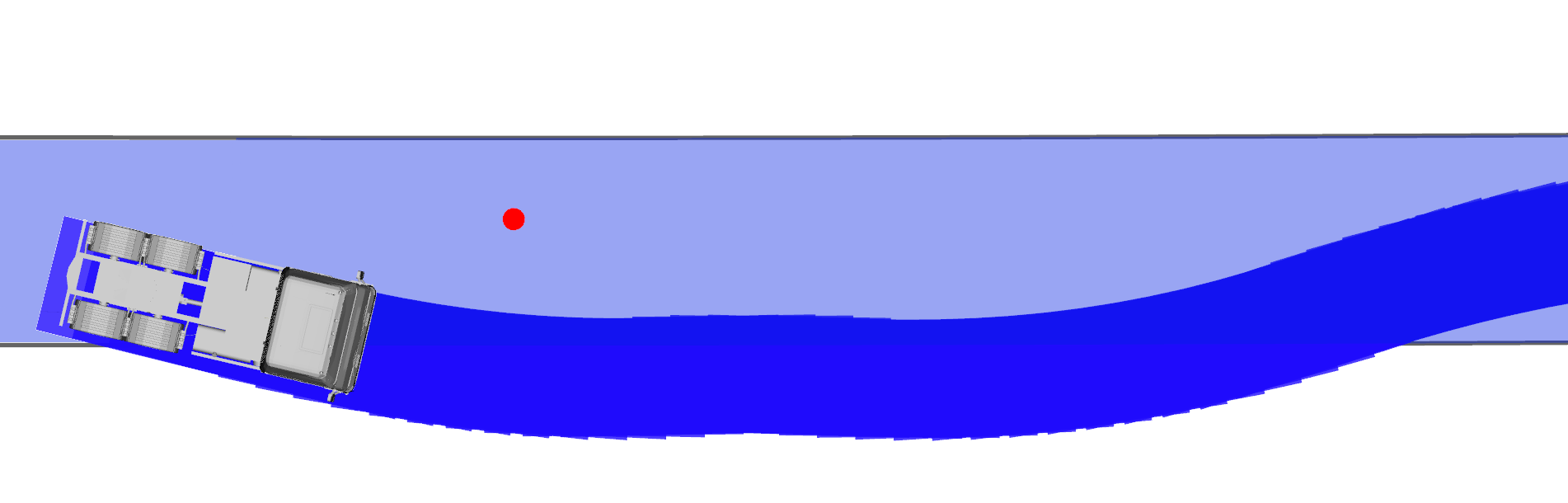}
        \label{fig:reducedmuCA:TA1}
    }\\
    \subfloat[NA, \mbox{$t_c=3.1$s \mbox{{(exiting lane)}}} ]{%
        \includegraphics[width=0.45\columnwidth]{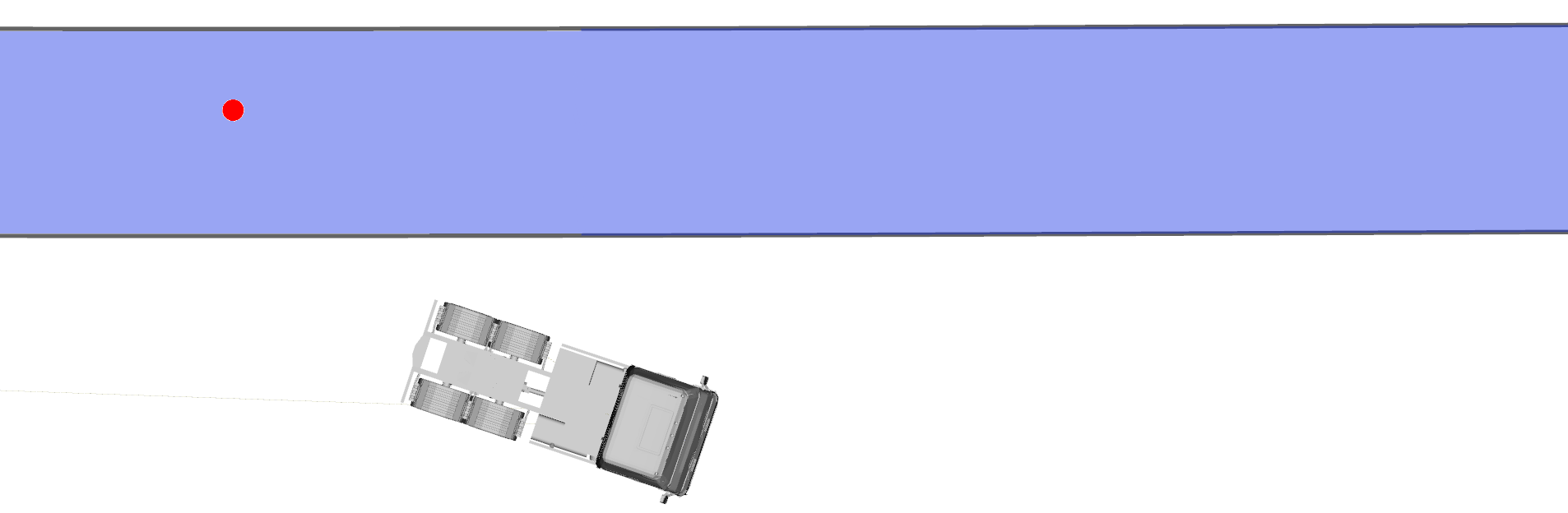}
        \label{fig:reducedmuCA:NA2}
    }
    \subfloat[TA, $t_c=3.1$s]{%
        \includegraphics[width=0.45\columnwidth]{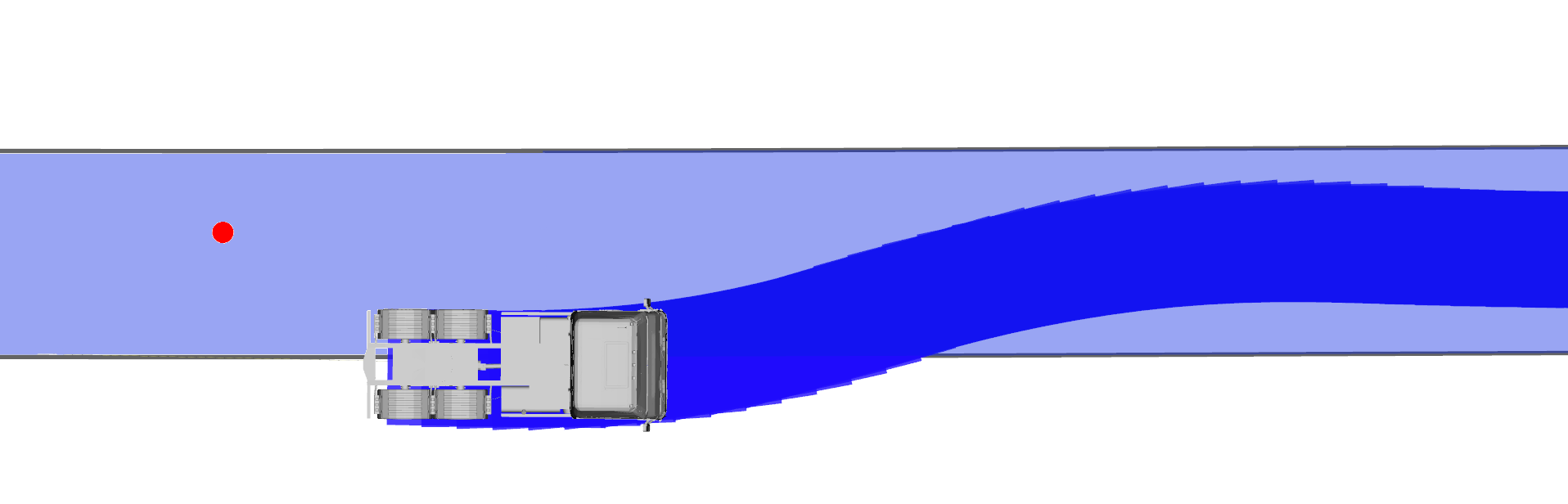}
        \label{fig:reducedmuCA:TA2}
    }\\
    \subfloat[NA, $t_c=5$s \mbox{{(accident)}}]{%
        \includegraphics[width=0.45\columnwidth]{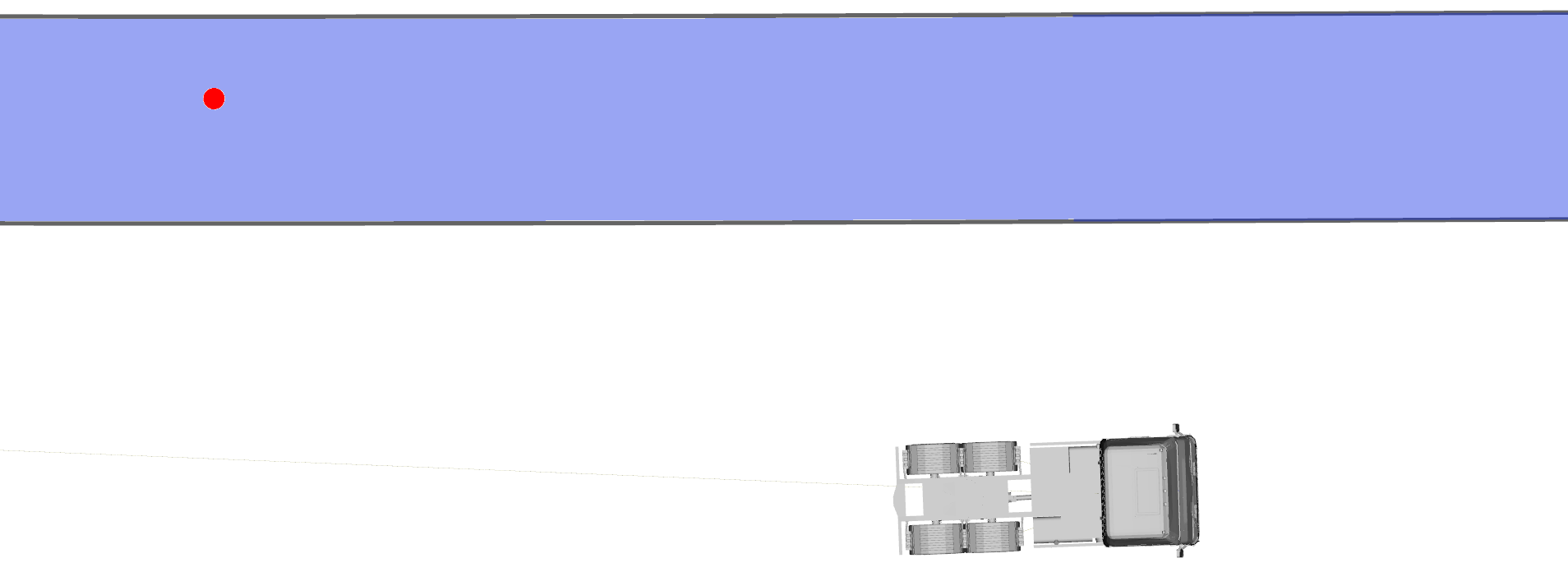}
        \label{fig:reducedmuCA:NA3}
    }
    \subfloat[TA, $t_c=5$s, \mbox{{(no accident)}}]{%
        \includegraphics[width=0.45\columnwidth]{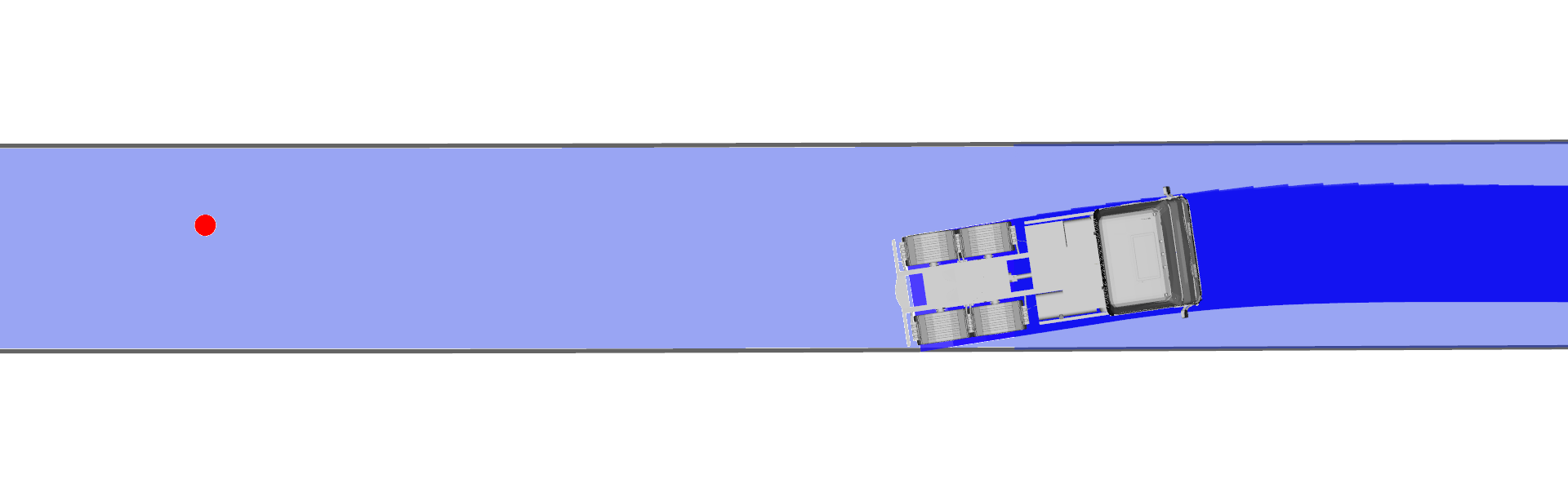}
        \label{fig:reducedmuCA:TA3}
    }\\
    \subfloat[Planned tire forces at $t_c=0$s (corresponding to \ref{fig:reducedmuCA:NA0} and \ref{fig:reducedmuCA:TA0})]{%
        \includegraphics[width=0.95\columnwidth]{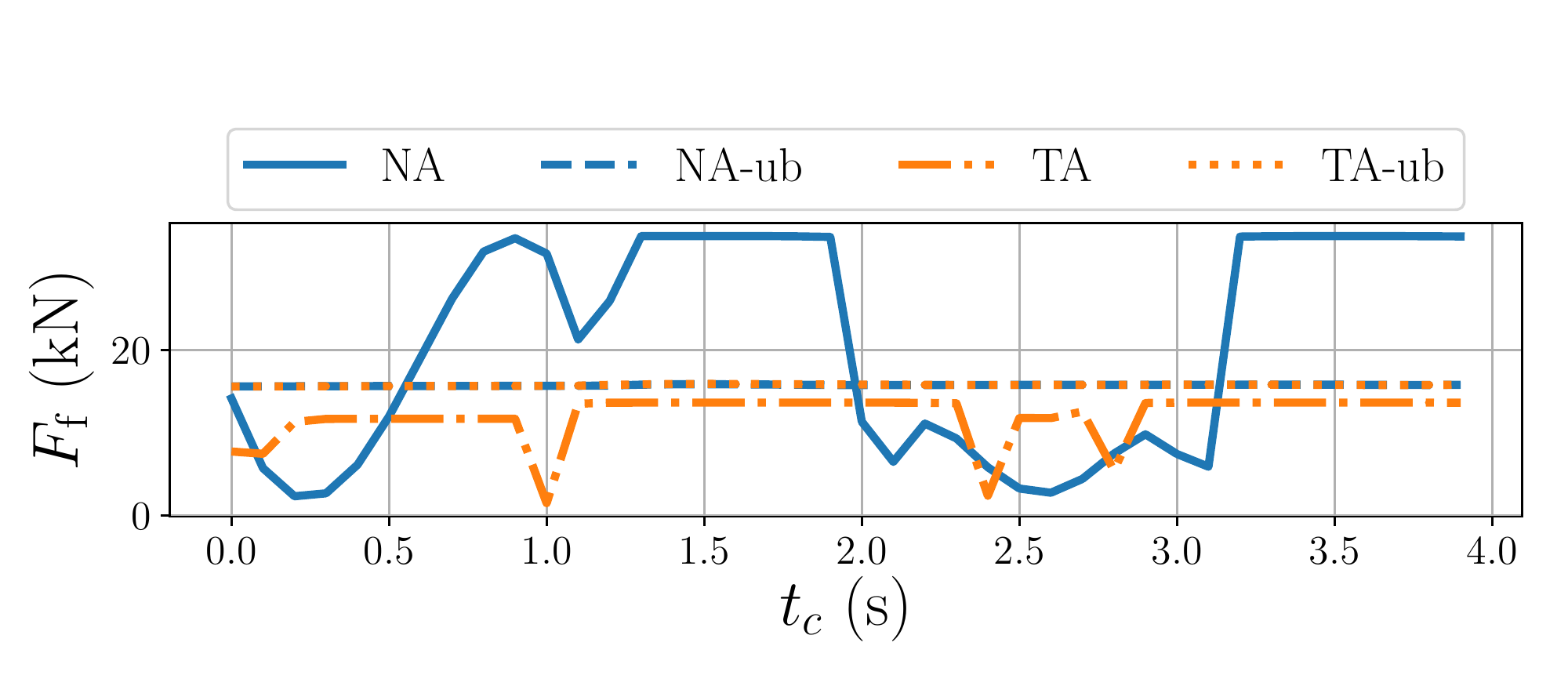}
        \label{fig:reducedmuCA:F}
    }\\
    \subfloat[Position w.r.t lane center]{%
        \includegraphics[width=0.45\columnwidth]{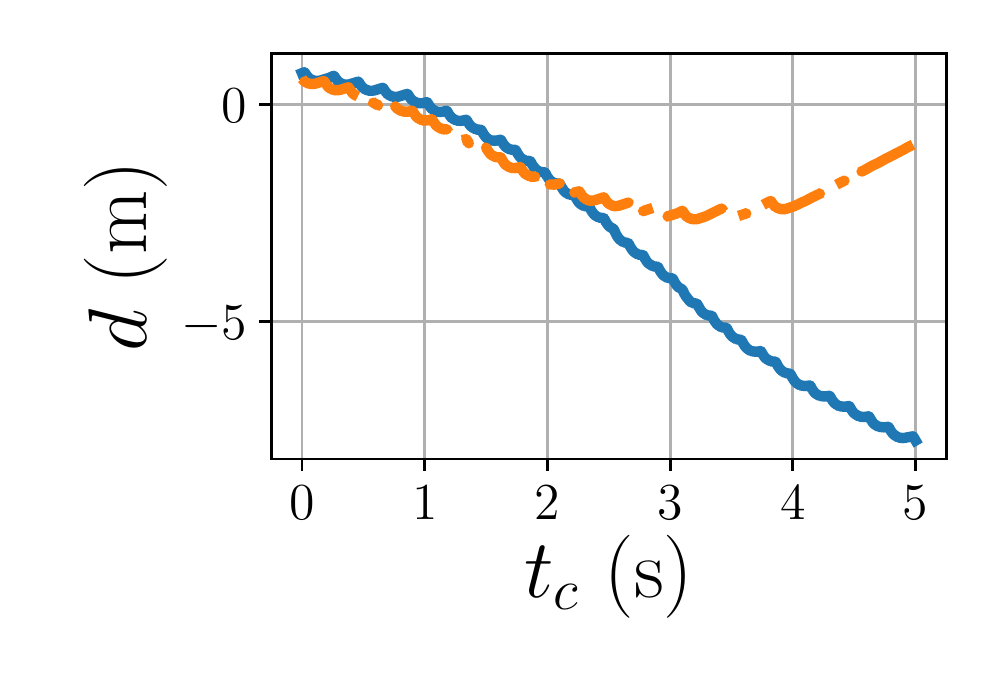}
        \label{fig:reducedmuCA:d}
    }
    \subfloat[Forward velocity]{%
        \includegraphics[width=0.45\columnwidth]{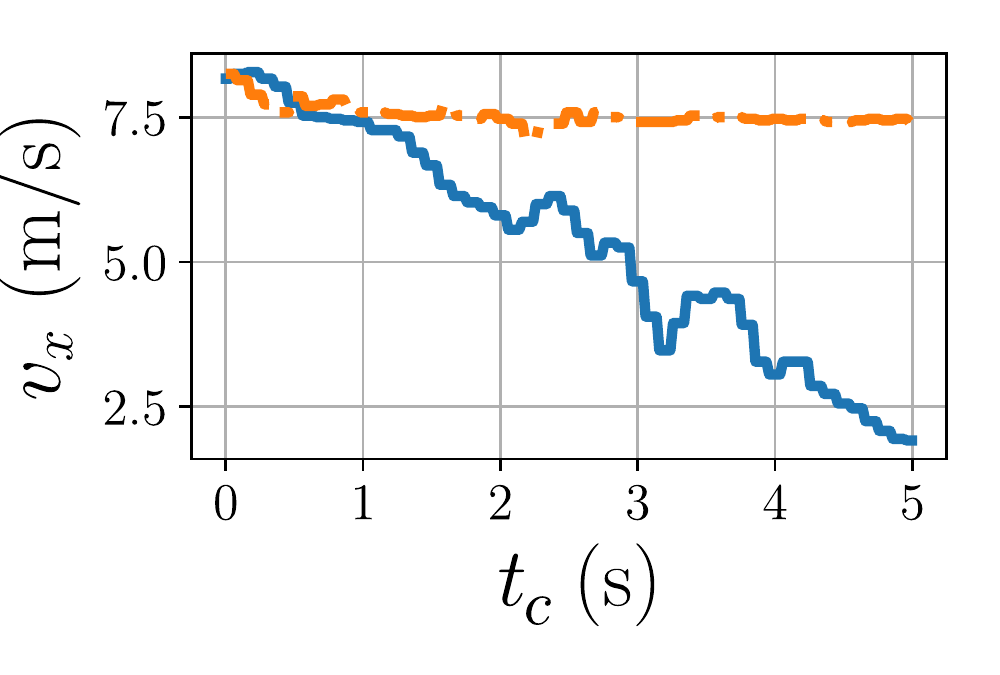}
        \label{fig:reducedmuCA:vx}
    }\\
    \subfloat[Front wheel slip angle]{%
        \includegraphics[width=0.45\columnwidth]{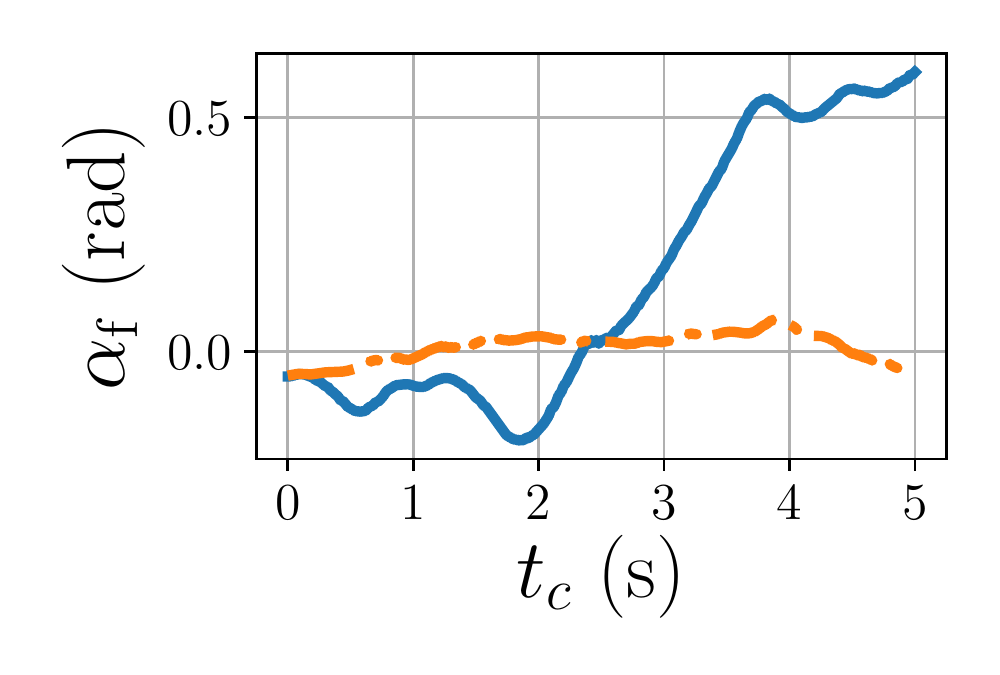}
        \label{fig:reducedmuCA:alphaf}
    }
    \subfloat[Lateral acceleration]{%
        \includegraphics[width=0.45\columnwidth]{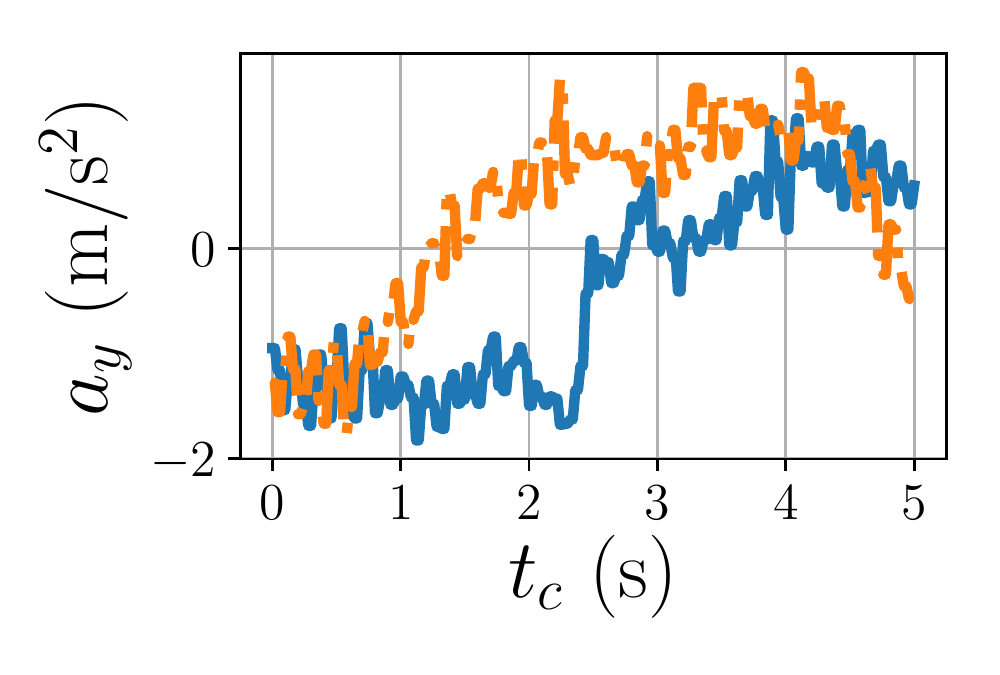}
        \label{fig:reducedmuCA:ay}
    }\\
    \end{minipage}\hfill
    \begin{minipage}[c]{0.45\textwidth}
    \centering

    \caption{Scenario (ii), collision avoidance at Low traction conditions. Subfigures \ref{fig:reducedmuCA:NA0} through \ref{fig:reducedmuCA:TA3} show overhead visualizations of the planned motion. The left and right columns corresponds to the non-adaptive (NA) and traction adaptive (TA) schemes respectively. Subfigure \ref{fig:reducedmuCA:F} show planned tire forces and prevailing force limits for the front tire at $t_c=0$s. Subfigures \ref{fig:reducedmuCA:d}, \ref{fig:reducedmuCA:vx}, \ref{fig:reducedmuCA:alphaf} and \ref{fig:reducedmuCA:ay} show comparisons in terms of deviation from the lane center, forward velocity, front wheel slip angle and lateral acceleration.}
    \label{fig:reduced_mu_coll_avoid:full}
    \end{minipage}
\end{figure}

The second scenario is aimed at replicating the conditions when a pedestrian has suddenly entered the lane from the side of the road. At the start of the second scenario, $t_c=0$s, the pedestrian has been detected by the vehicle's perception stack and the planner/controller starts reacting. The initial velocity of the vehicle is 8m/s and the pedestrian is detected 20m ahead of the vehicle. Traction conditions are the same as in the previous case, i.e., $\mu_{\textnormal{gt}}(s) = 0.8$ for $s<0$ and $\mu_{\textnormal{gt}}(s) = 0.2$ for $s \geq 0$. The NA scheme operates under static tire force constraints corresponding to a static friction estimate of $\mu_{\textnormal{sta}} = 0.8$, while the adaptive scheme sets its tire force constraints dynamically according to $\mu_{\textnormal{est}}(s) = \mu_{\textnormal{gt}}(s)$.
The objective of the vehicle is to avoid collision with the obstacle, while minimizing deviations from the lane center. 
The performance of the NA and TA schemes in this scenario is presented in Fig.~\ref{fig:reduced_mu_coll_avoid:full}. 

\subsubsection*{Not Adapting to Low Local Traction}
Under the non-adaptive scheme, we make the following observations. Upon detecting the pedestrian, Fig.~\ref{fig:reducedmuCA:NA0}, the NA scheme initiates an aggressive maneuver, i.e., high curvature given the speed, that is meant to quickly move the vehicle away from its collision course with the obstacle. However, the vehicle is not able to realize the aggressive motion, so the plan is gradually shifted forward, ahead of the vehicle. At $t_c=2$s, Fig.~\ref{fig:reducedmuCA:NA1}, the vehicle has eventually moved 2m to the right of the lane center, Fig.~\ref{fig:reducedmuCA:d}, enough to clear the obstacle. However, this occurs closer to the obstacle than was planned at $t_c=0$s. Also, at this point the vehicle exhibits heavy under-steering behavior while turning right (see plotted front slip angle in Fig.~\ref{fig:reducedmuCA:alphaf}). After clearing the obstacle, the vehicle starts steering left to return to the lane, but the grip is insufficient to counteract the motion out of the lane that was initiated during the first part of the maneuver. At $t_c=3.1$s, Fig.~\ref{fig:reducedmuCA:NA2}, the high slip angles have lead to a complete loss of control authority, and the vehicle is now veering out of the lane. At this point in the experiment, the planner disengages and full braking is applied. The vehicle eventually comes to a stop 7m outside the lane Fig.~\ref{fig:reducedmuCA:NA3}.  

The underlying reason is again revealed by the planned tire forces at $t_c=0$s, plotted in Fig.~\ref{fig:reducedmuCA:F}. The NA scheme (solid blue), plans tire forces that exceed the physical limit (dotted blue), and therefore, the planned motion is not physically attainable. The fact that the real vehicle cannot realize the planned motion has two separate unfavorable consequences in this case: First, the vehicle cannot change direction as fast as the planner dictates, therefore, the margin to the obstacle ends up being smaller than what was originally planned, see Figs.~\ref{fig:reducedmuCA:NA0} and \ref{fig:reducedmuCA:NA1}. Second, the overly aggressive maneuver leads to high slip angles, Fig.~\ref{fig:reducedmuCA:alphaf}. which eventually leads to  complete loss of control authority, Fig.~\ref{fig:reducedmuCA:NA2}, due to tire saturation \cite{rajamani2011vehicle}. 

\subsubsection*{Adapting to Low Local Traction}
As we saw in the previous scenario, the TA scheme dynamically updates tire force constraints such that the planned motions are dynamically feasible with respect to the local traction conditions. Therefore, in this case we observe that the initial plan from the TA scheme at $t_c=0$s, Fig.~\ref{fig:reducedmuCA:TA0} is a less aggressive maneuver compared to the NA scheme. At $t_c=2$s, Fig.~\ref{fig:reducedmuCA:TA0}, the vehicle has moved 2m to the right of the lane center, enough to clear the obstacle, without generating excessive front wheel slip, Fig.~\ref{fig:reducedmuCA:alphaf}. At $t_c=3.1$s, Fig.~\ref{fig:reducedmuCA:TA2}, the vehicle has cleared the obstacle and at $t_c=5$s, Fig.~\ref{fig:reducedmuCA:TA3}, the vehicle is returning from a comparatively small deviation from the lane, maintaining full control authority. 

Fig.~\ref{fig:reducedmuCA:F} reveals that the planned motion from the TA scheme at $t_c=0$s exploits approximately $90\%$ ($\lambda = 0.90$) of the available traction force for large parts of the maneuver (dash-dotted orange), while the NA scheme (solid blue) exceeds the traction bound by a large margin. Hence, apart from a small margin to the absolute traction limit, it delivers the most aggressive plan that is achievable at the prevailing conditions, yielding manageable slip angles, Fig.~\ref{fig:reducedmuCA:alphaf}, and therefore maintained control authority. As such, the critical scenario is mitigated. 

Furthermore, we observe from Fig.~\ref{fig:reducedmuCA:d}, showing deviation from the lane center, that for the first two seconds of the evasive maneuver, the sideways motion of the vehicle is almost identical for the two schemes. From this we get an indication that sacrificing control authority of the vehicle in an overly aggressive maneuver, does not improve the capacity to avoid colliding with the obstacle. It does however increase the risk of a secondary accident as the vehicle enters the opposing lane at high speed in an uncontrolled manner. 

\begin{figure}[!]
\captionsetup[subfigure]{}
\begin{minipage}[c]{0.5\textwidth}
\centering
    \subfloat[NA, $t_c=0$s]{%
        \includegraphics[width=0.45\columnwidth]{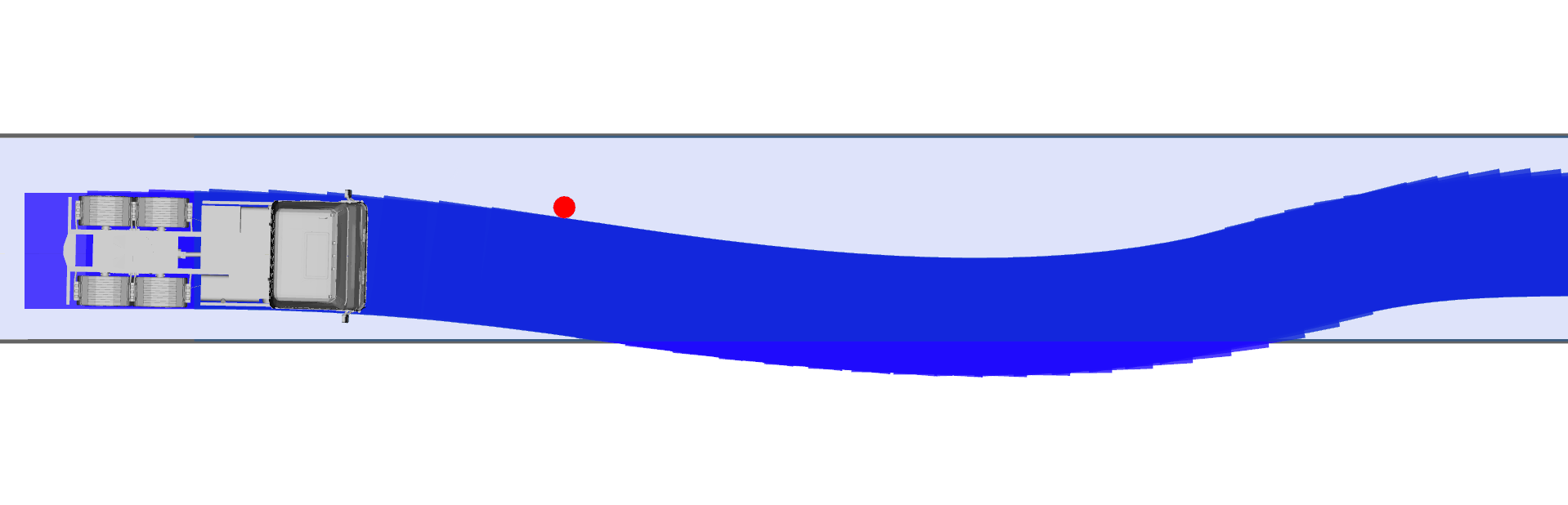}
        \label{fig:highmuCA:NA0}
    }
    \subfloat[TA, $t_c=0$s]{%
        \includegraphics[width=0.45\columnwidth]{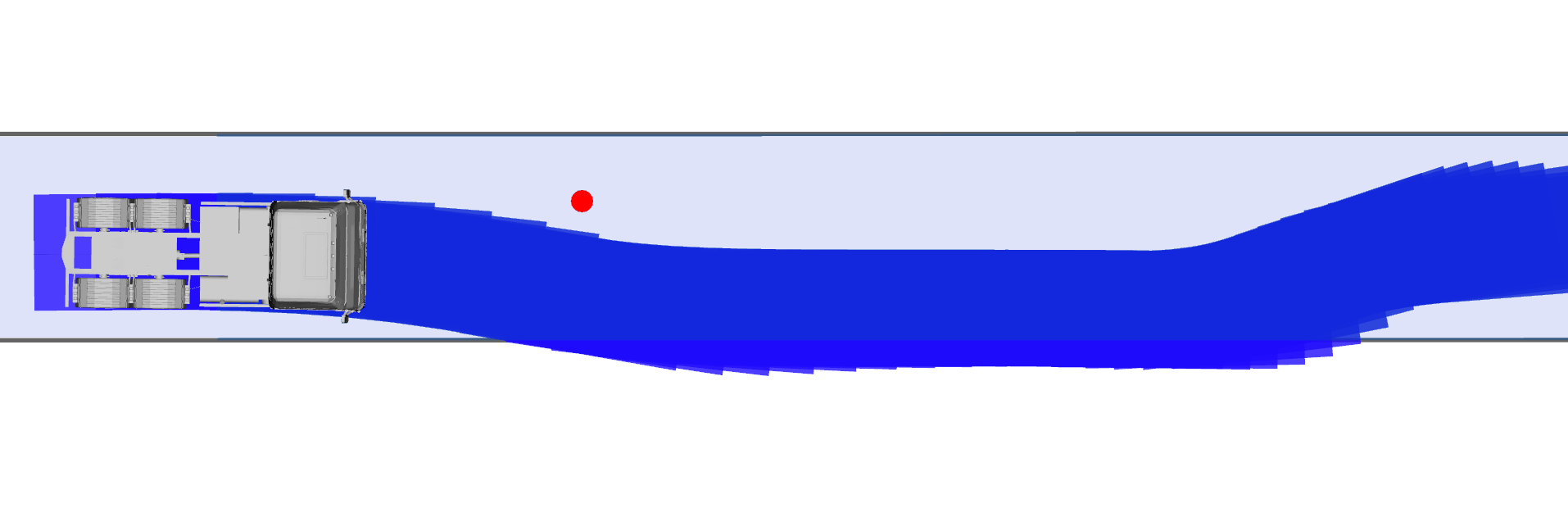}
        \label{fig:highmuCA:TA0}
    }\\
    \subfloat[NA, $t_c=0.7$s (accident)]{%
        \includegraphics[width=0.45\columnwidth]{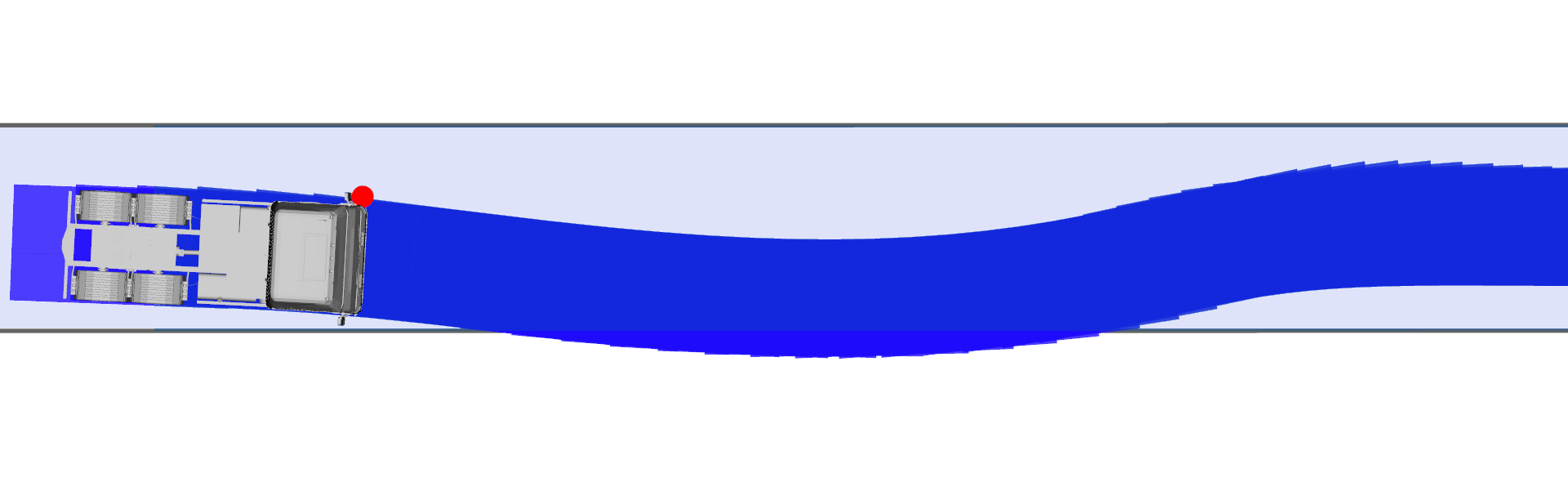}
        \label{fig:highmuCA:NA1}
    }
    \subfloat[TA, $t_c=0.7$s (evading)]{%
        \includegraphics[width=0.45\columnwidth]{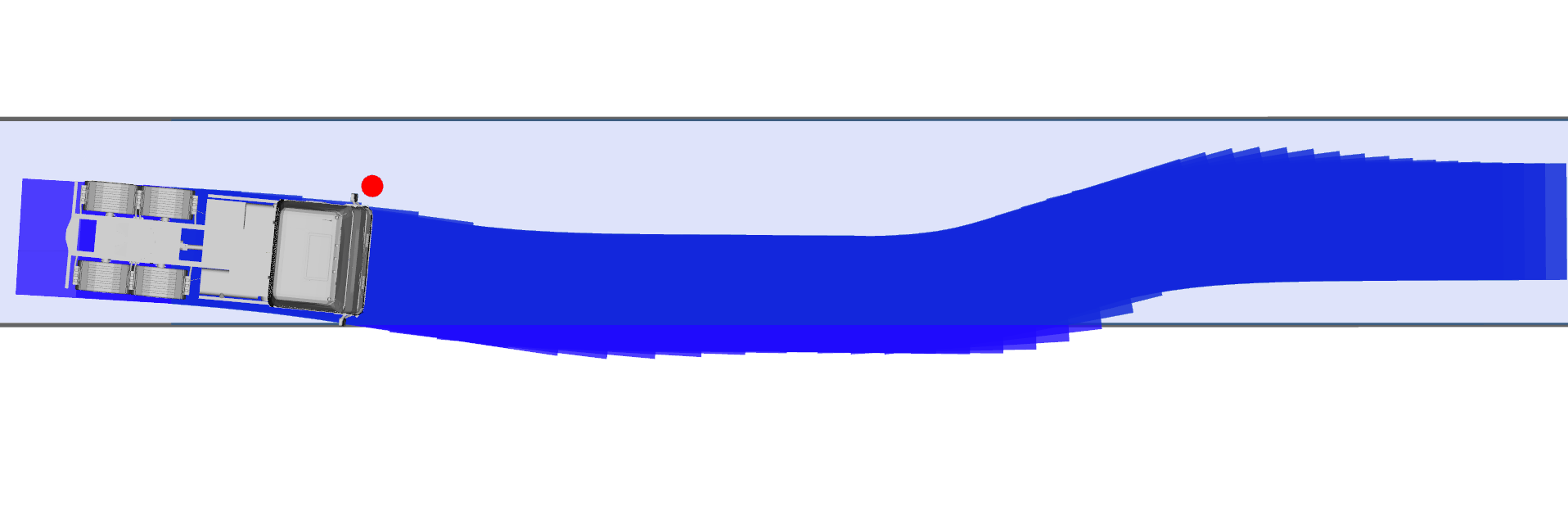}
        \label{fig:highmuCA:TA1}
    }\\
    \subfloat{%
        \includegraphics[width=0.45\columnwidth]{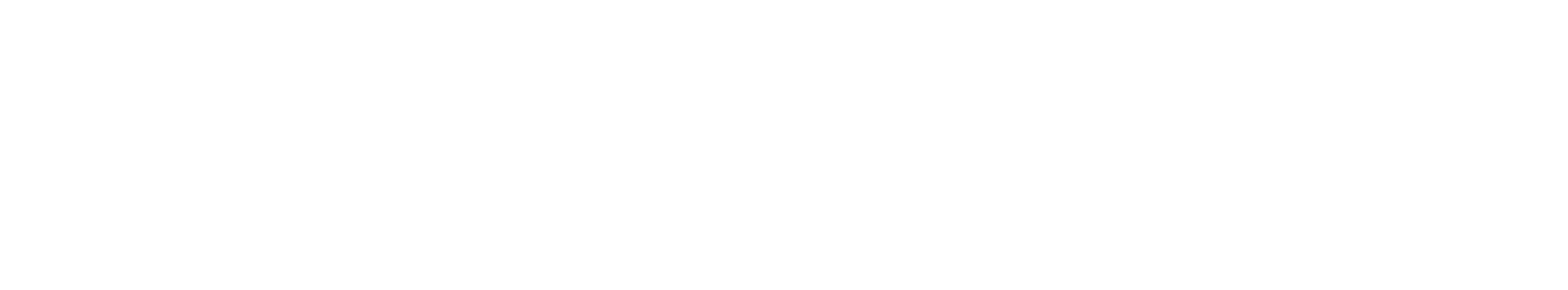}
    }
    \addtocounter{subfigure}{-1} 
    \subfloat[TA, $t_c=3$s (no accident) ]{%
        \includegraphics[width=0.45\columnwidth]{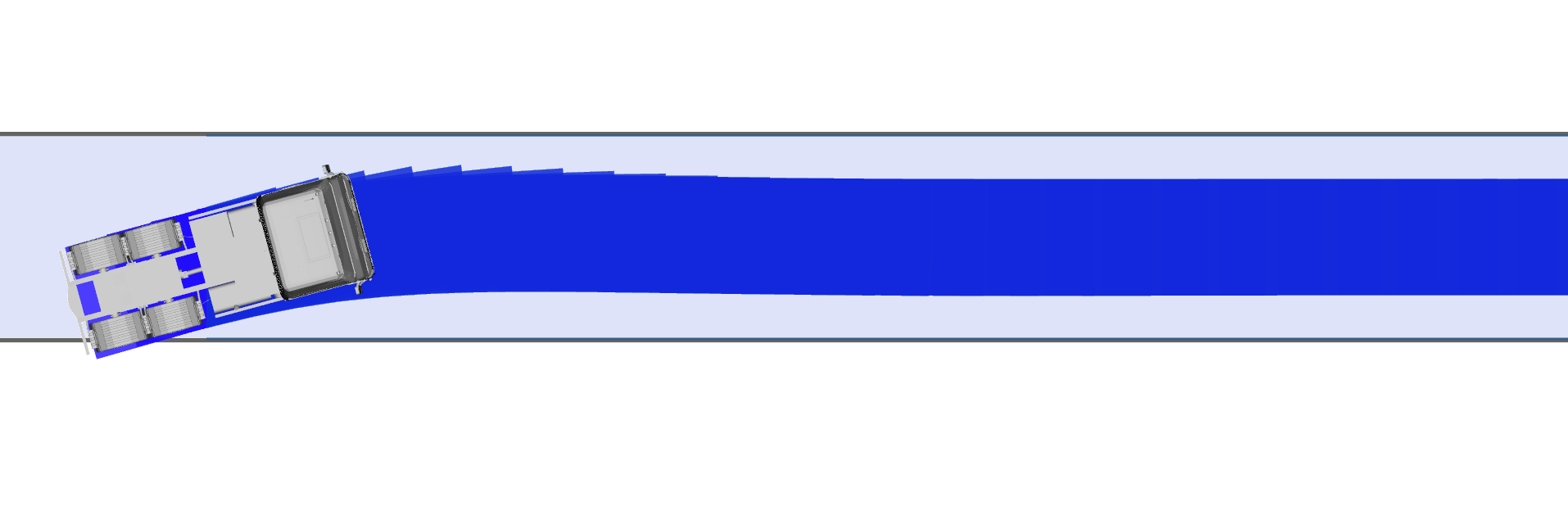}
        \label{fig:highmuCA:TA2}
    }\\
    \subfloat[Planned tire forces at $t_c=0$s (corresponding to \ref{fig:highmuCA:NA0} and \ref{fig:highmuCA:TA0})]{%
        \includegraphics[width=0.95\columnwidth]{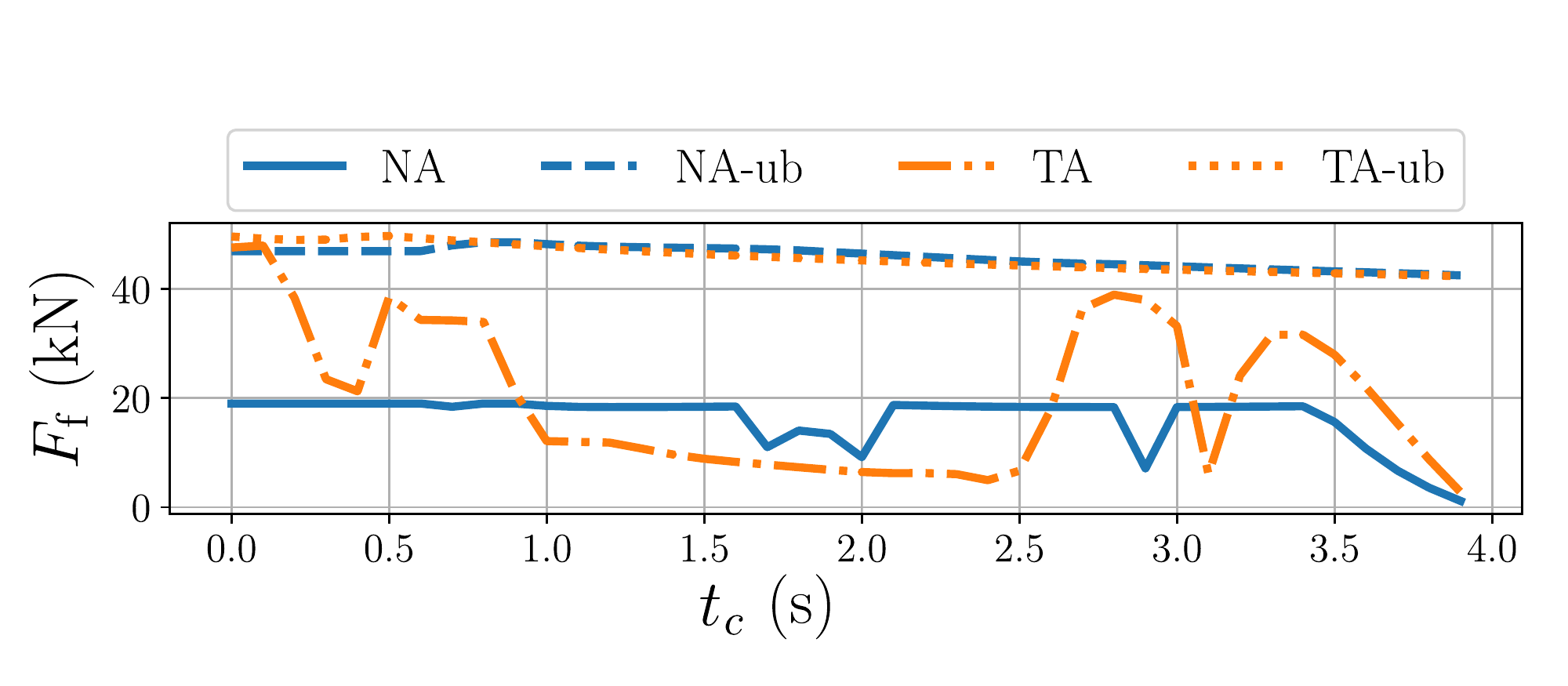}
        \label{fig:highmuCA:F}
    }\\
    \subfloat[Position w.r.t. lane center]{%
        \includegraphics[width=0.45\columnwidth]{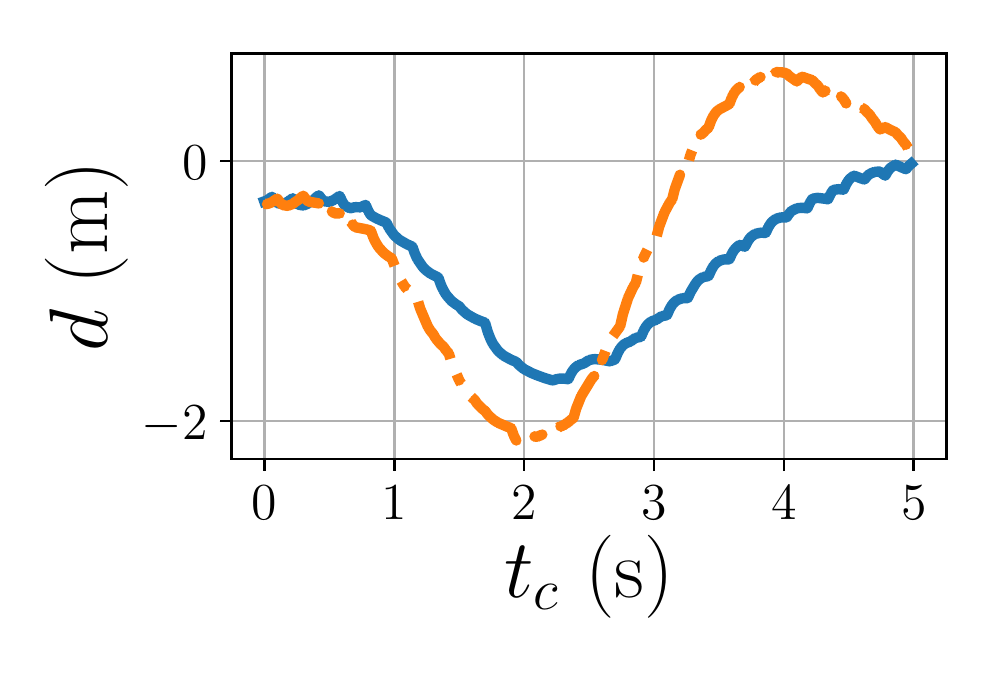}
        \label{fig:highmuCA:d}
    }
    \subfloat[Forward velocity]{%
        \includegraphics[width=0.45\columnwidth]{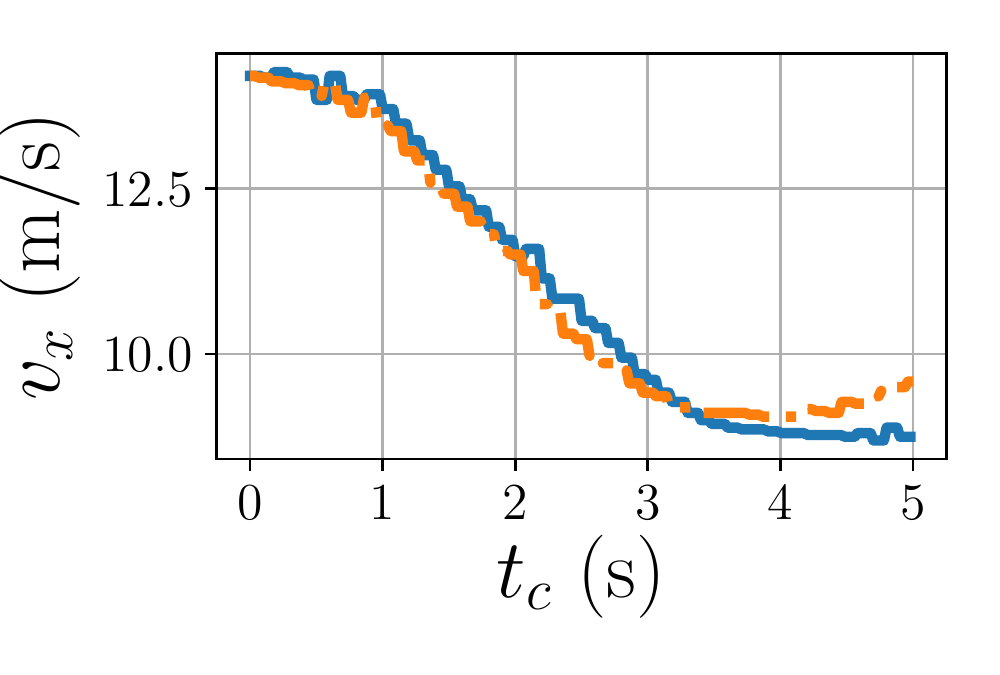}
        \label{fig:highmuCA:vx}
    }\\
    \subfloat[Slip angle of front wheel]{%
        \includegraphics[width=0.45\columnwidth]{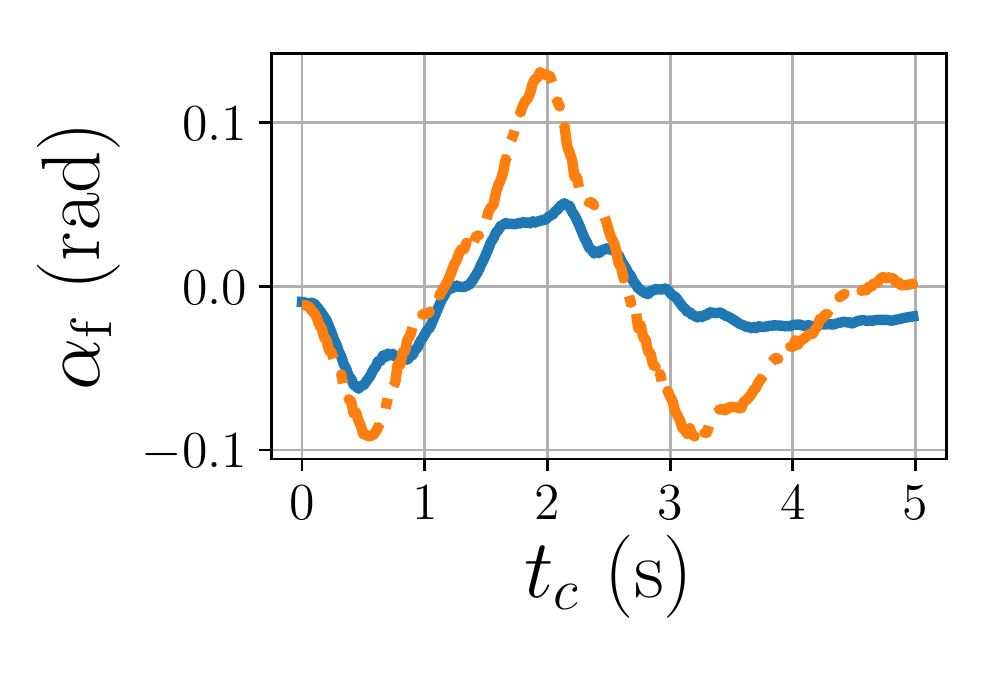}
        \label{fig:highmuCA:alphaf}
    }
    \subfloat[Lateral acceleration]{%
        \includegraphics[width=0.45\columnwidth]{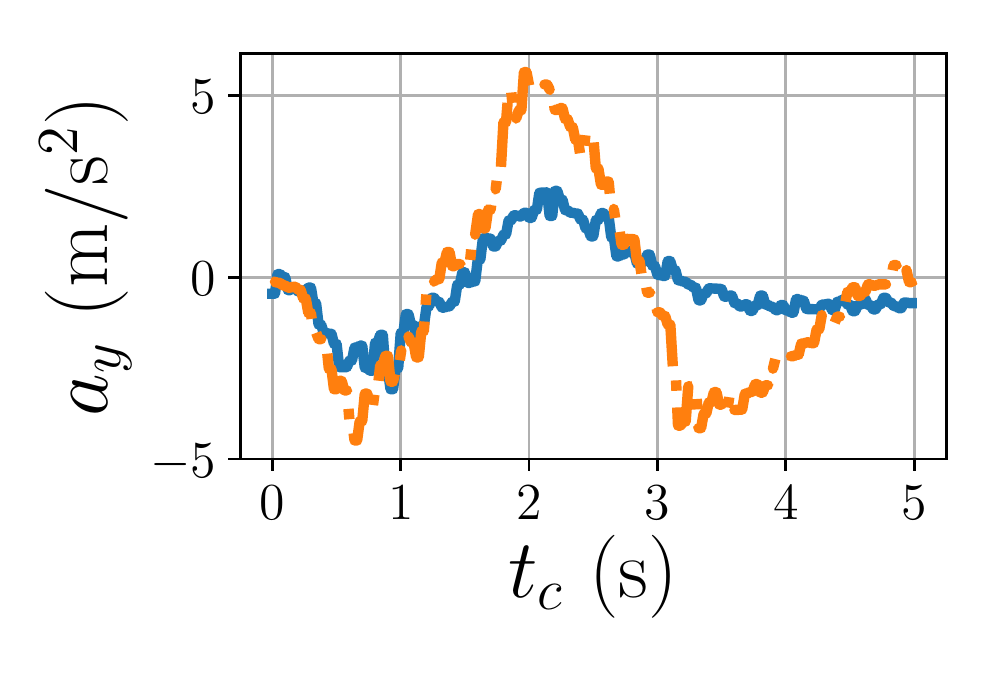}
        \label{fig:highmuCA:ay}
    }\\
    \end{minipage}\hfill
    \begin{minipage}[c]{0.45\textwidth}
    \centering
    \caption{Scenario (iii), collision avoidance at high $\mu$. Subfigures \ref{fig:highmuCA:NA0} through \ref{fig:highmuCA:TA2} show overhead visualizations of the vehicle position and the planned motion. The left and right columns of overhead views corresponds to the non-adaptive (NA) and traction adaptive (TA) schemes respectively. Subfigure \ref{fig:highmuCA:F} show planned tire forces and prevailing force limits for the front tire at $t_c=0$s. Subfigures \ref{fig:highmuCA:d}, \ref{fig:highmuCA:vx}, \ref{fig:highmuCA:alphaf} and \ref{fig:highmuCA:ay} show a comparison in terms of deviation from the lane center, forward velocity, front wheel slip angle and lateral acceleration. }
    \label{fig:highmuCA:full}
    \end{minipage}
\end{figure}

\subsection{Scenario (iii): Collision Avoidance at High Local Traction}
\label{sec:results:coll_avoid_high_mu}

A naive solution for handling scenarios 1 and 2 with static tire force constraints is to select a conservative static constraint, corresponding to the worst case traction. However, this strategy will restrict the utilization of the vehicle's capability at favorable traction conditions. Next, we investigate this aspect for collision avoidance at high $\mu$. 

In this scenario, $\mu_{\textnormal{gt}}(s) = 0.8$ for $s \geq 0$, and the pedestrian is detected 15m ahead of the vehicle. The initial velocity of the vehicle is 15m/s. The NA scheme operates under static tire force constraints corresponding to a \emph{conservative} static friction estimate of $\mu_{\textnormal{sta}} = 0.4$, while the adaptive scheme sets its tire force constraints dynamically according to $\mu_{\textnormal{est}}(s) = \mu_{\textnormal{gt}}(s)$ as in previous scenarios. Performance for the NA and TA schemes in this scenario is presented in Fig.~\ref{fig:highmuCA:full}. 

\subsubsection*{Not Adapting to High Local Traction}
For the NA scheme, we make the following observations. Upon detecting the pedestrian, Fig.~\ref{fig:highmuCA:NA0}, the planner initiates an evasive maneuver under conservative static tire force constraints. Given the vehicle state at the time when the planner/controller starts reacting to the obstacle, in combination with the restrictive tire force constraints, there is no collision free solution to the motion planning problem. As a result, the planner selects the least violating trajectory. At $t_c=0.7$s, Fig.~\ref{fig:highmuCA:NA1}, the vehicle is colliding with the obstacle at a velocity of 13.9 m/s. Notice that in this scenario the final motion of the vehicle corresponds well with the initial plan from the NA scheme, which was not the case in the low $\mu$ scenario. This is because in this case, the NA scheme is not violating the physical limitations in terms of tire forces, see Fig.
~\ref{fig:highmuCA:F} (blue dots). Rather, traction force utilization for the front wheel is consistently below $50\%$ throughout the maneuver. 
 
\subsubsection*{Adapting to High Local Traction}
Because $\mu_{\textnormal{est}}(s) > \mu_{\textnormal{sta}}$ in this case, adapting the tire force constraints means expanding the set of allowed tire forces compared to the NA case. Hence, the TA scheme is able to produce a more aggressive plan at $t_c=0$s, Fig.~\ref{fig:highmuCA:TA0}, that manages to avoid colliding with the obstacle. At $t_c=0.7$s, Fig.~\ref{fig:highmuCA:TA1}, the vehicle passes beside the obstacle as per the initial plan, and at $t_c=3$s, Fig.~\ref{fig:highmuCA:TA2}, the vehicle is returning to the lane, fully recovered from the incident. Fig.~\ref{fig:highmuCA:F} reveals that at $t_c=0$s, the TA scheme plans to exploit a larger part of the available traction. This is also reflected by the measured front slip angle, Fig.~\ref{fig:highmuCA:alphaf}. 

Also, in this scenario, because of the higher accelerations, the effect of dynamic normal loads become more apparent, see Fig.~\ref{fig:highmuCA:F}. In the beginning of the maneuver, when the vehicle is braking hard, the normal force, and hence the maximum horizontal force, is larger on the front tire and smaller on the rear. Notice that the TA scheme is able to exploit also this effect, and allocates 12\% higher front tire force to the front tire a the start of the maneuver (while the vehicle is braking hard) compared to the end of the maneuver (when the vehicle is no longer decelerating). Although this effect is small compared to the 50\% difference between the TA and NA schemes due to the friction estimate, it is not insignificant. The effect will be more pronounced at higher accelerations, and thus, is even more important to consider for vehicles that are smaller and more agile than our 8.5 ton test vehicle. 

The higher utilization of the available traction for the TA scheme leads to a quicker sideways motion compared to the NA scheme Fig.~\ref{fig:highmuCA:d}, and for this case, such a performance difference represents the difference between success and failure of the collision avoidance maneuver. Here, in contrast to Scenarios (i) and (ii), the TA scheme mitigates the critical situation by \emph{expanding} the tire force constraints instead of reducing them. In all three cases however, it utilizes around $90\%$ of the locally available traction. 

From this example we conclude that conservative friction estimation may reduce collision avoidance performance. To further investigate the extent of this phenomenon, we ran three configurations of the planning/ctrl schemes, TA with $\mu_{\textnormal{est}} = 0.8$, NA with $\mu_{\textnormal{sta}} = 0.6$ and NA with $\mu_{\textnormal{sta}} = 0.4$, through a sequence of 40 challenging collision avoidance scenarios each. A sequence of relative positions of the suddenly appearing obstacles was pre-generated by sampling from a uniform distribution, $-1.0\mathrm{m} \leq d_\mathrm{obs} \leq 1.0\mathrm{m}$, $s+11.0\mathrm{m} \leq s_\mathrm{obs} \leq s+13.0\mathrm{m}$. The initial velocity at obstacle detection was $10$m/s. 

\begin{figure}[t]
    \centering
    \includegraphics[width=0.6\columnwidth]{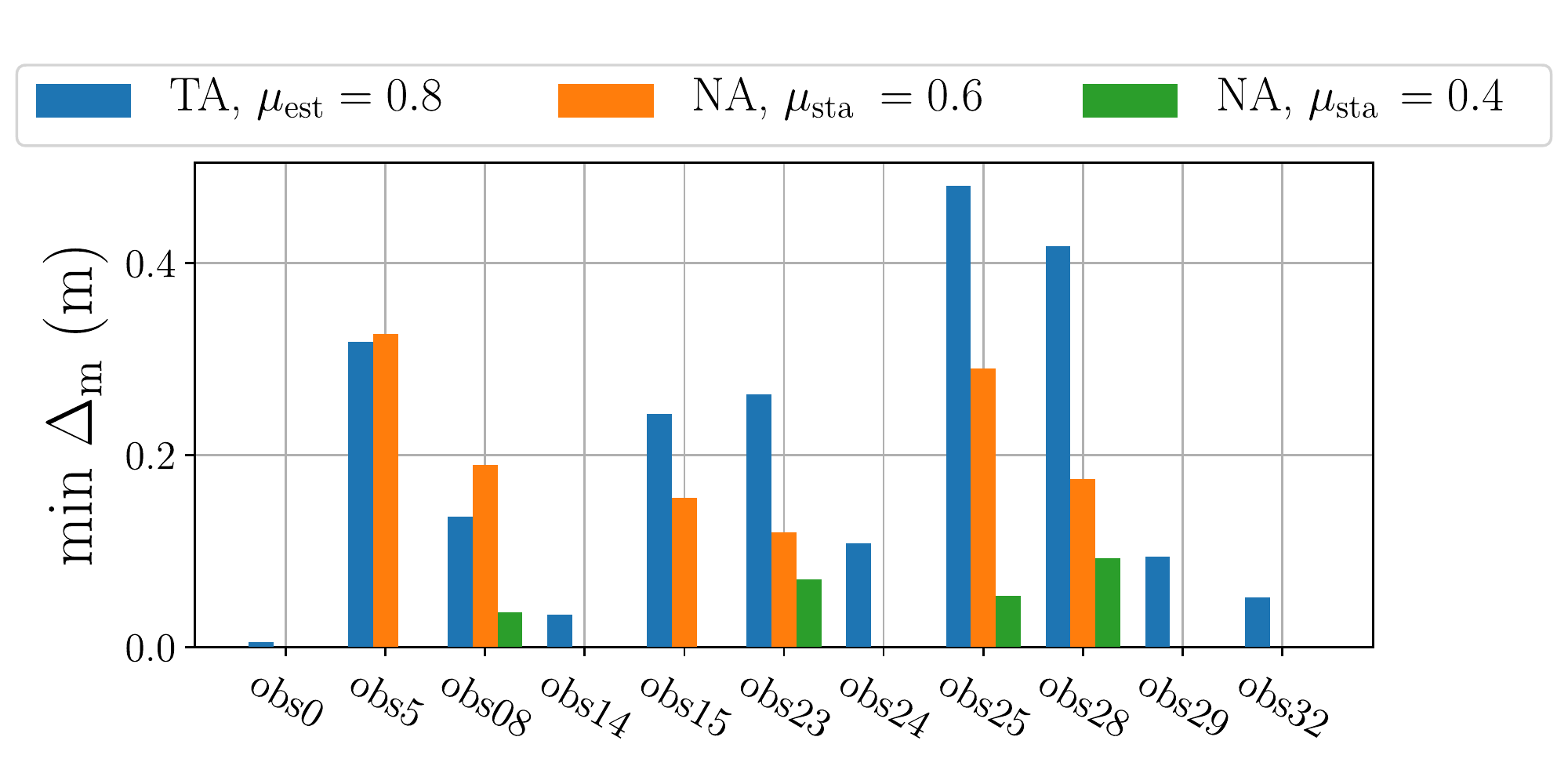}\\
    \caption{Minimum distances to avoided obstacles for three configurations of the planner: traction adaptive (TA) with $\mu_{\textnormal{est}} = 0.8$ (blue), non-adaptive (NA) with static tire force constraints corresponding to $\mu_{\textnormal{sta}} = 0.6$ (orange), non-adaptive (NA) with static tire force constraints corresponding to $\mu_{\textnormal{sta}} = 0.4$ (green). Zero value corresponds to collision. }
    \label{fig:ca_multi}
    \vspace{-0.3cm}
\end{figure}

With the three configurations, the vehicle managed to avoid collision in 11/40, 6/40 and 4/40 cases, respectively. Fig.~\ref{fig:ca_multi} shows individual differences in terms of minimum distance margin to the obstacle, $\Delta_\mathrm{m}$, for the 11 cases in which at least one configuration performed a successful avoidance maneuver. 

Note that the most conservative configuration (NA with $\mu_{\textnormal{sta}} = 0.4$), has the least amount of successful avoidance maneuvers, and the four successful avoidance maneuvers are very near misses, with less than $0.10$m of clearance to the obstacle. The least conservative configuration (TA with $\mu_{\textnormal{est}} = 0.8$) has the largest number of successful avoidance maneuvers and tends to also have the largest amount of clearance to the avoided obstacles. This is true for all cases except for obstacles 5 and 8, where (NA with $\mu_{\textnormal{sta}} = 0.6$) has slightly higher margin. We believe this stems from variations in the state of the vehicle when the obstacle is detected. In particular, the lateral position within the lane exhibited small variations during the experiment, which had noticeable effect on the margin to the obstacle. 

\subsection{Scenario (iv): Collision Avoidance with Discrete Decisions}
\label{sec:results:discrete_decisions}
As mentioned in Section \ref{sec:rti_lim}, the \RTIshort method handles situations when the initial guess $\hat{\mathcal{T}}_t$ violates state constraints, $\hat{x}_{k|t} \notin \mathcal{P}_{k|t}^\mathcal{X}$, for any $k \in \{0,1,\dots,N\}$, by softening the constraints and giving constraint violations a high cost. This in combination with the fact that the algorithm only searches for a solution locally around $\hat{\mathcal{T}}_t$ leads to sensitivity to highly suboptimal local minima. The problem occurs when the constraint configuration changes abruptly between planning iterations, e.g., at suddenly appearing obstacles, or in our approach to traction adaptation, at large changes in traction conditions. Therefore, we employ the sampling augmentation procedure, see Sections \ref{sec:feasibleinitguess} and \ref{sec:selectinitguess}, to mitigate this problem. In this section we evaluate the safety implications of sampling augmentation in critical scenarios with discrete decisions.

\begin{figure}[!]
\captionsetup[subfigure]{position=bottom} 
\begin{minipage}[c]{0.5\textwidth}
\centering
    \subfloat[RTI, $t_c=1$s]{%
        \includegraphics[width=0.45\columnwidth]{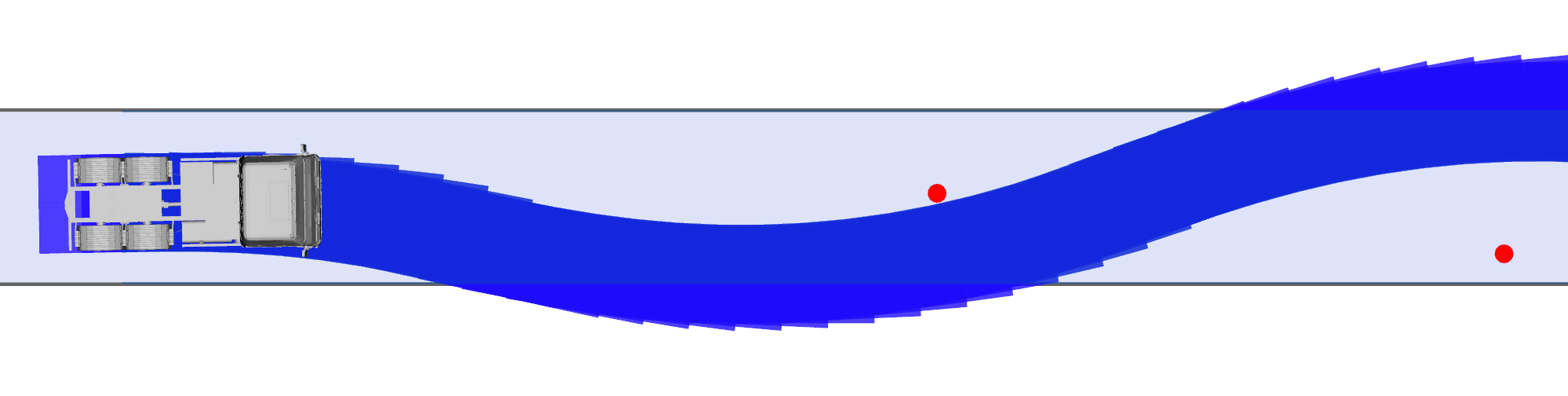}
        \label{fig:localmin:rti0}
    }
    \subfloat[SA, $t_c=1$s]{%
        \includegraphics[width=0.45\columnwidth]{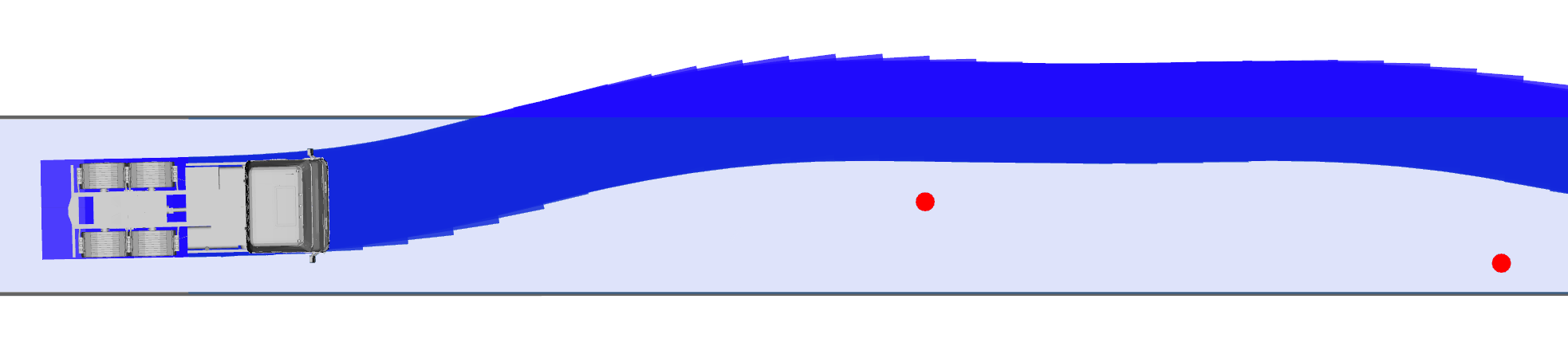}
        \label{fig:localmin:sarti0}
    }\\
    \subfloat[RTI, $t_c=2.5$s]{%
        \includegraphics[width=0.45\columnwidth]{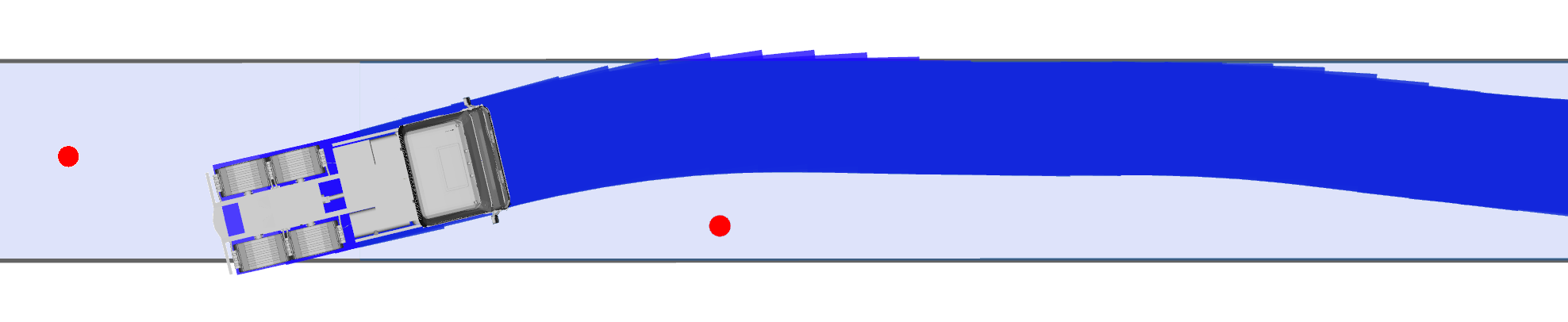}
        \label{fig:localmin:rti1}
    }
    \subfloat[SA, $t_c=2.5$s]{%
        \includegraphics[width=0.45\columnwidth]{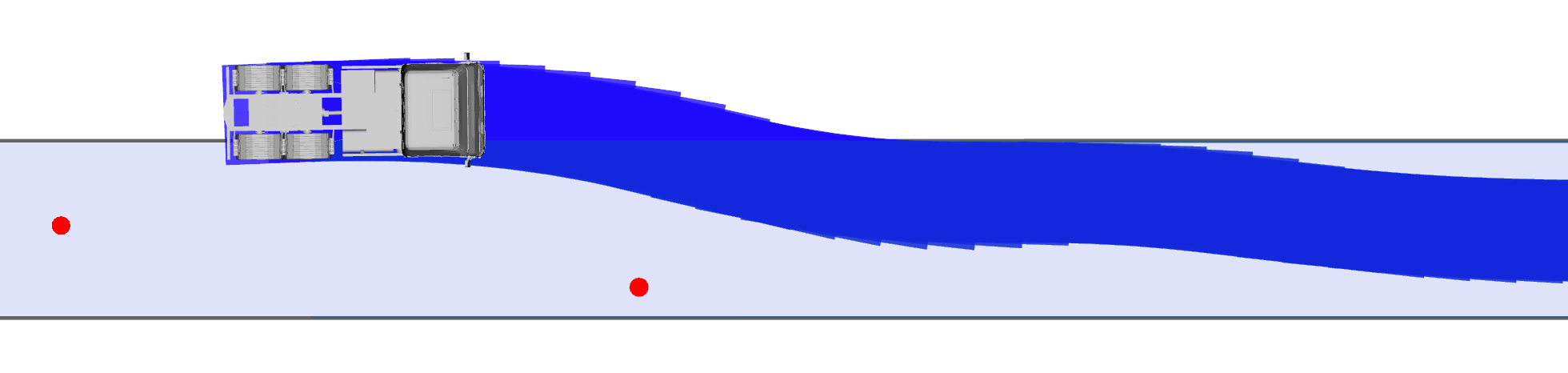}
        \label{fig:localmin:sarti1}
    }\\
    \subfloat[RTI, $t_c=5.2$s]{%
        \includegraphics[width=0.45\columnwidth]{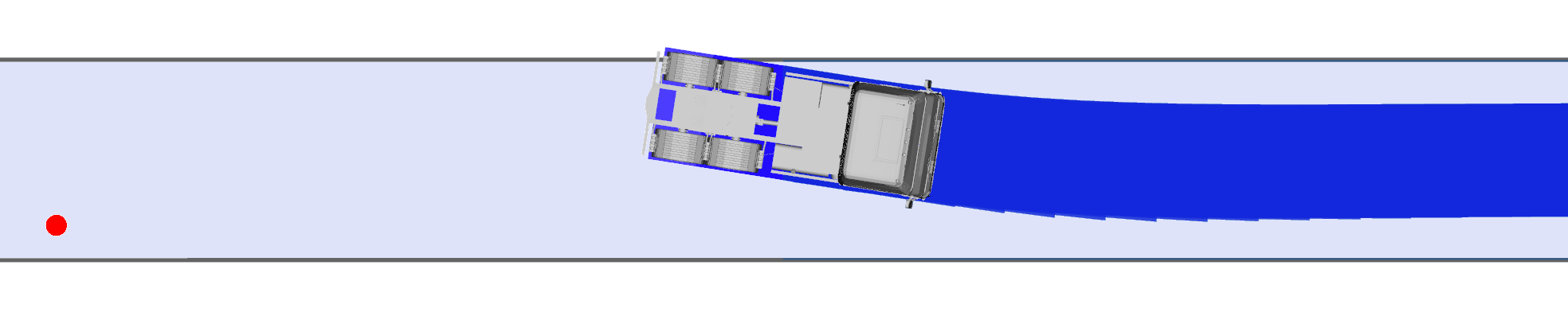}
        \label{fig:localmin:rti2}
    }
    \subfloat[SA, $t_c=4.5$s]{%
        \includegraphics[width=0.45\columnwidth]{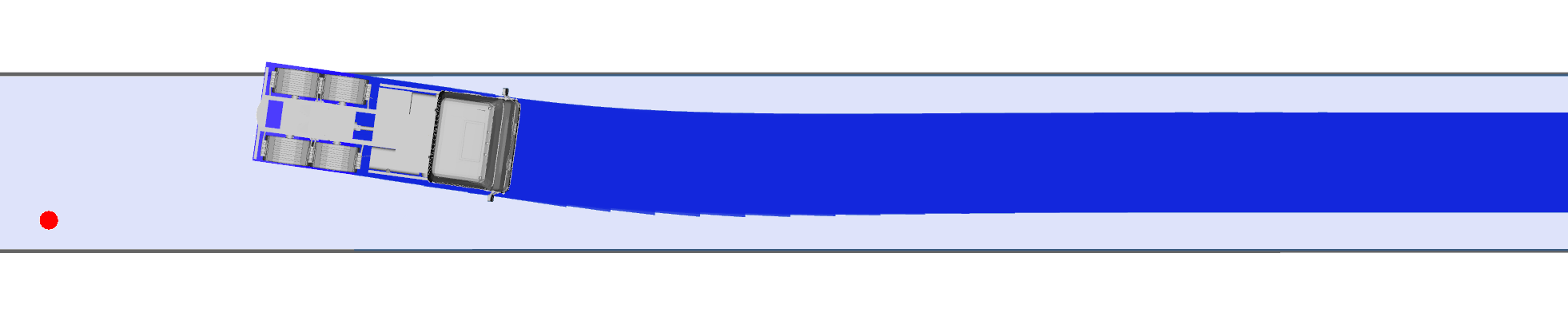}
        \label{fig:localmin:sarti2}
    }\\
    \subfloat[Planned tire forces at $t_c=1.0$s (corresponding to \ref{fig:localmin:rti0} and \ref{fig:localmin:sarti0})]{%
        \includegraphics[width=0.95\columnwidth]{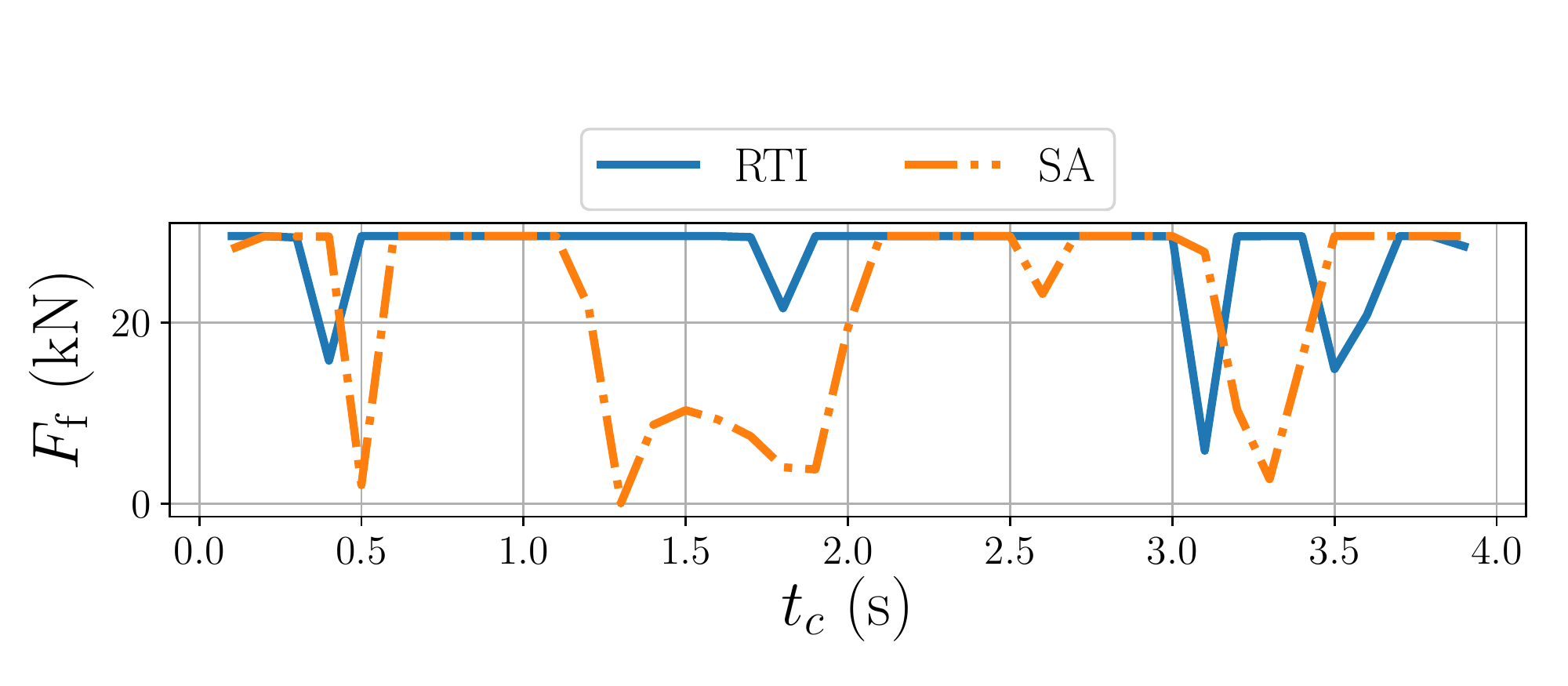}
        \label{fig:localmin:F}
    }\\
    \subfloat[Distance to obstacles]{%
        \includegraphics[width=0.45\columnwidth]{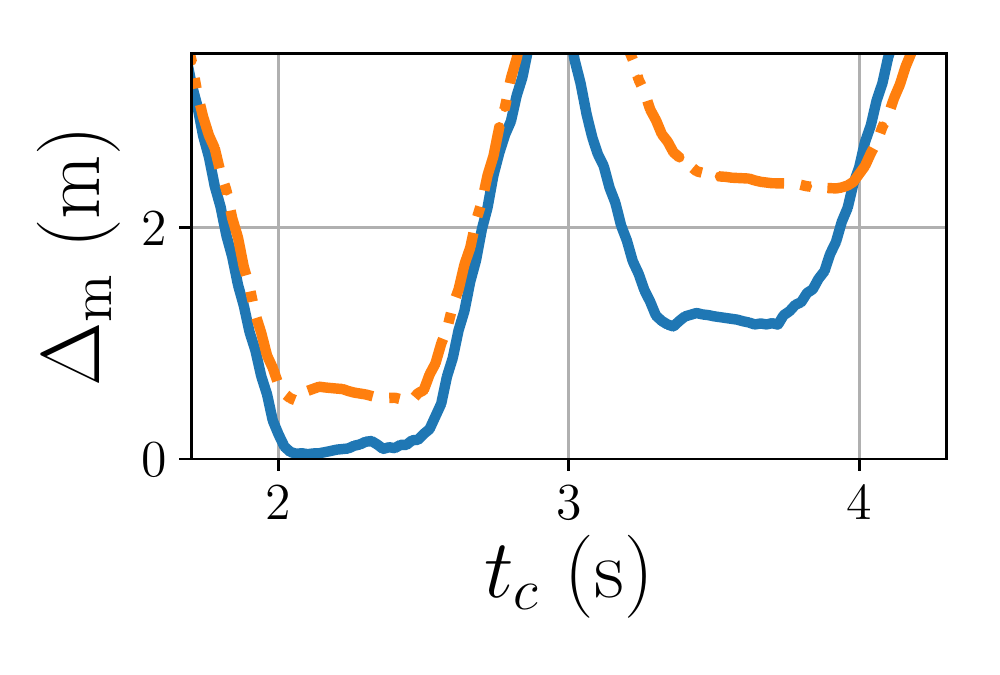}
        \label{fig:localmin:delta_m}
    }
    \subfloat[Initial guess selection with SA]{%
        \includegraphics[width=0.45\columnwidth]{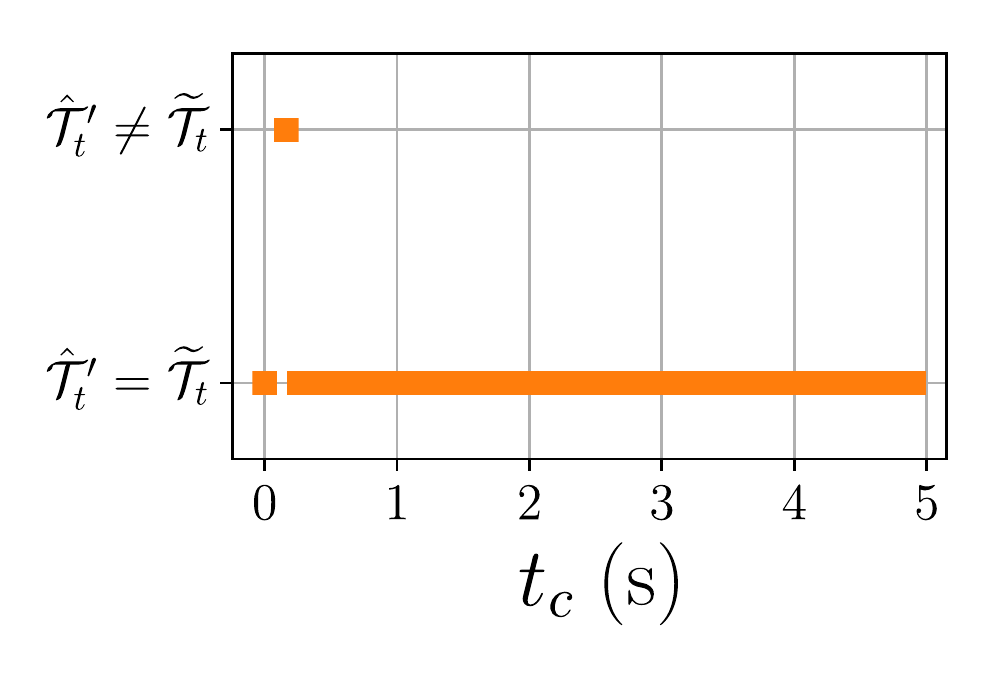}
    \label{fig:localmin:select}
    }
    \end{minipage}\hfill
    \begin{minipage}[c]{0.45\textwidth}
    \centering
    \caption{Scenario (iv), collision with discrete decisions. Subfigures \ref{fig:localmin:rti0} through \ref{fig:localmin:sarti2} show overhead visualizations of the vehicle position and the planned motion. The left and right columns of overhead views correspond to the non-augmented \RTIshort scheme (RTI) and sampling augmented (SA) schemes respectively. Subfigure \ref{fig:localmin:F} show planned tire forces for the front tires at $t_c=1.0$s. Subfigure \ref{fig:localmin:delta_m} shows a comparison in terms of smallest distance to either obstacle. Subfigure \ref{fig:localmin:select} shows which candidate was selected as initial guess with the SA scheme.}
    \label{fig:localmin:full}
    \end{minipage}
\end{figure}

Scenario (iv) is a collision avoidance scenario where two obstacles appear simultaneously 20m and 35m in front of the vehicle. The first obstacle appears slightly to the left of the lane center and the second obstacle appears on the right side of the lane. State constraints are set such that obstacles should be avoided by a margin of $0.5$m.
We compare performance of the non-augmented \RTIshort scheme (abbreviated RTI), with our sampling augmented variant (abbreviated SA). For this test both schemes use the same static tire force constraints. Results are presented in Fig.~\ref{fig:localmin:full}.  

\subsubsection*{Without 
Sampling Augmentation (RTI)}
When the obstacles are detected at $t_c=1.0$s, Fig~\ref{fig:localmin:rti0}, the non-augmented \RTIshort scheme produces an evasive plan that avoids the obstacles by going in between them (right of the first and left of the second obstacle). The vehicle manages to avoid collision, Figs.~\ref{fig:localmin:rti1}, \ref{fig:localmin:rti2}, but needs to use the maximum front tire force almost throughout the maneuver, Fig.~\ref{fig:localmin:F}, and has a very small distance margin $\Delta_\mathrm{m}$ to the obstacles (state constraints are violated), Fig.~\ref{fig:localmin:delta_m}.  

The decision to plan the maneuver between the obstacles stems from the last planning iteration before the obstacles appeared.
At the first planning iteration after the obstacle is detected, the solution from the previous iteration $\mathcal{T}^\star_{t-1}$ and hence the initial guess, $\hat{\mathcal{T}}_{t}$, is at the center of the lane. When the obstacles appear, $\hat{\mathcal{T}}_{t}$, violates the state constraints, $\hat{x}_{k|t} \notin \mathcal{P}_{k|t}^\mathcal{X}$ for some $k \in \{0,1,\dots, N\}$.
Since state constraints are soft, the optimization problem is still feasible, but the decision of adjusting the trajectory left or right of an obstacle is determined by minimizing cost locally around $\mathcal{T}^\star_{t-1}$. Since in this case, the first obstacle appears slightly to the left of the lane center, the planner makes the decision to go between the obstacles. 

\subsubsection*{With Sampling Augmentation (SA)}
When the sampling augmented \RTIshort algorithm is exposed to the same scenario, we observe that upon detecting the obstacle, the SA scheme instead produces an evasive plan to the left of both obstacles,  Fig.~\ref{fig:localmin:sarti0}. When following this plan, the vehicle is able to clear the obstacles more comfortably, Figs.~\ref{fig:localmin:rti1}, \ref{fig:localmin:rti2}. It uses the maximum front tire force for a smaller amount of the time, Fig.~\ref{fig:localmin:F}, keeps above the minimum margin to both obstacles (not violating state constraints), Fig.~\ref{fig:localmin:delta_m} and recovers more quickly after the incident (compare Fig.~\ref{fig:localmin:rti2} and Fig.~\ref{fig:localmin:sarti2}). 

Just as before, at the first planning iteration after the obstacles are detected, the initial guess $\tilde{\mathcal{T}}_{t}$ originating from $\mathcal{T}^\star_{t-1}$ is in collision with the obstacles. However, due to the sampling augmentation procedure, at $t_c=0.1$s, a sampled trajectory $\hat{\mathcal{T}}'_{t}$ is selected as initial guess. In the subsequent planning iterations throughout the maneuver, $\tilde{\mathcal{T}}_{t}$ is selected, Fig.~\ref{fig:localmin:select}. 

Fig.~\ref{fig:localmin:select} reveals that the sampled trajectories were only used in a single planning iteration just after the obstacles appeared. This indicates that sampling augmentation steps in exactly when needed, to avoid unfavorable local minima, i.e., at rapid changes in the constraint configuration of the motion planning problem. Otherwise, when changes to the constraint configuration are small, $\tilde{\mathcal{T}}_{t}$ typically has the lowest cost and gets selected. This result indicates that sampling augmentation alleviates Problem (iii) of Section \ref{sec:rti_lim} \RTIshort, without impairing the otherwise desirable properties of the \RTIshort method.

The sampling augmentation procedure should be viewed as an exploratory effort, that continuously evaluates a range of dynamically feasible maneuver alternatives, spread throughout the physically reachable drivable area, such that discrete decision making is made based on the full reachable state space of the vehicle (given current traction conditions), represented by $\hat{\mathcal{S}}_t \cup \tilde{\mathcal{T}}_{t}$ rather than being solely determined by local information around $\mathcal{T}^\star_{t-1}$.

\section{Conclusions and Future Work}
\label{sec:conclusions}
In this paper we address the problem of motion planning at the limits of handling under locally varying traction conditions. We evaluate the proposed traction adaptive algorithm by comparing it to an equivalent non-adaptive scheme in a sequence of critical scenarios at various traction conditions. 
Results from experimental evaluation indicate that traction adaptation improves capacity to avoid accident in all four tested critical scenarios. The proposed algorithm's performance in this regards stems from two properties:
\begin{enumerate} 
    \item Ensured dynamic feasibility of planned motions, even at varying local traction conditions.
    \item A high ratio (90\% for the tested configuration) of locally available traction is utilized to avoid collision if necessary.
\end{enumerate}

Moreover, we demonstrate experimentally that the sampling augmentation procedures (steps A-D of Section \ref{sec:method}) of the proposed algorithm successfully mitigates the computational challenges associated with infeasibility and sensitivity to local minima (listed in Section \ref{sec:rti_lim}) that emerges in the planning problem at rapid changes to the constraint configuration. 

In Section \ref{sec:relwork:friction_est} we briefly introduce a selection of approaches to obtaining predictive friction estimates online. It is intuitively clear that the characteristics and performance of such friction estimates will influence the planned behavior. We do not include that aspect in the scope of this paper, instead we present a follow-up study, dedicated to that topic \cite{svensson2021fusion}, where we conclude that combining camera based and traditional friction estimation techniques enable near-optimal traction adaptive motion planning and control. 
These preliminary results are obtained by emulating state of the art friction estimation methods in a simulated environment. Thus, the next step is to return to the test track and evaluate traction adaptive motion planning and control in closed loop with state of the art friction estimation.

\section*{Acknowledgments}
This work was supported in part by SAFER Open Research at AstaZero, Integrated Transport Resarch Lab at KTH, AutoDrive, grant agreement No 737469 and InSecTT, grant agreement No 876038. InSecTT (www.insectt.eu) has received funding from the ECSEL Joint Undertaking (JU). The JU receives support from the European Union’s Horizon 2020 research and innovation programme and Austria, Sweden, Spain, Italy, France, Portugal, Ireland, Finland, Slovenia, Poland, Netherlands, Turkey.
Disclaimer: The document reflects only the author’s view and the Commission is not responsible for any use that may be made of the information it contains.

In addition to the organizations that supported the project financially, the authors wish to express their gratitude towards Fredrik von Corswant, Henrik Biswanger as well as the staff at the AstaZero and Stora Holm test tracks for their support with the experiments. 
\bibliographystyle{ieeetr}
\bibliography{refs}
\begin{appendices}

\section{Experimental Setup}
\label{app:exp_setup}
The vehicle used for the experiments, depicted in Fig.~\ref{fig:motionfig:full}, was a Volvo FH16 6x4, a 750 hp tractor equipped with Volvo Dynamic Steering and I-Shift transmission, housed at the vehicle research laboratory \emph{Resource for Vehicle Research} (REVERE) at Chalmers University of Technology. Table~\ref{tab:static_params} outlines physical parameters of the vehicle.
The truck is equipped with multiple sensors to perceive the vehicle's state and surroundings, including lidar, stereo vision, differential GNSS and IMU.

\begin{table}[h] 
    \centering
    \begin{tabular}{|c|c|}
        \hline
        \bfseries Parameter & \bfseries Value\\
        \hline\hline
        $m$ & 8350 kg \\
        $I_z$ & 8150 $\text{kgm}^2$ \\
        $h$ & 1.0 m \\
        $l_{\textnormal{f}}$ & 1.20 m \\
        $l_{\textnormal{r}}$ & 2.20 m \\
        \hline
    \end{tabular}
    \vspace{0.2cm}
    \caption{Static physical parameters of test vehicle}
    \label{tab:static_params}
    \vspace{-0.3cm}
\end{table}

The computing system of the truck includes the microservices-based software stack OpenDLV\footnote{Cf.~\url{https://github.com/chalmers-revere/opendlv}} that enables transparent communication of distributed software components using inter-process communication (IPC), for example network communication and shared memory for low latency in real-time setups and for large volume sensor data handling.

The planning/control algorithms were integrated in a ROS stack of rudimentary autonomous functionality (localization, state estimation, track and object handling etc.) which was run on a PC with an Intel Core i7-7820HK CPU @ 2.9GHz and an Nvidia GeForce GTX 1070 GPU. On the truck's computing node, a ROS-to-OpenDLV bridge was realized for the project to enable the the ROS stack to access of the truck's sensors and actuators. 
Experiments were primarily conducted at the High Speed Area of the AstaZero test facility in Bor\aa s, Sweden, with complementary low $\mu$ tests at the Stora Holm test track in Gothenburg, Sweden. 

\section{Dynamic Bicycle Model with Time-varying Tire Force Limits in Road Aligned Coordinates}
\label{app:vehiclemodel}
\begin{figure}[h]
\captionsetup[subfigure]{}
\centering
    \subfloat[Top down view]{%
        \includegraphics[width=0.6\columnwidth]{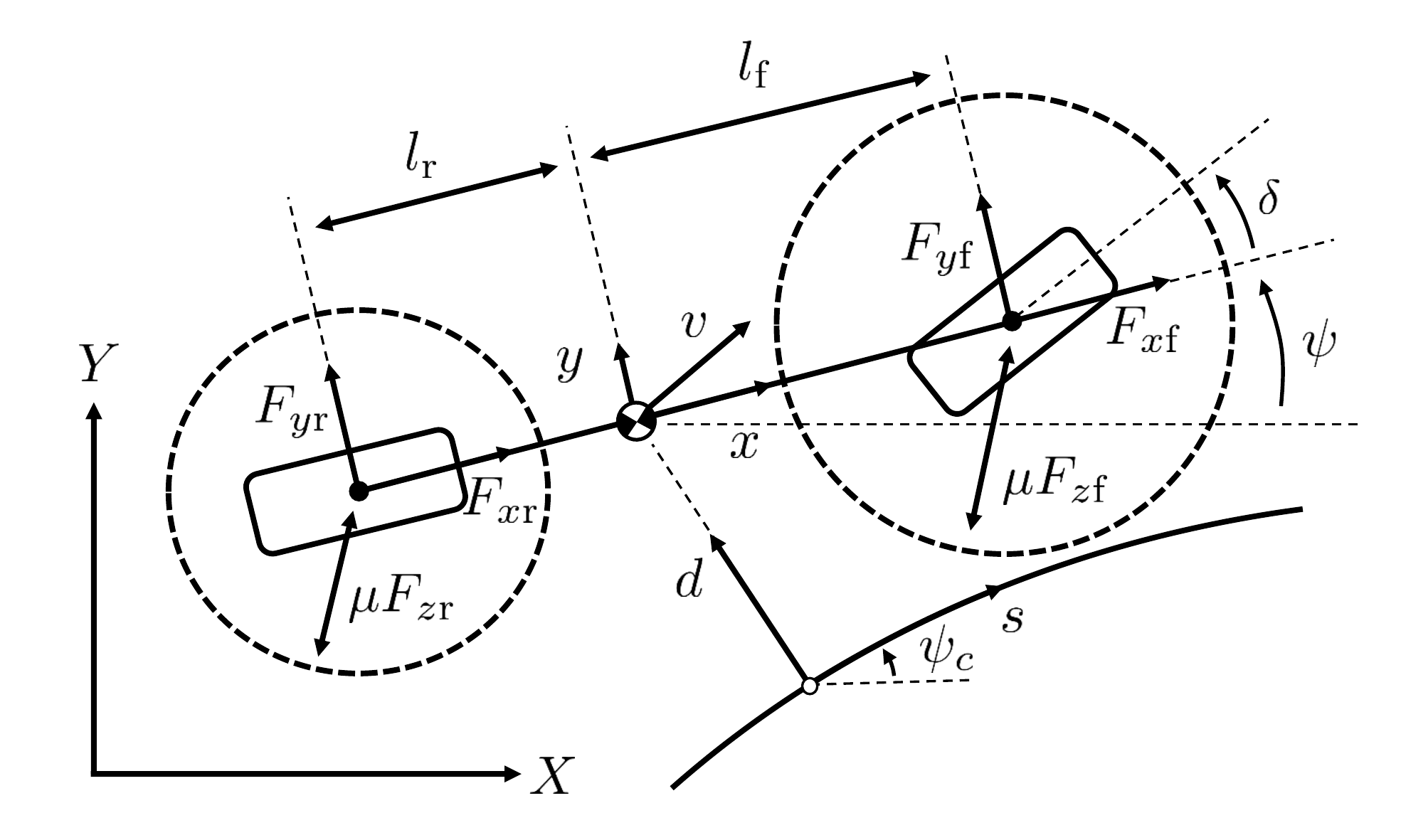}
        \label{fig:veh_model:dyn_model_frenet}
    }\\
    \subfloat[Left side view]{%
        \includegraphics[width=0.6\columnwidth]{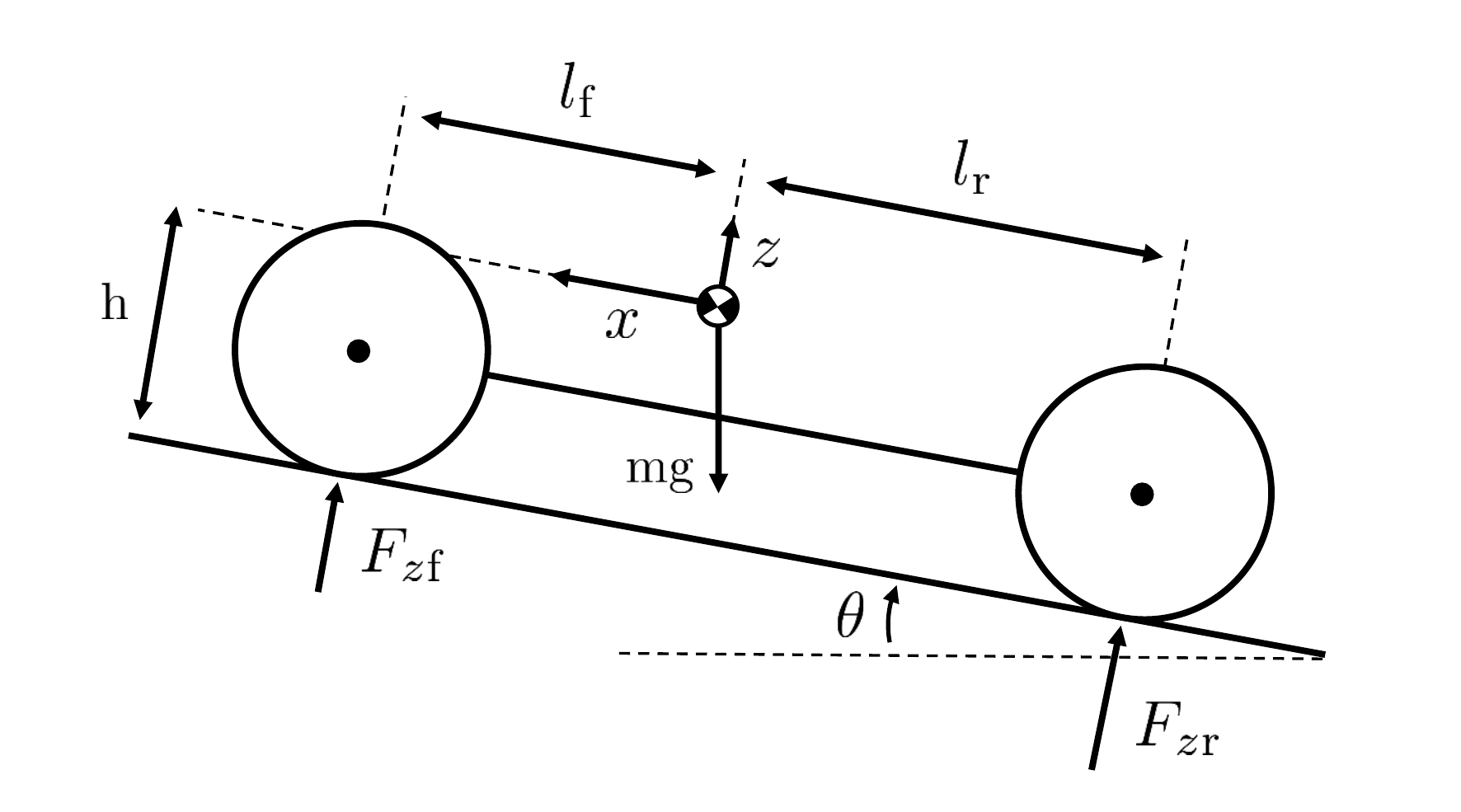}
        \label{fig:veh_model:normal_forces}
    }
    \caption{Geometric relations of the dynamic bicycle model with time-varying tire force limits in a road aligned coordinate frame.}
    \label{fig:veh_model:full}
\end{figure}

We derive a dynamic bicycle model based on standard vehicle dynamics literature \cite{rajamani2011vehicle}. See Fig.~\ref{fig:veh_model:full} for the geometric relations of the model.   
From Newton's second law for the lateral dimension we have that 
\begin{equation*}
\dot{v}_y = \frac{1}{m}(F_{{y\mathrm{f}}} + F_{{y\mathrm{r}}} ) - v_x \dot{\psi} + g \sin{\phi},
\end{equation*}
where $F_{{y\mathrm{f}}}$ and $F_{{y\mathrm{r}}}$ are lateral forces on the front and rear tires respectively. The term $v_x \dot{\psi}$ is the cross product term from the rotating vehicle-fixed coordinate system\footnote{Since $v_y$ is typically small, we ignore the corresponding term $v_y\dot{\psi}$ in the equation for $\dot{v}_x$.}
and $g \sin{\phi}$ is the contribution from the bank angle of the road. 
%
As for the longitudinal dynamics, we have that
\begin{equation*}
\dot{v}_x = \frac{1}{m}(F_{x\mathrm{f}} + F_{x\mathrm{r}} ) - g\sin(\theta),
\end{equation*}
where $F_{x\mathrm{f}}$ and $F_{x\mathrm{f}}$ are the controlled tractive forces on the front and rear tires respectively. The term $g\sin(\theta)$ represent the longitudinal acceleration contribution due to the inclination of the road. 
The yaw dynamics are given by moment balance around the $z$-axis as
\begin{equation*}
\ddot{\psi} = \frac{1}{I_z} \left( l_{\mathrm{f}} F_{{y\mathrm{f}}} - l_{\mathrm{r}} F_{{y\mathrm{r}}} \right),
\end{equation*}
where $I_z$ is the vehicle's moment of inertia about the $z$-axis, and $l_\mathrm{f}$ and $l_\mathrm{f}$ are the distances from the center of mass to the front and rear axle respectively. 
In order to capture the motion of the vehicle relative to the road we use a road aligned coordinate frame \cite{micaelli1993trajectory}.
A coordinate $s$ represents the progress along the centerline of the lane and a coordinate $d$ represents the lateral deviation from the centerline, as depicted in Fig.~\ref{fig:veh_model:dyn_model_frenet}. The $d$-coordinate increases to the left. Variables $\psi_c$ $\kappa_c$ denote the path tangent, and path curvature at the point on the centerline, where the vehicle position is perpendicular to the centerline tangent. $\Delta \psi = \psi-\psi_c$ represents the relative angle between the vehicle yaw angle and the tangent of the centerline. Geometric relations of the coordinate system definition give us the evolution of $s$, $d$ and $\Delta \psi$. Finally we express the continuous time dynamics as
\begin{align}\label{eq:dynamics_cont}
& \dot{s} = \frac{v_x \cos{(\Delta \psi)} - v_y \sin{(\Delta \psi)}}{1-d \kappa_c}, \nonumber \\
& \dot{d} = v_x\sin{(\Delta \psi)} + v_y \cos{( \Delta \psi)}, \nonumber \\
& \Delta \dot{\psi} = \dot{\psi} - \kappa_c \frac{v_x \cos{(\Delta \psi)} - v_y \sin{(\Delta \psi)}}{1-d \kappa_c}, \nonumber \\
& \ddot{\psi} = \frac{1}{I_z} \left( l_{\mathrm{f}} F_{{y\mathrm{f}}} - l_{\mathrm{r}} F_{{y\mathrm{r}}} \right),  \nonumber \\
& \dot{v}_x = \frac{1}{m} \left( F_{{x\mathrm{f}}} + F_{{x\mathrm{r}}} \right) - g \sin( \theta), \nonumber    \\
& \dot{v}_y = \frac{1}{m} \left( F_{{y\mathrm{f}}} + F_{{y\mathrm{r}}} \right) - v_x \dot{\psi} + g \sin{\phi}.
\end{align}
By selecting the state and control vectors as $ x = [s, d, \Delta \psi, \dot{\psi}, v_{x}, v_{y}]^\top$ and $ u = [F_{{y\mathrm{f}}}, F_{{x\mathrm{f}}}, F_{{x\mathrm{r}}}]^\top$ we can compactly write the planning model \eqref{eq:dynamics_cont} as $\dot{x} = f_p\left(x,u\right)$. 

The rear lateral force $F_{y\mathrm{r}}$ is not included among the control inputs, instead it is computed from the state through a time-varying linear tire model 
\begin{align} \label{eq:linear_tire}
    F_{{y\mathrm{r}}} = \tilde{C}_\mathrm{r}(\mu) \alpha_\mathrm{r}
\end{align}
with the rear slip angle $\alpha_\mathrm{r} = -\arctan{(( v_y-l_\mathrm{r} \dot{\psi})) / v_x }$. The time-varying rear cornering stiffness $\tilde{C}_\mathrm{r}$ is obtained by linearizing Pacejka's Magic Formula tire model, \cite{pacejka1992magic},
\begin{align*}
    F_y = & D(\mu) \sin \left( \right. C(\mu) \arctan \left( \right. B(\mu) \alpha_\mathrm{f} + \cdots  \\ 
    & - E(\mu) \left( \right. B(\mu) \alpha_\mathrm{f} - \arctan \left( \right. B(\mu) \alpha_\mathrm{f} \left. \right) \left. \right) \left. \right) \left. \right),
\end{align*}
with parameters $B(\mu)$, $C(\mu)$, $D(\mu)$ and $E(\mu)$ obtained as functions of $\mu$ from a look-up table that approximately associate the tire parameters to the friction coefficient for various road surface conditions. 
\begin{remark}
In general, the linear tire model used in \eqref{eq:linear_tire} does not accurately predict the relation between slip angle and tire force at high slip angles \cite{rajamani2011vehicle}. However, in this context, the combined lateral and longitudinal force command is always inside the friction circle, by a margin determined by the utilization factor $\lambda$. Inside such a boundary, a time-varying linear model corresponds well with the Magic Formula.
Considering the computational trade-off mentioned in Section \ref{sec:prel}, this design represents a compromise between complexity and accuracy that have proven to be practically viable for our application. 
\end{remark}

The normal forces $F_{z\mathrm{f}}$ and $F_{z\mathrm{r}}$ acting on the tires of the model are obtained from moment balances about the contact points of the rear and front tires respectively as
\begin{equation}
\begin{aligned}
    &F_{z\mathrm{f}} = \frac{1}{l_\mathrm{f}+l_\mathrm{r}} \left( -m \dot{v}_x h - mgh \sin{\theta} + mgl_\mathrm{r} \cos{\theta} \right), \\
    &F_{z\mathrm{r}} = \frac{1}{l_\mathrm{f}+l_\mathrm{r}} \left( m \dot{v}_x h + mgh \sin{\theta} + mgl_\mathrm{f} \cos{\theta} \right).
    \label{eq:pitchdynamics}
\end{aligned}
\end{equation}
See geometric relations in Fig.~\ref{fig:veh_model:normal_forces}. 
Combining these relations with the friction estimate $\mu$ we define front and rear \emph{time-varying}\footnote{
Variables $\mu$, $F_{z\mathrm{f}}$ and $F_{z\mathrm{r}}$ vary in time and hence, so do $F_\mathrm{f}^{(\mathrm{ub})}$ and $F_\mathrm{r}^{(\mathrm{ub})}$. Just as with the state variables in \eqref{eq:dynamics_cont} the time dependency is omitted from the equations to simplify notation.}
force boundaries as
\begin{equation*}
    F_\mathrm{f}^{(\mathrm{ub})} = \lambda \mu F_{z\mathrm{f}}, \quad F_\mathrm{r}^{(\mathrm{ub})} = \lambda \mu F_{z\mathrm{r}},
\end{equation*}
where $\lambda \in [0,1]$ is a user adjustable traction utilization factor. If $\lambda = 1.0$, the algorithm is allowed to utilize the full friction circle. The experiments for this paper were run at $\lambda = 0.9$. To comply with these bounds, tire force commands need to fulfill
\begin{equation} \label{eq:tireforce_bounds}
\begin{aligned}
    \sqrt{F_{{x\mathrm{f}}}^2 + F_{{y\mathrm{f}}}^2} \leq &  F_\mathrm{f}^{(\mathrm{ub})}, \quad \sqrt{F_{{x\mathrm{r}}}^2 + F_{{y\mathrm{r}}}^2} \leq &  F_\mathrm{r}^{(\mathrm{ub})},
\end{aligned}
\end{equation}
where $F_{y\mathrm{r}}$ is determined by \eqref{eq:linear_tire}.
%
In addition to the bounded tire forces \eqref{eq:tireforce_bounds}, the drivetrain of the vehicle imposes additional constraints
\begin{equation} \label{eq:drivetrain_bounds}
    F_{{x\mathrm{f}}} \leq ~0, \quad
    F_{{x\mathrm{r}}} \leq ~F_\mathrm{r}^{(\mathrm{ub,dr})},
\end{equation}
due to rear wheel drive and limited tractive force from the motor $F_\mathrm{r}^{(\mathrm{ub,dr})}$. 

\begin{figure}[h]
    \centering
    \includegraphics[width=0.6\columnwidth]{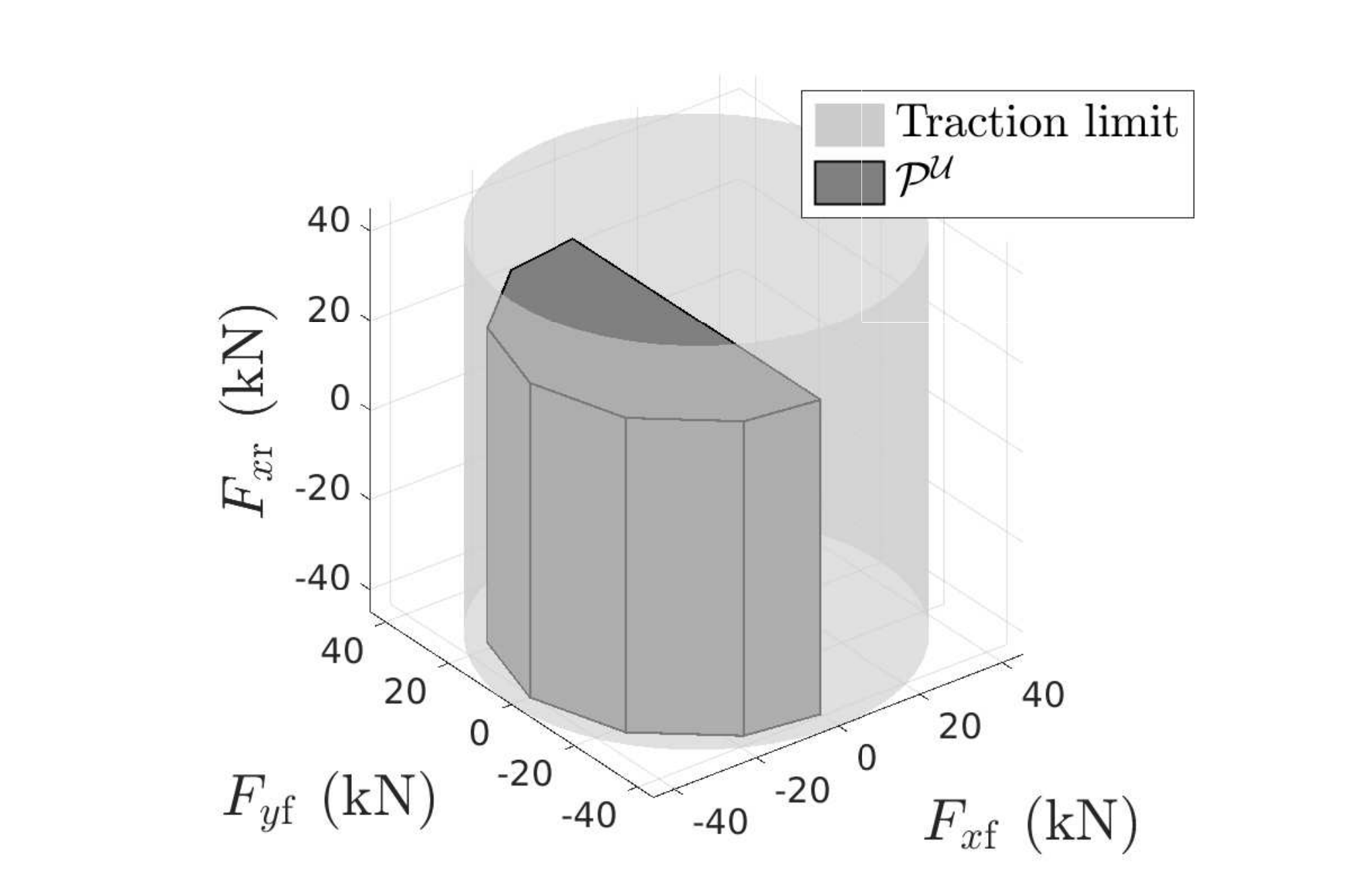}\\
    \caption{Input constraint polytope $\mathcal{P}^\mathcal{U}$ (for single prediction step $k$) bounding control signals produced by the algorithm (dark gray) and theoretical traction limits according to \eqref{eq:forcelimit}.}
    \label{fig:U_polytope}
    \vspace{-0.3cm}
\end{figure}

Finally, the algorithmic framework used in this paper requires linear input constraints. To accommodate this, we under-approximate the input constraints \eqref{eq:tireforce_bounds} and \eqref{eq:drivetrain_bounds} by a three-dimensional convex polytope $\mathcal{P}^\mathcal{U}$, specific to each prediction step $k$, visualized in Fig.~\ref{fig:U_polytope}.

\section{Low-Level Control Interface}
\label{app:ctrl_interface}
The purpose of the control interface is to translate tire force commands, $F_{{y\mathrm{f}}}$, $F_{{x\mathrm{f}}}$ and $F_{{x\mathrm{r}}}$ from the planned trajectory $\mathcal{T}^\star_t$  into the actual control inputs of the vehicle. For our particular test vehicle, those are steering angle request $\delta_{\mathrm{req}}$ and longitudinal acceleration request $a_\mathrm{req}$.

For the longitudinal control input, the transformation is trivially computed as $a_\mathrm{req} = (F_{x\mathrm{f}} + F_{x\mathrm{r}})/m$.   
For the steering input $\delta_{\mathrm{req}}$, we begin by computing the kinematic component of the steering angle, given by $\delta_\mathrm{kin}$, i.e., the steering angle that would yield the planned motion in the absence of tire slip. We fit a circle segment of curvature $\rho$ to the Cartesian representation of the position states $s_{k|t}^\star$ and $d_{k|t}^\star$ for $k \in \{0,1,\dots,(N_\mathrm{fit})\}$ of $\mathcal{T}^\star_t$, where $N_\mathrm{fit}$ is selected such that only the initial part of the trajectory is used. Then, the kinematic steering angle component is obtained as $\delta_\mathrm{kin} = \rho(l_\mathrm{f}+l_\mathrm{r})$.  
The dynamic steering angle component, i.e., the desired front wheel slip angle, is obtained as $\alpha_\mathrm{f} = F_{y\mathrm{f}}/\tilde{C}_\mathrm{f}$, where the time-varying cornering stiffness $\tilde{C}_\mathrm{f}$ is obtained by linearizing the Pacejka Magic Formula tire model as per the procedure described in Appendix \ref{app:vehiclemodel}.
The steering angle request is then obtained as $\delta_{\mathrm{req}} = \delta_\mathrm{kin} + \alpha_\mathrm{f}$. Finally, the control commands $\delta_{\mathrm{req}}$ and $a_\mathrm{req}$ are passed to the vehicle via the ROS-to-OpenDLV bridge described in Appendix \ref{app:exp_setup}.

\end{appendices}

\end{document}